\documentclass[12pt]{article}
\usepackage{amsmath}
\usepackage[usenames, dvipsnames]{xcolor}

\usepackage{tocloft}
\usepackage[]{todonotes}
\usepackage{verbatim}
\usepackage{mathrsfs}
\usepackage[margin=.8in]{geometry}
\usepackage{dsfont}
\usepackage{tikz}
\usepackage{setspace,caption}
\usepackage{lipsum}
\usetikzlibrary{matrix,arrows,calc}
\usepackage{slashed}        \usepackage{graphicx}       \usepackage{subcaption}     \usepackage{psfrag}       \usepackage{tensor}       \usepackage{fouridx}        \usepackage{bm}         \usepackage{multirow}       \usepackage{soul}         \usepackage{bbold}        \usepackage{multicol}       \usepackage{tikz-cd}
\usepackage{rotating}
\usepackage{soul}

\usetikzlibrary{arrows}

 \tikzset{
commutative diagrams/.cd,
arrow style=tikz,
diagrams={>=latex}}

\usepackage{amssymb}
\usepackage{mathtools}

\usepackage{feynmf}
\usepackage{marvosym}

\usepackage{import}

\usepackage[utf8]{inputenc}
\usepackage{amsmath}
\usepackage{tikz}
\usepackage{extpfeil}
\usepackage{amsfonts}
\usepackage{amssymb} 
\usepackage{bm}
\usepackage{graphicx}
\usepackage{array}
\usepackage{lipsum}
\usepackage{float}
\usepackage{multirow}
\usepackage{hhline}
\usepackage{tabularx}
\usepackage{graphicx}
\usepackage{afterpage}
\usepackage{longtable}
\usepackage{epstopdf}

\usepackage[titletoc]{appendix}

\newcommand{\fontsizeshift}{12pt}

\newcommand{\cN}{\mathcal{N}}

\newcommand{\cO}{\mathcal{O}}
\newcommand{\bE}{\mathbb{E}}

\newcommand{\bR}{\mathbb{R}}

\newcommand{\amnew}[1]{{\color{black} #1}}
\newcommand{\jimnew}[1]{{\color{black} #1}}
\newcommand{\amred}[1]{{\color{black} #1}}

\newcommand{\be}{\begin{equation}}
\newcommand{\ee}{\end{equation}}
\newcommand{\bea}{\begin{eqnarray}}
\newcommand{\eea}{\end{eqnarray}}

\newcommand{\gnet}{Gauss-net}

\newcommand{\din}{d_\text{in}}
\newcommand{\dout}{d_\text{out}}
\newcommand{\sgp}{S_{\text{GP}}}
\newcommand{\zgp}{Z_{\text{GP}}}
\newcommand{\ggp}{G_{\text{GP}}}
\newcommand{\zgpzero}{Z_{\text{GP},0}}
\newcommand{\wick}{\text{Wick}}

\newcommand{\sumabcdef}{\sum_{\mathcal{P}(abcdef)}}

\definecolor{cobalt}{RGB}{44, 98, 120}
\definecolor{celadon}{rgb}{0.67, 0.88, 0.69}
\definecolor{dm}{cmyk}{.20, 0, .30, 0}
\definecolor{burgundy}{rgb}{0.5, 0.0, 0.13}
\definecolor{plotBlue}{RGB}{94, 130, 181}

\usepackage{amsfonts}
\usepackage{amsthm}
\usepackage{amsmath, multicol}
\usepackage{indentfirst}
\usepackage{graphicx}
\usepackage{graphics}
\usepackage{braket}
\usepackage[T1]{fontenc}
\usepackage[english]{babel}
\usepackage{color,soul}
\numberwithin{equation}{section}

\usepackage{url}

\usepackage{flip-acronyms} \usepackage{tikzfeynman}

\newcommand{\email}[1]{\url{#1}}

\newcommand*\xoverline[2][0.75]{    \sbox{\myboxA}{$\m@th#2$}    \setbox\myboxB\null    \ht\myboxB=\ht\myboxA    \dp\myboxB=\dp\myboxA    \wd\myboxB=#1\wd\myboxA    \sbox\myboxB{$\m@th\overline{\copy\myboxB}$}    \setlength\mylenA{\the\wd\myboxA}    \addtolength\mylenA{-\the\wd\myboxB}    \ifdim\wd\myboxB<\wd\myboxA       \rlap{\hskip 0.5\mylenA\usebox\myboxB}{\usebox\myboxA}    \else
        \hskip -0.5\mylenA\rlap{\usebox\myboxA}{\hskip 0.5\mylenA\usebox\myboxB}    \fi}
\makeatother

\begin{document}

\newcommand{\main}{.}
\begin{titlepage}

\setcounter{page}{1} \baselineskip=15.5pt \thispagestyle{empty}

\bigskip\

\vspace{2cm}
\begin{center}
{\LARGE \bfseries Neural Networks and Quantum Field Theory}
\end{center}

\vspace{0.25cm}

\begin{center}

James Halverson, Anindita Maiti, and Keegan Stoner \\ 

\vspace{.5 cm}
\emph{Department of Physics, Northeastern University \\ Boston, MA 02115}\\

\vspace{.5 cm}
\small{\texttt{\{j.halverson,  maiti.a,  stoner.ke\}@northeastern.edu}}

\vspace{0.5cm}
\end{center}

\vspace{1cm}
\noindent

We propose a theoretical understanding of neural networks in terms of Wilsonian effective field theory. 
The correspondence relies on the fact that many asymptotic neural networks are drawn from Gaussian processes, the analog of non-interacting field theories. 
Moving away from the asymptotic limit yields a non-Gaussian process and corresponds to turning on particle interactions,
allowing for the computation of correlation functions of neural network outputs with Feynman diagrams.
Minimal non-Gaussian process likelihoods are determined by the most relevant non-Gaussian terms, according to the flow in their coefficients induced by the Wilsonian renormalization group. This yields a direct connection between overparameterization and simplicity of neural network likelihoods.
Whether the coefficients are constants or functions may be understood in terms of GP limit symmetries,
as expected from 't Hooft's technical naturalness.
General theoretical calculations are matched to neural network experiments in the simplest class of models allowing the correspondence.
Our formalism is valid for any of the many architectures that becomes a GP in an asymptotic limit, a property preserved under
certain types of training.

\noindent
 \vspace{3.1cm}

\end{titlepage}
\tableofcontents
\newpage

\section{Introduction \label{sec:intro}}

The relationship between asymptotic neural networks and Gaussian
processes provides a strong hint towards 
a theoretical understanding of deep learning. Rather than considering a neural network to be determined by draws from a parameter space distribution, this perspective considers neural networks themselves as draws from a function space distribution. The essential idea is that a family
of neural network architectures
\be
f_{\theta,N}: \bR^{d_\text{in}} \to \bR^{d_{out}}
\ee
indexed by parameters $\theta$ and a discrete hyperparameter $N$ admits a limit $N\to \infty$
in which networks are drawn from a Gaussian process (GP), i.e. a Gaussian
distribution on function space, 
\be 
P[f] \sim \text{exp}\left[-\frac12 \int d^{d_\text{in}}x \, d^{d_\text{in}}x' f(x) \Xi(x,x') f(x')\right],
\ee 
where the functional / operator inverse of $\Xi(x,x')$ is the GP kernel. Of 
course, in practice one usually studies networks with large-but-finite 
$N$. These should be drawn from a distribution that receives $1/N$
corrections relative to the Gaussian distribution, i.e., a non-Gaussian
process (NGP). Learning is then a data-induced flow of the function space distribution; recent literature and the above argument together suggest that the distribution remains an NGP during training.

The idea that neural networks are drawn from a non-Gaussian (but close-to-Gaussian) distribution on function space is immediately suggestive to physicists: such mathematics provides the backbone of perturbative quantum
field theory (QFT), the framework that underlies numerous physical systems, from superconductors to the Standard Model of particle physics.
From 
the perspective of Feynman's path integral, non-interacting (``free'') 
quanta or particles are described by appropriate Gaussian
distributions on function (field) space, where details depend on symmetries and
particle properties such as Lorentz or rotational invariance and spin. When the fields do not have any vacuum expectation value the corresponding GP has mean zero and the $2n$-pt
correlation functions are entirely determined by the $2$-pt
statistics. Physically, this corresponds to $n$ particles propagating past one
another without interacting. Such quantum field theories are exactly solvable
due to their Gaussian nature. Interactions between particles arise precisely due to 
non-Gaussian corrections to the log likelihood, known as the action
in physics.

\vspace{.5cm}

In this paper\footnote{We use alphabetical authorship as in high energy physics. Maiti and Stoner contributed equally to this work.} we develop a sharp correspondence\footnote{Developing the correspondence
will require using language from both communities in a way that will sometimes be 
obvious to experts in one of the fields. We hope the additional clarity is worth the tedium.} between neural networks
and quantum field theory. We will introduce a framework
known as Wilsonian effective field theory (EFT) for studying neural networks, utilizing it to determine minimal NGP likelihoods associated with neural network architectures,
making modifications from the usual physics contexts when necessary. 
For brevity, we will refer to the general idea as a NN-QFT correspondence.

\textbf{Related work.}  An EFT approach to neural networks is possible whenever a family of architectures admits a GP limit, which is the case for many modern
architectures. Though the original NN-GP correspondence \cite{neal}
was in the context of infinitely wide single-layer fully-connected networks,
which admit computable kernels \cite{williams}, recent work has 
shown that infinitely wide deep fully-connected networks \cite{lee,Matthews2018GaussianPB}
are drawn from GPs, as are deep convolutional networks in the
infinite channel limit \cite{Novak2018BayesianCN,GarrigaAlonso2019DeepCN}.
In \cite{yangTPorig,yangTP1, yangTP2}, Yang developed a language for understanding which architectures admit GP limits, which was utilized to demonstrate that any standard architecture admits a GP limit, i.e. any architecture that is a composition of multilayer perceptrons, recurrent neural networks, skip connections \cite{skip1, skip2}, convolutions \cite{conv16, Fukushima1980NeocognitronAS, rumelhart1985learning, lecun1998gradient, lecun1999object} or graph convolutions \cite{bruna2013spectral, henaff2015deep,Duvenaud2015MolFingerPrintConvolutionNetwork, li2015gated, defferrard2016convolutional, kipf2016semi}, pooling \cite{lecun1998gradient, lecun1999object}, batch \cite{ioffe2015batch} or layer \cite{ba2016layer} normalization, and / or attention \cite{bahdanau2014neural, vaswani2017attention}. Furthermore, though these results apply to randomly initialized neural
networks, appropriately trained networks are also drawn from GPs \cite{Lee2019WideNN,Jacot2018NeuralTK}. NGPs have been used to model finite neural networks in \cite{antognini2019finite, naveh2020predicting, Yaida2019NonGaussianPA, cohen2020learning}, with some key differences from our work. For these reasons, we believe that an EFT approach to 
neural networks is possible under a wide variety of circumstances.

While working on this project two papers appeared that also utilize Feynman
diagrams in neural networks, and we would therefore like to differentiate our work. In \cite{Dyer2020AsymptoticsOW} diagrams were associated to a different class of Gaussian integrals associated with neural network parameters drawn from a Gaussian distribution; they were used to put bounds on correlation functions in infinite width neural networks. In contrast, our Gaussian (and non-Gaussian) integrals are over the function space associated with the GP or NGP, crucially relying on the central limit theorem.  \cite{Yaida2019NonGaussianPA} focuses on preactivation distribution flow associated with perturbative corrections in the width parameter.
Both of these interesting papers differ from our approach, which uses Feynman diagrams in our application of Wilsonian effective field theory to determine NGP likelihoods. It is applicable to any family of architectures admitting a GP limit, and knowledge of the GP limit and EFT together allow for the prediction of NGP correlation functions. Couplings are extracted from experiments and matched to theory predictions for their flow under the renormalization group.

\textbf{Our contributions.} In this work we alternate between a general development of the theory
behind a NN-QFT correspondence and its experimental verification\footnote{We provide an implementation of our code at \url{https://github.com/keeganstoner/nn-qft}.}. We do so in
the simplest class of architectures admitting such a description,
single-layer fully-connected networks. By construction, the point 
is precisely to \emph{not} take an infinite width limit $N\to \infty$,
instead treating the neural networks as perturbations of the associated
GP, modeled by perturbing the GP to an NGP by adding corrections according to the rules of Wilsonian EFT. A summary of our results:
\begin{itemize}
\item \textbf{Higher-point GP statistics.} We review how $n$-pt correlation
functions in GPs, $n>2$, may be computed using non-interacting Feynman diagrams and demonstrate the experimental falloff to GP predictions as $N\to \infty$ in
three classes of single-layer networks.
\item \textbf{Neural Net and NGP Likelihoods from Wilsonian EFT.} We develop rules for treating the NGPs associated with finite $N$ networks. Non-Gaussian terms correspond to particle interactions and have coefficients (``couplings'') corresponding to interaction strengths. We compute the $4$-pt and $6$-pt functions using interacting Feynman diagrams. In the single-layer networks, $4$-pt couplings are extracted from experiments. These couplings are used to make predictions for the $6$-pt function \amnew{and $4$-pt function evaluated on test set inputs}, which we verify experimentally.
\item \textbf{Couplings and GP Limit Symmetries.} Whether the coefficient of a non-Gaussian term in the NGP likelihood is a constant or a function is of practical importance. We demonstrate that this may be understood in terms of 't Hooft's notion (``technical naturalness'') that the couplings may be small when setting them to zero recovers a symmetry. The coefficient is nearly a constant when the GP limit is translation invariant, i.e., if setting its variance to zero recovers a symmetry.
\item \textbf{Minimal NGP Likelihoods and Renormalization Group (RG) Flow.} Our approach introduces a parameter $\Lambda$ known as the cutoff, so technically there is an infinite class of NGPs designed to describe one set of experiments. If all are equally effective, the predictions must not depend on $\Lambda$, which yields a differential equation known as a $\beta$-function governing the couplings at different values of the cutoff. The induced flow in coupling space is known as Wilsonian RG flow, and the couplings are said to ``run''. We compute the $\beta$-functions for $4$-pt couplings in some of our examples and experimentally verify their running. As a limit in $\Lambda$ is taken, the flow renders some coefficients negligible, yielding a minimal NGP likelihood.
\end{itemize}
These elements, particularly EFT and RG flow, constitute the essentials of an NN-QFT correspondence, as they constitute the essentials of a modern approach to QFT.

\emph{For our ML readers:} this approach immediately connects overparameterization with neural net likelihood simplicity. That is, neural networks with increasingly large numbers of parameters are drawn from increasingly simple distributions. This was already implicit in the connection between asymptotic neural networks and GPs, and also in the fact that non-Gaussian corrections to GPs should be $1/N$ suppressed. However, the Wilsonian renormalization group adds another layer to the story: at any fixed $N$ there is a family of NGPs indexed by a continuous parameter $\Lambda$, and by taking limits in $\Lambda$ many of the non-Gaussian terms in the NGP likelihood become negligible, leading to simpler NGP likelihoods. \amnew{This physics-motivated technique introduces a} gained simplicity, \amnew{verified in the results of our NN experiments}: to excellent approximation, a \emph{single number} is all that it takes to correct GP correlation functions to NGP (neural network output) correlation functions, despite the fact that in moving away from the GP limit an infinite number of neural network
parameters are lost. \amnew{For other architectures it may be that additional parameters that are the coefficients of non-Gaussian terms in the probability distribution function are needed; for our architectures we also study three-parameter models, but find that one suffices to model the correlation functions to excellent approximation in experiments.}

We also recognize that our experimental focus on randomly initialized networks, rather than trained networks, means that we have left an understanding of learning to future work. This is intentional, as our primary goal in this work is to develop EFT techniques for treating the NGPs associated with neural networks. We emphasize that though we have not treated trained networks directly, our techniques apply \amnew{to trained networks that are close to GPs. Shown in \cite{yangTP1,yangTP2, yangTP3}, randomly initialized infinite width NNs are GPs, and further, the GP property persists under appropriate training \cite{yangTP1,yangTP3,yang2020feature,Lee2019WideNN,Jacot2018NeuralTK}. Such trained networks at finite width should be effectively described using our techniques, which only requires being close to a GP.}

\emph{For our physicist readers:} in applying EFT techniques to the non-Gaussian processes from which neural networks are drawn, there are a number of important changes from the usual cases in QFT. Some of them include:
\begin{itemize}
\item \textbf{Tree-level divergences.} In the absence of $\delta$-functions that collapse integrals associated with internal points, divergences may arise in tree-level diagrams. Of course, proper regularization and renormalization renders them a feature, not a bug.
\item \textbf{Lack of derivatives.} Gaussian process kernels are functions on input space, but \jimnew{their inverses, which appear in the GP probability distribution function, in general need not have derivatives (though sometimes  cases they could). If not}, they are akin to space-dependent mass terms, rather than kinetic terms, \jimnew{which are sometimes considered in cosmology \cite{Khoury:2003rn}.}  We plan to include derivatives in future work.
\item \textbf{Coupling functions.} Most notably, couplings in NGPs are not necessarily constants, and their spatial (input) variance can be understood in terms of technical naturalness. 
\end{itemize}
Physicist readers may also notice that we have followed a school of thought in QFT due to Coleman, which implores the student to first understand the basics of 
perturbation theory and renormalization in the case of spin-$0$ particles, since the introduction of higher spin particles does not change the basic conceptual framework. Coleman's approach is even more natural in a NN-QFT correspondence, since the scalar outputs associated with neural networks mean that (as functions) they should be understood as scalar fields, which correspond to spin-$0$ particles. Of course this could change with further developments in neural networks, but it is appropriate for now. Another element of Coleman's school is that perturbation theory in QFT is ``just Gaussian integrals'' (with some additional decorations), which we review in detail in Appendix \ref{app:gaussian}.

\medskip
This paper is organized is follows. In Section \ref{sec:FreeGP} we review 
Gaussian processes and explain how they correspond to free field theories. We demonstrate the higher-point correlation functions may be computed from two-point statistics using non-interacting Feynman diagrams, and experimentally demonstrate the falloff to GP predictions in the $N\to \infty$ limit. In Section \ref{sec:EFT} we introduce a treatment of the NGPs associated with finite $N$ networks in terms of Wilsonian EFT, including the introduction of interacting Feynman diagrams for the computation of correlation functions. We extract $4$-pt couplings and use them to make $6$-pt predictions, which are verified experimentally. Technical naturalness is discussed in regards to when and why couplings should be constants versus functions. In Section \ref{sec:rg} we introduce Wilsonian RG as applied to neural networks, computing some $\beta$-functions that govern the running of couplings and verifying them experimentally. We argue that $6$-pt couplings are negligible due to being irrelevant, in the Wilsonian sense.

\section{Asymptotic Neural Networks, Gaussian Processes, and Free Field Theory\label{sec:FreeGP}}

In this Section we wish to draw a sharp analogy between Gaussian
processes (GPs), neural networks, and techniques from free field theory.
In this analogy we will facilitate computations of correlation functions
in the GP limit with Feynman diagrams, representing correlation functions 
in terms of the kernel $K(x,x')$ that determines the GP. We will then specify to a
concrete class of neural networks --- single layer feedforward networks
in their infinite width limit --- that exemplify the general idea, demonstrating
that theoretical calculations with Feynman diagrams in the GP and associated
combinatorics agree with 
experiments, up through the 6-pt functions.

The essential idea connecting Gaussian processes and free field theory is
simple to state. Some classes of neural network architectures admit a limit $N\to \infty$
where a randomly initialized neural network in that class is equivalent
to a draw from a Gaussian process, i.e. the neural network outputs evaluated
on fixed inputs are
described by draws from a multivariate Gaussian distribution. Meanwhile,
a field configuration
$f(x)$ in a free field theory is also drawn from a multivariate Gaussian
distribution, and it is precisely the Gaussian nature of the associated
path integral that makes the theory solvable, i.e. all correlation functions
of the fields $f(x)$ may be computed exactly. The free field theory is a GP, with the
kernel describing the dynamical propagation of field quanta.

Some aspects of this Section are stated in the literature, see
e.g. \cite{antognini2019finite, naveh2020predicting, Yaida2019NonGaussianPA}, but we wish to fully develop the formalism in order
to facilitate new results obtained in later Sections. The essentials of the GP-Free QFT correspondence
are presented in Table \ref{tab:GPFREEQFT}.

\subsection{General Theory}

\begin{table}[t]
    \centering 
    \begin{tabular}{|c|c|}
        \hline
    GP / asymptotic NN& Free QFT \\ \hline
    input $x$ & external space or \jimnew{momentum space point}\\
    kernel $K(x_1,x_2)$ & Feynman propagator \\ 
    asymptotic NN $f(x)$ & free field \\ 
    log-likelihood & free action $\sgp$ \\ \hline
    \end{tabular}
    \caption{Correspondence between quantities in the GP / asymptotic neural network and free QFT.}
    \label{tab:GPFREEQFT}
    \end{table}

Let us develop the connection between neural networks, Gaussian processes,
and free field theory. For simplicity we assume that 
the mean $\mu$ of the GP is zero, corresponding to zero vacuum expectation
value (VEV) in the field theory, an assumption which we will relax in subsequent
work.

Consider a family of neural network architectures with learnable parameters
$\theta$ and a discrete hyperparameter $N$,
\begin{equation}
f_{\theta,N}: \bR^{\din} \to \bR^{\dout},
\end{equation}
where at initialization the learnable parameters are drawn as $\theta \sim P(\theta)$.
The parameter distribution $P(\theta)$ and the network architecture together 
induce an implicit distribution on function space from which the neural network is drawn,
$P(f)$. We will often drop the explicit subscripts $\theta, N$ for brevity.

For many architectures there is a limit $N\to \infty$ in which
the distribution on functions becomes a Gaussian process, which means that the 
neural network outputs $\{f(x_1),\dots, f(x_k)\}$ evaluated on
any fixed set of $k$ inputs $\{x_1,\dots,x_k\}$ are drawn from a multivariate Gaussian
distribution $\cN(\mu, \Xi^{-1})$,
\begin{equation}
  \{f(x_1),\dots, f(x_k)\} \sim \cN(\mu, \Xi^{-1}),
\end{equation}
which by assumption in this paper has 
$\mu=0$. The inverse covariance matrix $\Xi$ is determined by the kernel
function $K(x,x')$ as $(\Xi^{-1})_{ij} = K_{ij} := K(x_i,x_j)$. Since $\mu=0$,
the GP is entirely determined by its covariance, which in turn is entirely
determined by the kernel. 
Correlation functions between $n$ outputs can be expressed as
\begin{equation}
  G^{(n)}({x_1},\dots, {x_n}) = \frac{\int df \,\, f_{1}\dots f_{n}\, e^{-\frac12 f_i \Xi_{ij} f_j}}{Z},
\end{equation}
and are called $n$-pt functions,
where the partition function $Z= \int df\,\, e^{-S}$, and $S=-\frac12 f_i \Xi_{ij} f_j$
is the log-likelihood, or "action" in physics language. Einstein summation is implicit, and $f_i:=f(x_i)$
is a vector of outputs on a fixed set of inputs $\{x_i\}$ with dimension $\din = d$. 

Of course, the Gaussian process is defined for any $\{x_1,\dots,x_k\}$
for any $k$, and therefore it is natural to take the continuum limit, 
in which case the correlation functions become 
\begin{equation}
  G^{(n)}({x_1},\dots, {x_n}) = \frac{\int df \,\, f(x_{1})\dots f(x_{n})\, e^{- S}}{Z},
\end{equation}
where the log-likelihood is now
\begin{equation}
	\label{eqn:gp_nonlocal_S}
S = \frac12 \int d^{d_\text{in}}x \, d^{d_\text{in}}x'\,\,  f(x) \Xi(x,x') f(x') 
\end{equation}
and $\Xi(x,x')=K^{-1}(x,x')$ is the inverse covariance function, defined by
\begin{equation}
	\label{eq:functionalinverseK}
\int d^{d_\text{in}} x' \, K(x,x') \, \Xi(x',x'') = \delta^{(d_\text{in})}(x-x''),
\end{equation}
where $\delta^{(d_\text{in})}(x-x'')$ is the $d_\text{in}$-dimensional Dirac delta function.
This equation is the continuum analog of the relation $(\Xi^{-1})_{ij} = K_{ij}$ 
in the discrete case.

\jimnew{We note important special cases, some of which arise below, that will also be the subject of follow-up work. Suppose in \eqref{eqn:gp_nonlocal_S} we had $\Xi(x,x') = \Xi(x) L_\sigma(x,x')$ where $L_\sigma(x,x')$ is any family of functions that limits to $L_0(x,x')=\delta^{(\din)}(x-x')$; for instance, a family of normalized Gaussians of variance $\sigma^2$. In that limit, the double-integral collapses and one is left with
\begin{equation}
	S = \frac12 \int d^{d_\text{in}}x \,\, f(x) \Xi(x) f(x),
\end{equation}
which defines a \emph{local} Gaussian process, denoting that the associated GP log-likelihood only has one integral over the input space. In this case the associated inversion formula is instead $\Xi(x)K(x,x')=\delta^{(\din)}(x-x')$. If a kernel satisfies this identity, it means it is the kernel of a local GP.  At finite $\sigma$ it can be thought of as a scale of non-locality, and one might call such a GP \emph{semi-local} and try to understand the utility of the scale $\sigma$ in neural networks. Similarly, one could consider a log-likelihood with $f$ constant and no input integrals $S=\frac12 \, f \,\Xi\, f$, which one might call an \emph{ultra-local} GP, with associated inversion formula $K \Xi = 1$.
}

\vspace{1cm}
Both the discrete and continuum  versions of the GP
$n$-pt functions may be computed exactly
using standard Gaussian integral techniques reviewed in Appendix \ref{app:gaussian}. This
would not be the case if the action contained beyond-quadratic
terms, though a perturbative expansion may be available for suitably 
small coefficients of beyond-quadratic terms; see Section \ref{sec:EFT}.

A direct physics analog of a Gaussian process is a free field theory,
which describes a quantum field $\phi(x)$ without any interaction terms. 
The field $\phi(x)$ depends on the $d$-dimensional coordinates of space\footnote{Spacetime,
in the Lorentzian case. We will focus on the Euclidean case throughout. \jimnew{Depending on the structure of the log-probability, momentum space might be a better analog of the input.}} $x$
and the associated field theory is defined by the path integral 
\begin{equation}
Z = \int D\phi\,  e^{-S[\phi]}
\end{equation}
in terms of an action $S[\phi]$, which is quadratic in the case of
a free field theory. A famous example is free scalar field theory,
which has 
\begin{equation}
S[\phi] = \int  d^dx \, \phi(x) (\Box+m^2)\phi(x),
\end{equation}
with $\Box := \partial_\mu \partial^\mu$ and $m$ the 
mass of the bosonic particle associated to $\phi$. 
The functional inverse of $(\Box + m^2)$ is known as the propagator,
the $2$-pt correlation function in the free field theory, and 
is the analog of the GP kernel. By virtue of being Gaussian,
correlation functions in the free theory may be computed exactly.

\amnew{Following Table \ref{tab:GPFREEQFT}, an input $x$ to the NN corresponds to a point in space or momentum space $\mathbb{R}^{\din}$ in QFT. Computed in natural units $\hbar = c = 1$ of high energy physics, the neural network kernel, log-likelihood, and output $f(x)$ are the analogs of the free field theory propagator, free action and free field $\phi(x)$, respectively.}

\vspace{1cm}
A central point of this work is that taking an infinite-width neural network to a large-but-finite width
neural network corresponds to moving away from the GP to a non-Gaussian
process (NGP),
which in field theory corresponds to turning on interactions. We will
do this in Section \ref{sec:EFT}.

\subsection{Neural Network Correlation Functions with Feynman Diagrams} \label{correlation_fn_section}

Anticipating their utility when we move away from the GP, in this Section
we derive Feynman rules for the diagrammatic computation of correlation
functions in the GP. The ability to represent such computations diagrammatically,
in both quantum field theories and Gaussian processes, follows directly from basic properties of Gaussian
integrals reviewed in Appendix \ref{app:gaussian}.

The partition function of the Gaussian process is
\be 
\zgp[J]=\frac{\int df\, e^{-\sgp-\frac12\int d^{d_\text{in}}x\,  J(x)f(x) -\frac12\int  d^{d_\text{in}}y \, J(y)f(y)}}{\zgpzero}, 
\ee
where we have included the terms involving the source $J(x)$, and 
$\zgpzero := \int df e^{-\sgp}$ is the associated action, or (negative) log-likelihood, is
\be
\sgp = \frac12 \int d^{d_\text{in}} x \, d^{d_\text{in}} y \,\,  f(x) \Xi(x,y) f(y).
\ee
The $n$-pt correlation functions are defined by
\begin{equation}
  \ggp^{(n)}(x_{1},\dots, x_{n}) = \frac{\int df \,\, f(x_{1})\dots f(x_{n})\, e^{-\sgp}}{\zgpzero},
\end{equation}
We will consistently label GP quantities with a subscript, since in moving away from the
GP limit in neural network architectures 
we will use effective field theory to determine deviations from the GP quantities using the NGP effective actions $S=\sgp + \Delta S$.

Since all of the $f$-dependent terms in $\zgp$ are quadratic or below, it can be 
evaluated exactly by completing the square and performing the Gaussian integral, yielding
\be  
\zgp[J]= \exp{\left(\frac12\int d^{d_\text{in}}x \, d^{d_\text{in}}y\, J(x) K(x,y) J(y) \right) } \label{eqn:GPint1}, 
\ee
where the $\zgpzero$ factor has canceled. The correlation functions may be written as
\be
\label{eqn:ggpddj}  \ggp^{(n)}(x_1, \dots, x_n) 
  = \left[ \left(-\frac{\delta}{\delta J(x_1)}\right)\dots \left(-\frac{\delta}{\delta J(x_n)}\right) \zgp\right] \bigg|_{J=0}. \,
\ee
The basic pattern of the computation of \eqref{eqn:ggpddj} emerges from the fact that 
taking these functional $J$-derivatives either pulls down factors from the exponential or 
hits previously-pulled-down $J$-factors, using $\delta J(x)/\delta J(y) = \delta^{d_\text{in}}(x-y)$ repeatedly\footnote{Details can be found in Appendix \ref{app:gaussian}.}.
The $\delta$-functions make the kernels depend on external points, and depending on $n$ there
will be many terms, each with $n/2$ kernel factors. Those terms contain
the information of which external points appear in kernels together, motivating
the definition of the set of ways to connect the points $(x_1,\dots,x_n)$ in pairs,
\be
\label{eqn:ggp214}
\wick(x_1,\dots,x_n) = \{P \in \text{Partitions}(x_1,\dots,x_n)\,\, | \,\, |p| = 2 \,\, \forall p \in P\}.
\ee
This is known as the set of Wick contractions, and it has cardinality 
$|\wick(x_1,\dots,x_n)| = (n-1)!!$. With this definition,
the procedure of computing $\ggp^{(n)}(x_1,\dots,x_n)$ via $J$-derivatives yields
\be
\label{eqn:ggpwick}
  \ggp^{(n)}(x_1, \dots, x_n) =  
  \sum_{p \in \wick(x_1,\dots,x_n)} K(a_1,b_1)\dots K(a_{n/2},b_{n/2}) 
\ee
where we write each element
$p \in \wick(x_1,\dots,x_n)$ as $p={(a_1,b_1),\dots,(a_{n/2},b_{n/2})}$. 
This simple expression may be read
\begin{center}
  ``sum over all ways of pairing up elements in $\{x_1,\dots,x_n\}$,\\
  and in each term write a kernel factor $K(a_i,b_i)$ for each of the pairs $(a_i,b_i)$,''
\end{center}
giving a simple rule for writing down the answer for $\ggp^{(n)}(x_1, \dots, x_n)$,
for any $n$, even when the combinatorics of $\wick(x_1,\dots,x_n)$ become grotesque.

\vspace{1cm}

Diagrammatic representations of the GP follow immediately by a simple rule change:
\begin{center}
  ``sum over all ways of connecting the points $\{x_1,\dots,x_n\}$ in pairs,\\
  and in each term draw a line between the points in the pair $(a_i,b_i)$.''
\end{center}
Both of these simple colloquial expressions yield  correct ways to represent \eqref{eqn:ggpwick},
one in terms of a sum of kernel factors, and the other in terms of a sum of diagrams.
The diagram-to-analytic map between the two is clearly that for each line between $a_i$ and $b_i$ in a diagram\footnote{These are, of course, the Feynman diagrams of physics, and the rules are known as 
Feynman rules.},
write a kernel factor $K(a_i,b_i)$.  For instance, in the case $n=2$, $\wick(x_1,x_2) = \{\{(x_1,x_2)\}\}$ and we have
\bea
\ggp^{(2)}(x_1,x_2) &=& K(x_1,x_2)\nonumber \\ 
&=& 
\begin{tikzpicture}[line width=1pt, scale=0.5, baseline=0*\fontsizeshift]
		\coordinate (x1) at (-1,0);
	\coordinate (x3) at (1,0);
	\coordinate (ls1) at (0,.3);
	\draw[scs] (x1) -- (x3);
	\draw[fill] (x1) circle (.05);
	\draw[fill] (x3) circle (.05);
	\node at ($(x1) + 2*(ls1)$) {$x_1$};
	\node at ($(x3) + 2*(ls1)$) {$x_2$};
\end{tikzpicture}.
\eea
Similarly, 
$\wick(x_1,x_2,x_3,x_4) = \{\{(x_1,x_2),(x_3,x_4)\},\{(x_1,x_3),(x_2,x_4)\},\{(x_1,x_4),(x_2,x_3)\}\}$ and we have
\bea \label{4-ptwick}
  \ggp^{(4)}(x_1,x_2,x_3,x_4) &=& K(x_1,x_2)K(x_3,x_4) + K(x_1,x_3)K(x_2,x_4) + K(x_1,x_4)K(x_2,x_3) \nonumber \\
   &=& \begin{tikzpicture}[line width=1pt, scale=0.5, baseline=-.5*\fontsizeshift]
		\coordinate (C) at (0,0);
	\coordinate (x1) at (-1,1);
	\coordinate (x2) at (-1,-1);
	\coordinate (x3) at (1,1);
	\coordinate (x4) at (1,-1);
	\coordinate (ls) at (0,-.3);
	\coordinate (ls1) at (0,.3);
	\draw[scs] (x1) -- (x2);
	\draw[scs] (x3) -- (x4);
	\draw[fill] (x1) circle (.05);
	\draw[fill] (x2) circle (.05);
	\draw[fill] (x3) circle (.05);
	\draw[fill] (x4) circle (.05);
		\node at ($(x1) + 2*(ls1)$) {$x_1$};
	\node at ($(x2) + 2*(ls)$) {$x_2$};
	\node at ($(x3) + 2*(ls1)$) {$x_3$};
	\node at ($(x4) + 2*(ls)$) {$x_4$};
\end{tikzpicture} \, \, + \, \,  
\begin{tikzpicture}[line width=1pt, scale=0.5, baseline=-.5*\fontsizeshift]
		\coordinate (C) at (0,0);
	\coordinate (x1) at (-1,1);
	\coordinate (x2) at (-1,-1);
	\coordinate (x3) at (1,1);
	\coordinate (x4) at (1,-1);
	\coordinate (ls) at (0,-.3);
	\coordinate (ls1) at (0,.3);
	\draw[scs] (x1) -- (x3);
	\draw[scs] (x2) -- (x4);
	\draw[fill] (x1) circle (.05);
	\draw[fill] (x2) circle (.05);
	\draw[fill] (x3) circle (.05);
	\draw[fill] (x4) circle (.05);
		\node at ($(x1) + 2*(ls1)$) {$x_1$};
	\node at ($(x2) + 2*(ls)$) {$x_2$};
	\node at ($(x3) + 2*(ls1)$) {$x_3$};
	\node at ($(x4) + 2*(ls)$) {$x_4$};
\end{tikzpicture} \, \,  +
\begin{tikzpicture}[line width=1pt,scale=0.5,baseline=-.5*\fontsizeshift]
		\coordinate (C1) at (-.2,-.2);
	\coordinate (C2) at (.2,.2);
	\coordinate (x1) at (-1,1);
	\coordinate (x2) at (-1,-1);
	\coordinate (x3) at (1,1);
	\coordinate (x4) at (1,-1);
	\coordinate (ls) at (0,-.3);
	\coordinate (ls1) at (0,.3);
	\draw[scs] (x1) -- (x4);
	\draw[scs] (x2) -- (C1);
	\draw[scs] (C2) -- (x3);
	\draw[fill] (x1) circle (.05);
	\draw[fill] (x2) circle (.05);
	\draw[fill] (x3) circle (.05);
	\draw[fill] (x4) circle (.05);
		\node at ($(x1) + 2*(ls1)$) {$x_1$};
	\node at ($(x2) + 2*(ls)$) {$x_2$};
	\node at ($(x3) + 2*(ls1)$) {$x_3$};
	\node at ($(x4) + 2*(ls)$) {$x_4$};
\end{tikzpicture}
\eea
In connecting points in pairs for any odd $n$, there is always a leftover point, which corresponds to a factor
of $J$ in every term in the analytic expression. Since $J$ is set to zero after taking functional $J$-derivatives
in \eqref{eqn:ggpddj}, we conclude that $\ggp^{(n)}(x_1,\dots,x_n)=0$ for any odd $n$.

In the analogy to free QFT, quantities in the GP map to associated quantities in the QFT, summarized in Table \ref{tab:GPFREEQFT}.
Remembering that the GP can sometimes be realized by asymptotic neural networks, the neural network inputs are the points 
in space in the QFT, and the kernel is the Feynman propagator, which represents the probability or amplitude of propagation of a 
particle from one point to another. Notably, due to the Gaussian nature of $\zgp$, all diagrams in the diagrammatic
expressions for $\ggp^{(n)}(x_1,\dots,x_n)$ are simple connections of pairs of points in space, flying past one another
without interacting. When $\zgp$ is corrected by non-Gaussian terms,
``interactions'' arise in a way that we will make concrete in Section \ref{sec:EFT}. Thus, the GP / asymptotic neural networks correspond to
free (non-interacting) field theories, and moving away from the asymptotic limit corresponds to turning on interactions.

\subsection{Examples: Infinite Width Single-Layer Networks} \label{sec:networksetup}

\bigskip

To experimentally realize the theoretical ideas of this paper, 
such as utilizing effective
field theory and Wilsonian renormalization group flow for understanding neural 
networks, we must introduce concrete architectures that will be used in
experiments and their corresponding GPs. Specifically, we now introduce the 
three single-layer architectures we study in this paper and also review
the correspondence between infinite width single-layer networks and Gaussian processes.

Consider a fully-connected neural network with one hidden layer and elementwise nonlinearity $\sigma$, defined by $f(x) = W_{1} (\sigma(W_{0} x + b_{0})) + b_{1}$, where the weights and biases, $W_{i}$ and $b_{i}$, characterize the affine transformations for each layer. Including the spaces associated with the hidden layers, 
\begin{equation}
\label{structure}
  f_{\theta,N}: \bR^{\din} \xrightarrow[]{W_{0}, \, b_{0}} \bR^{N} \xrightarrow[]{\sigma} \bR^{N} \xrightarrow[]{W_{1}, \, b_{1}} \bR^{d_\text{out}}.
\end{equation}
The weight and bias parameters, collectively labeled as $\theta$, are i.i.d. and drawn from a Gaussian distribution. Specifically, the biases are drawn from $\mathcal{N}(\mu_{b}, \sigma_{b}^{2})$ and the weights in each layer $W_{0}$, $W_{1}$ are drawn from $\mathcal{N}(\mu_{W}, \sigma_{W}^{2}/\din)$ and $\mathcal{N}(\mu_{W}, \sigma_{W}^{2}/N)$, respectively, so they are normalized with respect to the input dimension of the associated layer. The first linear layer takes the input $x$ to a preactivation $z_{0}$
\begin{equation}
\label{eq:preact}
  z^{j}_{0} = \sum_{i = 1}^{\din} W^{ij}_{0} x^{i} + b^{j}_{0},
\end{equation}
which is then acted on by the elementwise nonlinearity $\sigma$, giving a postactivation $x^{j}_{1} = \sigma(z^{j}_{0})$, that is acted on by the final linear layer, yielding
\begin{equation} \label{last_layer}
  f_{\theta,N}(x) = z^{k}_{1} = \sum_{j = 1}^{N} W^{jk}_{1} x^{j}_{1} + b^{k}_{1},
\end{equation}
the output of the neural network.

By the Central Limit Theorem (CLT), in the infinite width limit the network outputs are drawn from a Gaussian distribution on function space \cite{neal}; i.e. the network outputs are drawn from a GP. This arises because the weight part of output layer of the network defined in \eqref{last_layer} is the sum of $N$ i.i.d. terms. In the infinite width limit $N \rightarrow \infty$, we get a finite\footnote{Since the elements of $W$ are properly normalized.} sum over these independent parameters. Thus by the CLT we have a neural network output that is selected from a Gaussian distribution, i.e., the neural network evaluated on any finite collection of inputs is drawn from a multivariate Gaussian distribution. This is precisely what defines a GP. \amnew{We present in Figure \ref{NNschematics} a schematic of a single-layer feedforward neural network, where the arrows represent the linear layers of weights and biases, and the nonlinear activation function is applied elementwise to each node in the hidden layer.  Since the outputs are real scalars, the closest QFT analog of the infinite $N$ neural network is a free scalar theory, albeit with a different two-point function (kernel).}
\amnew{

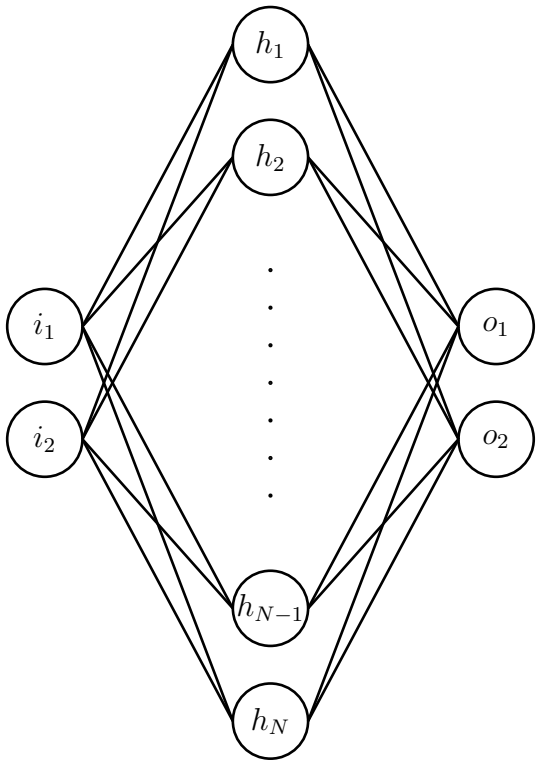
\begin{figure}[] 
  \centering
  \begin{tikzpicture}[line width=1pt, scale=0.5, baseline=-.5*\fontsizeshift]
		\coordinate (C) at (0,0);
	\coordinate (i1) at (-6,1.5);
	\coordinate (i2) at (-6,-1.5);
	\coordinate (ri1) at (-5,1.5);
	\coordinate (ri2) at (-5,-1.5);
	\coordinate (li1) at (-7,1.5);
	\coordinate (li2) at (-7,-1.5);
	\coordinate (o1) at (6,1.5);
	\coordinate (o2) at (6,-1.5);
	\coordinate (lo1) at (5,1.5);
	\coordinate (lo2) at (5,-1.5);
	\coordinate (ro1) at (7,1.5);
	\coordinate (ro2) at (7,-1.5);

	\coordinate (lh1) at (-1,9);
	\coordinate (lh2) at (-1,6);
	\coordinate (lh3) at (-1,-6);
	\coordinate (lh4) at (-1,-9);
	\coordinate (h1) at (0,9);
	\coordinate (h2) at (0,6);
	\coordinate (h3) at (0,-6);
	\coordinate (h4) at (0,-9);
	\coordinate (rh1) at (1,9);
	\coordinate (rh2) at (1,6);
	\coordinate (rh3) at (1,-6);
	\coordinate (rh4) at (1,-9);

	\draw (h1) circle (1);
	\draw (h2) circle (1);
	\draw (h3) circle (1);
	\draw (h4) circle (1);

	\draw (i1) circle (1);
	\draw (i2) circle (1);
	\draw (o1) circle (1);
	\draw (o2) circle (1);

	\draw[scs] (ri1) -- (lh1);
	\draw[scs] (ri1) -- (lh2);
	\draw[scs] (ri1) -- (lh3);
	\draw[scs] (ri1) -- (lh4);
	\draw[scs] (ri2) -- (lh1);
	\draw[scs] (ri2) -- (lh2);
	\draw[scs] (ri2) -- (lh3);
	\draw[scs] (ri2) -- (lh4);

	\draw[scs] (lo1) -- (rh1);
	\draw[scs] (lo1) -- (rh2);
	\draw[scs] (lo1) -- (rh3);
	\draw[scs] (lo1) -- (rh4);
	\draw[scs] (lo2) -- (rh1);
	\draw[scs] (lo2) -- (rh2);
	\draw[scs] (lo2) -- (rh3);
	\draw[scs] (lo2) -- (rh4);

	\draw[fill] (0,3) circle (.02);
	\draw[fill] (0,2) circle (.02);
	\draw[fill] (0,1) circle (.02);
	\draw[fill] (0,0) circle (.02);
	\draw[fill] (0,-3) circle (.02);
	\draw[fill] (0,-2) circle (.02);
	\draw[fill] (0,-1) circle (.02);

	\coordinate (x1) at (-1,1);
	\coordinate (x2) at (-1,-1);
	\coordinate (x3) at (1,1);
	\coordinate (x4) at (1,-1);
	\coordinate (ls) at (0,-.3);
	\coordinate (ls1) at (0,.3);
	
	\node at (i1) {$i_1$};
	\node at (i2) {$i_2$};
	\node at (o1) {$o_1$};
	\node at (o2) {$o_2$};
	\node at (h1) {$h_1$};
	\node at (h2) {$h_2$};
	\node at (h3) {$h_{N-1}$};
	\node at (h4) {$h_N$};

				\end{tikzpicture} \, \, 
\caption{Single-layer fully-connected network of width $N$, with $\din=\dout=2$.}
\label{NNschematics}
 \end{figure}

 Any neural network $f$ with a linear output layer, such as the single-layer networks in our examples, has a structure that affects its study: it is generally of the form
 \begin{equation}
 f = f_b + f_W,
 \end{equation}
 where $f$ are drawn from independent processes\footnote{For our initializations they are mean-free processes, which is assumed in the derivations but can be relaxed.}. Due to independence, the $2$-pt function associated to $f$ is the sum of the $2$-pt functions associated to $f_b$ and $f_W$. Furthermore, the higher-point correlators associated to $f$ decompose into the higher-point correlators associated to $f_b$ and $f_W$ in the natural way.
 For instance, just as 
 \begin{equation}
G^{(2)}(x_1,x_2) = G^{(2)}_b(x_1,x_2) + G^{(2)}_W(x_1,x_2),
 \end{equation}
 where subscript denotes the correlation functions associated to $f_b$ and $f_W$, we also have 
 \begin{equation}
	G^{(4)}(x_1,x_2,x_3,x_4) = G^{(4)}_b(x_1,x_2,x_3,x_4) + G^{(4)}_W(x_1,x_2,x_3,x_4) + \sum_{i\neq j \neq k \neq l} G^{(2)}_b(x_i,x_j) G^{(2)}_W(x_k,x_l),
\end{equation}
and similarly for the higher-point functions. In networks with a linear output layer with Gaussian biases, $f_b$ is a bias term that is drawn from an ultra-local Gaussian process at all $N$ (since it doesn't depend on $N$), and $f_W$ is the weight-dependent term that becomes a GP as $N\to \infty$, where the type of GP (non-local, semi-local, local) is model-dependent and depends on the choice of activation function.
 }

One must derive the kernel associated to
a given infinite width architecture in order to 
compute correlation functions in the associated GP. Kernel derivations are done in detail in Appendix \ref{kernel_derivation_appendix} via two methods: the first method described in \cite{williams}
computes the exact $2$-pt function for all widths, and a second method in
\cite{yang} applies only in the GP limit. In the architectures that we study,
the methods happen to agree for all widths $N$, though in general it is only required that the former method recover the
result of the latter in the infinite width limit. We also build upon the work of \cite{yang} in creating a network with a translation invariant Gaussian kernel.
 
The networks differ only in their activation functions
$\sigma$. We now introduce the networks that we study via their
activation functions and 
the associated GP kernels.

\subsubsection*{Erf-net}

The network architecture that we call Erf-net is defined by an error function activation
\begin{equation}
  \sigma(z) = \text{erf}(z) = \frac{2}{\sqrt{\pi}} \int_{0}^{z} dt \, e^{-t^{2}}.
\end{equation}
The associated GP kernel is
\begin{eqnarray}
  K_{\text{Erf}}(x, x') = \sigma_{b}^2 + \sigma_{W}^2\, \frac{2}{\pi} \arcsin \Bigg[ \frac{  2(\sigma_{b}^2 + \frac{\sigma_{W}^{2}}{\din}\, x x^\prime ) }{\sqrt{\left( 1 +  2(\sigma_{b}^2 + \frac{\sigma_{W}^{2}}{\din}\, x^{2} )\right) \left( 1 + 2(\sigma_{b}^2 + \frac{\sigma_{W}^{2}}{\din}\,x'^{2}) \right)  }} \Bigg] \label{erfkernel},
\end{eqnarray}   
which is derived via both methods of \cite{yang} and \cite{williams} in \eqref{innerkernel}.

\subsubsection*{ReLU-net}

The network architecture that we call ReLU-net is defined by ReLU activation function:
\begin{equation}
  \sigma(z) =
    \begin{cases} 
    0 & z < 0 \\
    z & z \geq 0 \\
 \end{cases}
\end{equation}
The associated GP kernel is
\begin{eqnarray}
 \label{relukernel}
  K_{\text{ReLU}}(x, x') &=& \sigma_{b}^2 + \sigma_{W}^2\, \frac{1}{2\pi} \sqrt{(\sigma_{b}^2 + \frac{\sigma_{W}^{2}}{\din}\, x\cdot x)(\sigma_{b}^2 + \frac{\sigma_{W}^{2}}{\din}\, x^\prime\cdot x^\prime)}(\sin \theta + (\pi - \theta)\cos\theta ), \\
  \theta &=& \arccos \Bigg[\frac{\sigma_{b}^2 + \frac{\sigma_{W}^{2}}{\din}\, x\cdot x^\prime}{\sqrt{(\sigma_{b}^2 + \frac{\sigma_{W}^{2}}{\din}\, x\cdot x)(\sigma_{b}^2 + \frac{\sigma_{W}^{2}}{\din}\, x^\prime\cdot x^\prime)}} \Bigg], \nonumber
\end{eqnarray} 
which is derived via both methods of \cite{yang} and \cite{williams} in \eqref{deriverelukernel}.

\subsubsection*{\gnet}

While the two previous examples are well-studied in the literature, such as \cite{williams} and \cite{chosaul}, in
this Section we introduce a new activation function in order to obtain a translation invariant GP kernel. This architecture is obtained by adding a normalization layer after the usual exponential activation, according to the process
\begin{eqnarray}
x \rightarrow \exp(W \, x + b ) \rightarrow \frac{\exp(W \, x + b )}{\sqrt{K_{\text{exp}}(x,x)}} \rightarrow f(x),
\end{eqnarray}
where $K_{\text{exp}}(x,x)$ is the $2$-pt function of the intermediate exponential activation layer. 

The resulting activation
is\footnote{We thank Greg Yang for discussions of activations
that yield translationally invariant kernels.}
\begin{eqnarray}
  \sigma(x) = \frac{\exp{(W \, x + b)}}{\sqrt{\exp{ [ 2(\sigma_{b}^{2} + \frac{\sigma_{W}^{2}}{\din} x^{2}) ] }}},
\end{eqnarray}
which unlike usual activations depends on both the preactivation and the input; we have written it entirely in terms of the input
by writing $z = W x + b$.
The associated GP kernel
\begin{eqnarray}
  K_{\text{Gauss}}(x, x') = \sigma_{b}^2 + \sigma_{W}^2\,\exp\left[- \frac{\sigma_{W}^{2}|x - x^\prime |^2}{2 \din} \right], \label{gnetkernel}
\end{eqnarray}  
is derived via both the methods of \cite{yang} and \cite{williams} in \ref{Gaussnet}. The kernel
is particularly simple, a Gaussian\footnote{Hence our choice of the name "\gnet".} in the Euclidean distance between the two inputs. Accordingly, the kernel is invariant under the translation map $x\to x+c$, $x' \mapsto x'+c$
for any constant vector $c$.
The normalization in $\sigma$ is crucial for translation invariance.

\subsection{Experiments: Falloff to GP Feynman Diagrams at Large Width} \label{sec:gpexperiments}

We now wish to demonstrate that in the infinite width limit the experimental results for the $n$-pt functions 
$G^{(n)}$ converge to those of the Gaussian process, $\ggp^{(n)}$. For simplicity we will consider the case of a 
single-dimensional output, $d_\text{out}=1$. Accordingly, we define the deviation in the $n$-pt function, \amnew{in terms of experimental $n$-pt correlation of NN outputs $G^{(n)}$ defined in \eqref{defineGn}, and theoretical predictions by free field theory $\ggp^{(n)}$, given below}
\begin{equation}
    \Delta G^{(n)}(x_1,\dots,x_n) = G^{(n)}(x_1,\dots x_n) - \ggp^{(n)}(x_1,\dots,x_n).
\end{equation}
For measuring the size of the deviation with respect to the experimental results, it is convenient to define
the normalized deviation $m_n(x_1,\dots,x_n)=\Delta G^{(n)}/\ggp^{(n)}(x_1,\dots x_n)$.

The $2$-pt deviation, given below, is the difference between the experimental $2$-pt function and the kernel of the corresponding Gaussian process. 
\begin{eqnarray}
    \Delta G^{(2)} &=& G^{(2)}(x_1,x_2) - \ggp^{(2)}(x_{1}, x_{2}) \nonumber
  \\ &=& \mathbb{E} (f(x_{1})\, f(x_{2})) - K(x_{1}, x_{2}) \nonumber
  \\ &=& \frac{1}{n_{\text{nets}}} \sum_{\alpha}^{n_{\text{nets}}} f_{\alpha}(x_{1}) f_{\alpha} (x_{2}) - K(x_{1}, x_{2})
\end{eqnarray}
$f_{\alpha} (x_{i})$ denotes the output of the $\alpha^\text{th}$ network for the input $x_{i}$.

The $4$-pt and $6$-pt deviations are expressed using Wick contractions of products of the kernels evaluated at Wick pairs $p = (x_{i}, x_{j})$. The $4$-pt deviation is given by 
\begin{eqnarray}
 \label{eq:4-ptdev}
    \Delta G^{(4)} &=& G^{(4)}(x_1,x_2, x_3, x_4) - \ggp^{(4)}(x_{1}, x_{2},x_3,x_4) \nonumber
  \\ &=& \mathbb{E} (f(x_1)\,f(x_2)\, f(x_3)\, f(x_4)) - \sum_{p \in \wick(x_1,x_2, x_3, x_4)} K(p_{1}) K(p_{2}) \nonumber
  \\ &=& \frac{1}{n_{\text{nets}}} \sum_{\alpha}^{n_{\text{nets}}} f_{\alpha}(x_{1}) f_{\alpha} (x_{2}) f_{\alpha}(x_{3}) f_{\alpha} (x_{4})  
  - \bigg[ K(x_{1}, x_{2}) K(x_{3}, x_{4})  \\ &+&  K(x_{1}, x_{3}) K(x_{2}, x_{4}) + K(x_{1}, x_{4}) K(x_{2}, x_{3}) \bigg] \nonumber \\
  \nonumber
  \\ \nonumber &=& \frac{1}{n_{\text{nets}}} \sum_{\alpha}^{n_{\text{nets}}} f_{\alpha}(x_{1}) f_{\alpha} (x_{2}) f_{\alpha}(x_{3}) f_{\alpha} (x_{4}) 
  - \Bigg[ \, \, \begin{tikzpicture}[line width=1pt, scale=0.5, baseline=-.5*\fontsizeshift]
		\coordinate (C) at (0,0);
	\coordinate (x1) at (-1,1);
	\coordinate (x2) at (-1,-1);
	\coordinate (x3) at (1,1);
	\coordinate (x4) at (1,-1);
	\coordinate (ls) at (0,-.3);
	\coordinate (ls1) at (0,.3);
	\draw[scs] (x1) -- (x2);
	\draw[scs] (x3) -- (x4);
	\draw[fill] (x1) circle (.05);
	\draw[fill] (x2) circle (.05);
	\draw[fill] (x3) circle (.05);
	\draw[fill] (x4) circle (.05);
		\node at ($(x1) + 2*(ls1)$) {$x_1$};
	\node at ($(x2) + 2*(ls)$) {$x_2$};
	\node at ($(x3) + 2*(ls1)$) {$x_3$};
	\node at ($(x4) + 2*(ls)$) {$x_4$};
\end{tikzpicture} \, \, + \, \,  
\begin{tikzpicture}[line width=1pt, scale=0.5, baseline=-.5*\fontsizeshift]
		\coordinate (C) at (0,0);
	\coordinate (x1) at (-1,1);
	\coordinate (x2) at (-1,-1);
	\coordinate (x3) at (1,1);
	\coordinate (x4) at (1,-1);
	\coordinate (ls) at (0,-.3);
	\coordinate (ls1) at (0,.3);
	\draw[scs] (x1) -- (x3);
	\draw[scs] (x2) -- (x4);
	\draw[fill] (x1) circle (.05);
	\draw[fill] (x2) circle (.05);
	\draw[fill] (x3) circle (.05);
	\draw[fill] (x4) circle (.05);
		\node at ($(x1) + 2*(ls1)$) {$x_1$};
	\node at ($(x2) + 2*(ls)$) {$x_2$};
	\node at ($(x3) + 2*(ls1)$) {$x_3$};
	\node at ($(x4) + 2*(ls)$) {$x_4$};
\end{tikzpicture} \, \,  +
\begin{tikzpicture}[line width=1pt,scale=0.5,baseline=-.5*\fontsizeshift]
		\coordinate (C1) at (-.2,-.2);
	\coordinate (C2) at (.2,.2);
	\coordinate (x1) at (-1,1);
	\coordinate (x2) at (-1,-1);
	\coordinate (x3) at (1,1);
	\coordinate (x4) at (1,-1);
	\coordinate (ls) at (0,-.3);
	\coordinate (ls1) at (0,.3);
	\draw[scs] (x1) -- (x4);
	\draw[scs] (x2) -- (C1);
	\draw[scs] (C2) -- (x3);
	\draw[fill] (x1) circle (.05);
	\draw[fill] (x2) circle (.05);
	\draw[fill] (x3) circle (.05);
	\draw[fill] (x4) circle (.05);
		\node at ($(x1) + 2*(ls1)$) {$x_1$};
	\node at ($(x2) + 2*(ls)$) {$x_2$};
	\node at ($(x3) + 2*(ls1)$) {$x_3$};
	\node at ($(x4) + 2*(ls)$) {$x_4$};
\end{tikzpicture}\,\,  \Bigg]
 \end{eqnarray}
The last line follows from \eqref{4-ptwick}.

The $6$-pt function is obtained similarly using products of three kernels, where we will use the abbreviation $K(x_{i}, x_{j}) =: K_{ij}$ going forward. It is given by
\begin{eqnarray}
 \label{eq:6-ptdevsec2.4}
    \Delta G^{(6)} &=& G^{(6)}(x_1,x_2, x_3, x_4, x_5, x_6) - \sum_{p \in \wick(x_1,x_2, x_3, x_4, x_5, x_6)} K(p_{1}) K(p_{2}) K(p_{3}) \nonumber
  \\ &=& \frac{1}{n_{\text{nets}}} \sum_{\alpha}^{n_{\text{nets}}} f_{\alpha}(x_{1}) f_{\alpha} (x_{2}) f_{\alpha}(x_{3}) f_{\alpha} (x_{4}) f_{\alpha} (x_{5}) f_{\alpha} (x_{6})  \nonumber
  \\ &-& \bigg[ K_{12} K_{34} K_{56} +  K_{12} K_{35} K_{46} + K_{12} K_{36} K_{45} + K_{13} K_{24} K_{56} + K_{13} K_{25} K_{46} + K_{13} K_{26} K_{45} \nonumber
  \\ &+& K_{14} K_{23} K_{56} + K_{14} K_{25} K_{36} + K_{14} K_{26} K_{35} + K_{15} K_{23} K_{46} + K_{15} K_{24} K_{36} + K_{15} K_{26} K_{34}  \nonumber
  \\ &+& K_{16} K_{23} K_{45} + K_{16} K_{24} K_{35} + K_{16} K_{25} K_{34} \bigg].
 \end{eqnarray}
 We remind the reader that in Section \ref{correlation_fn_section} we discussed a simple check of the combinatorics: the
 Wick contractions for $\ggp^{(n)}$ ($n$ even) involve the number of ways of connecting $n$ points in pairs, which
 is $(n-1)!!=(n-1)\,(n-3)\,\,\dots\,1.$ For the $n=6$ case this predicts $15$ terms, which we see explicitly in the last equality of \eqref{eq:6-ptdevsec2.4}.

The $2$-pt, $4$-pt and $6$-pt deviations, denoted by $\Delta G^{(n)}$, are calculated experimentally at widths 
\begin{equation}
	N \in \{2, 3, 4, 5, 10, 20, 50, 100, 500, 1000\}
\end{equation} for all three architectures. To do this, for each architecture we
calculate the $n$-pt correlation functions 
\begin{equation}
  G^{(n)}(x_1,\dots,x_n) = \frac{1}{n_\text{nets}} \sum_{\alpha\in \text{nets}}^{n_\text{nets}}\, f_\alpha(x_1)\, \dots \, f_\alpha(x_n)
  \label{defineGn}
\end{equation}
 of outputs $f_{\alpha}(x_i)$ from \amnew{100} experimental runs with \amnew{$10^5$} networks each, with weights and biases $\theta$ drawn as described in Section \ref{sec:networksetup} with $\mu_W = \mu_b = 0$ and $\sigma_b=\sigma_W=1$ for Erf-net and \gnet.
Since our experiments have $d_\text{in}=1$, 
this is $b^{i}\sim \cN(0,1)$ for $i=1,2$ and $W^0\sim\cN(0,1)$, 
$W^1\sim\cN(0,1/N)$.  All experiments are done with $100$ experiments of $10^5$ nets each, for a total of $n_\text{nets} = 10^{7}$. We choose ReLU-net to have a bias of 0 for scaling reasons explained in Section \ref{sec:rge}, so $\sigma_b=0$ for this case. \amnew{Inputs of ReLU-net are chosen to be all positive so that the kernel is always nonzero; the kernel is zero for opposite-signed inputs in the case $\din=1$ and $\sigma_b=0$. Inputs of Erf-net are chosen to be positive to simplify the functional form in Section \ref{sec:EFT}, so that the kernel is always positive. These networks are defined for any $x \in \mathbb{R^{\din}}$, but the inputs used for experiments were chosen where the finite width NGP is well approximated by local-operator correction terms to the associated log-likelihood. More generally, a neural network may only have some subset of the input space where local correction terms to the GP can give a good description of the NGP at finite width; we will systematically study non-local terms in follow-up work. }
\begin{table}[t]
	\centering
	\begin{tabular}{|l|l|l|}
	\hline
			 & inputs $\{x_{i}\}$                                     & $(\sigma_{W}^2, \, \sigma_{b}^{2})$ \\ \hline
	\gnet    & $\{-0.01, -0.006, -0.002, +0.002, +0.006, +0.01\}$ & $(1, 1)$                          \\ \hline
	Erf-net  & $\{ +0.002, +0.004, +0.006, +0.008, +0.010, +0.012\}$              & $(1, 1)$                          \\ \hline
	ReLU-net & $\{ +0.2, +0.4, +0.6, +0.8, +1.0, +1.2\}$          & $(1, 0)$                          \\ \hline
	\end{tabular}
	\caption{Parameters that define the networks used in GP and non-GP experiments in Sections \ref{sec:gpexperiments} and \ref{sec:EFTforSingle}.}
	\label{tab:parameters}
	\end{table}

\begin{figure}[thb]
    \centering
  \makebox[\textwidth][c]{
  \includegraphics[width=0.36\textwidth]{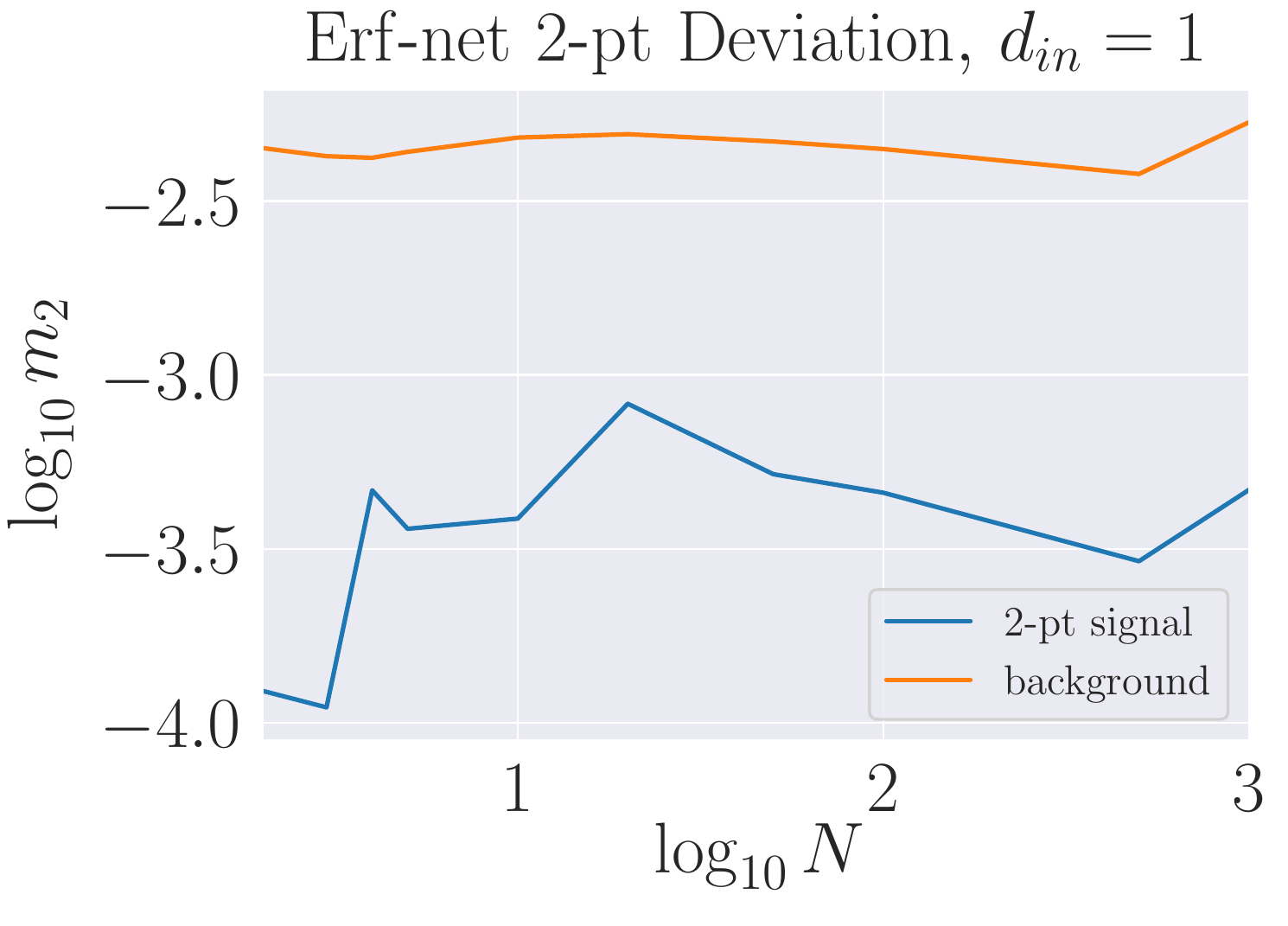}
  \includegraphics[width=0.36\textwidth]{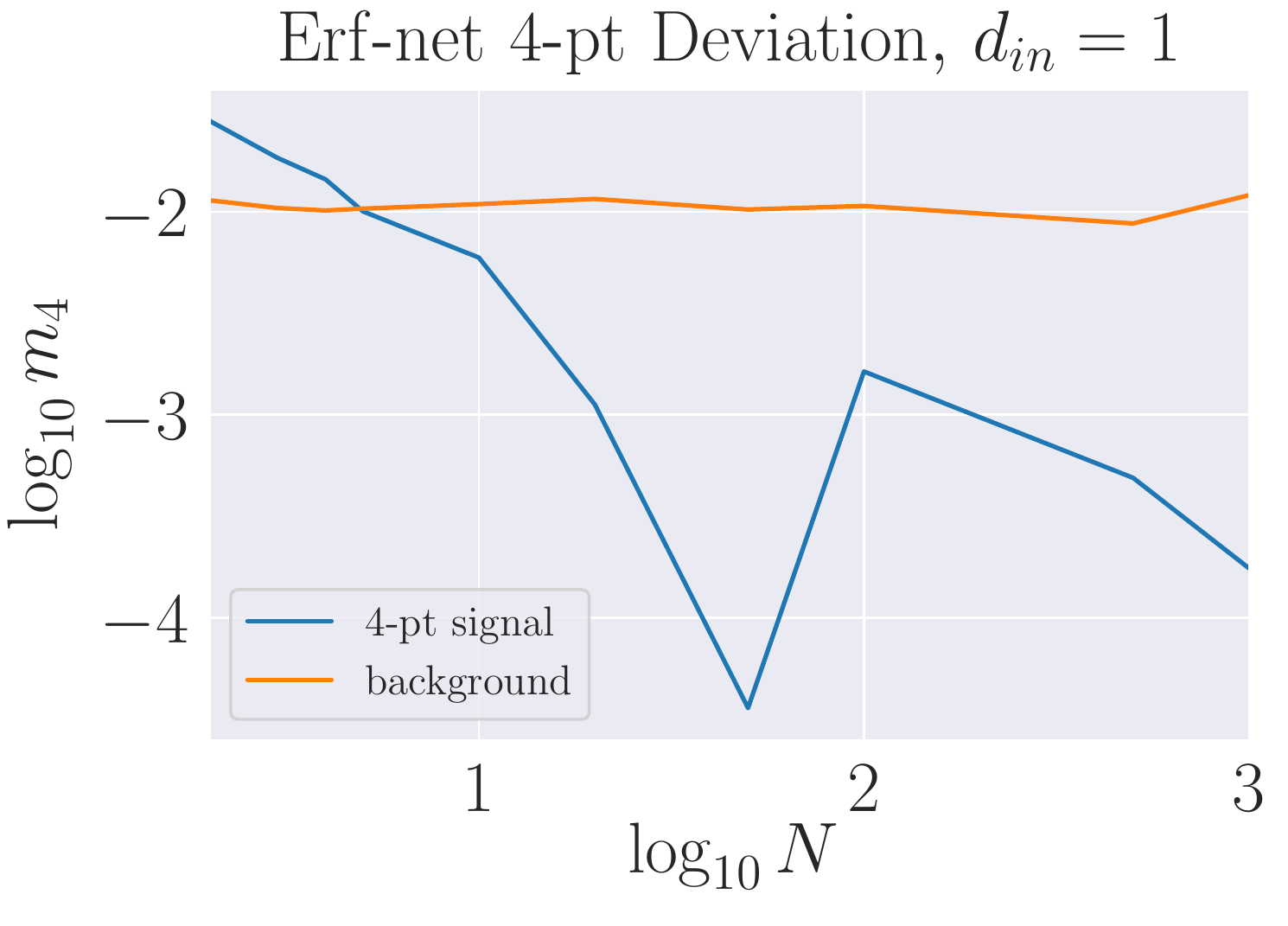}
  \includegraphics[width=0.36\textwidth]{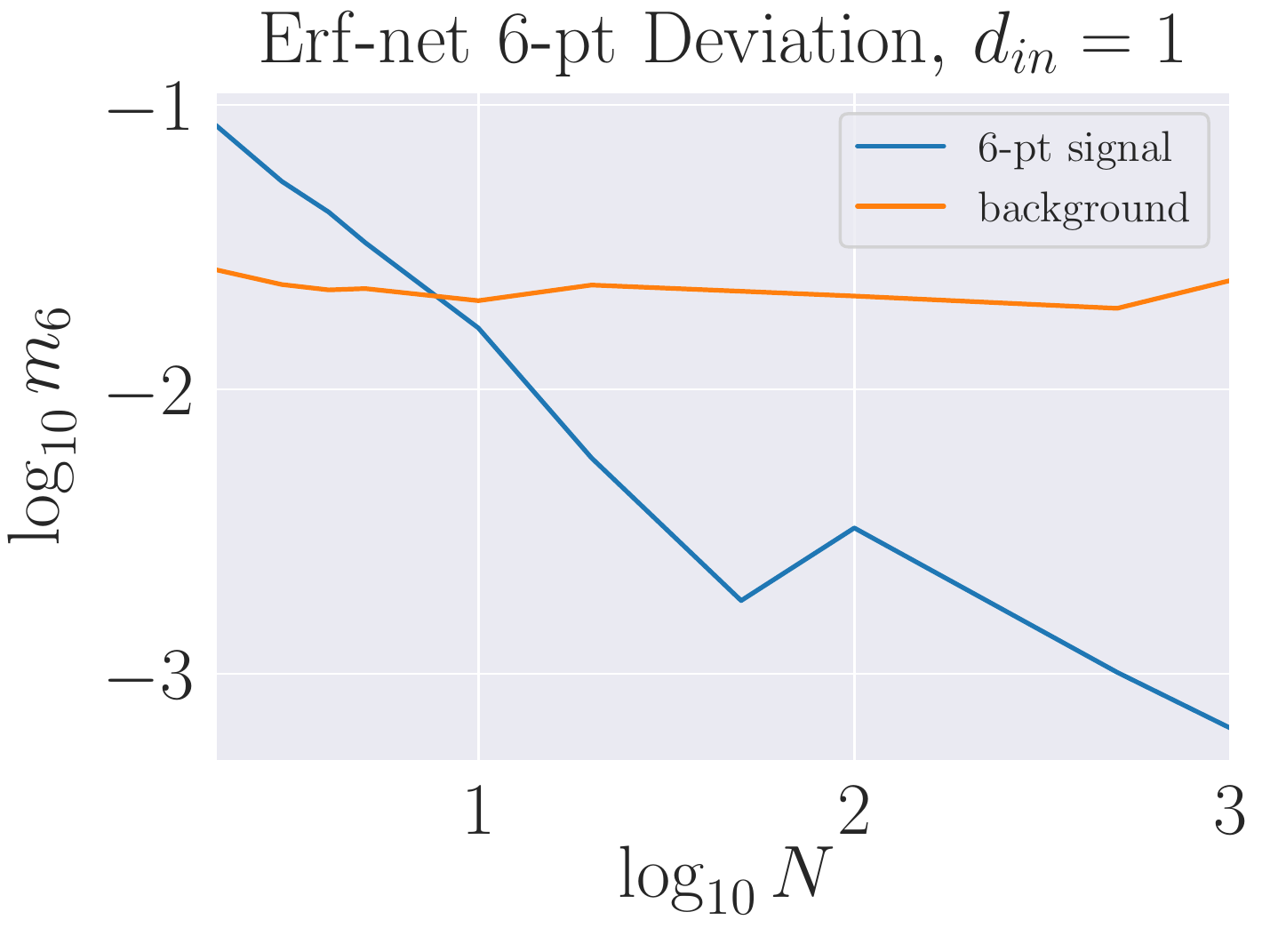}}
  \makebox[\textwidth][c]{
  \includegraphics[width=0.36\textwidth]{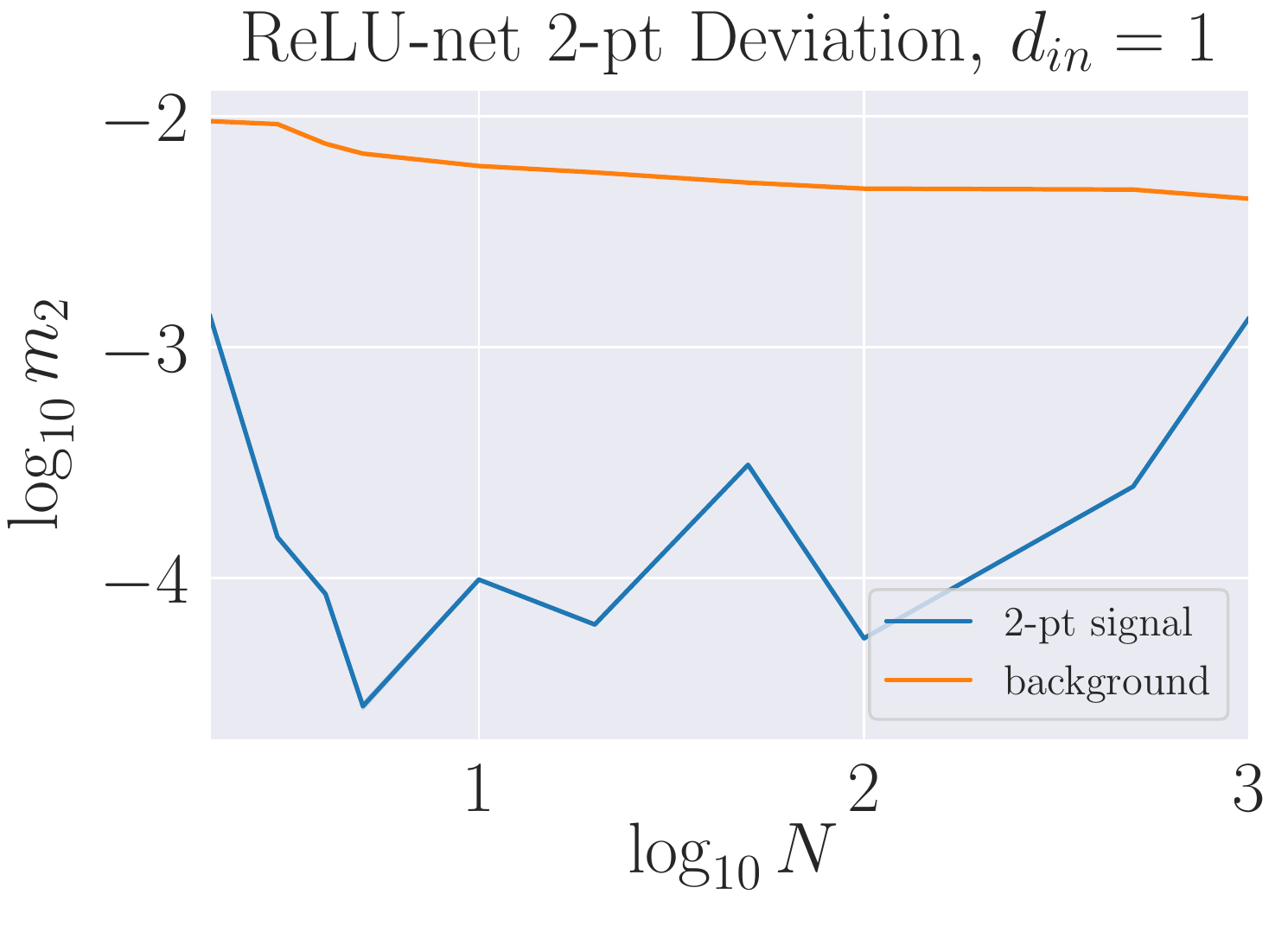}
  \includegraphics[width=0.36\textwidth]{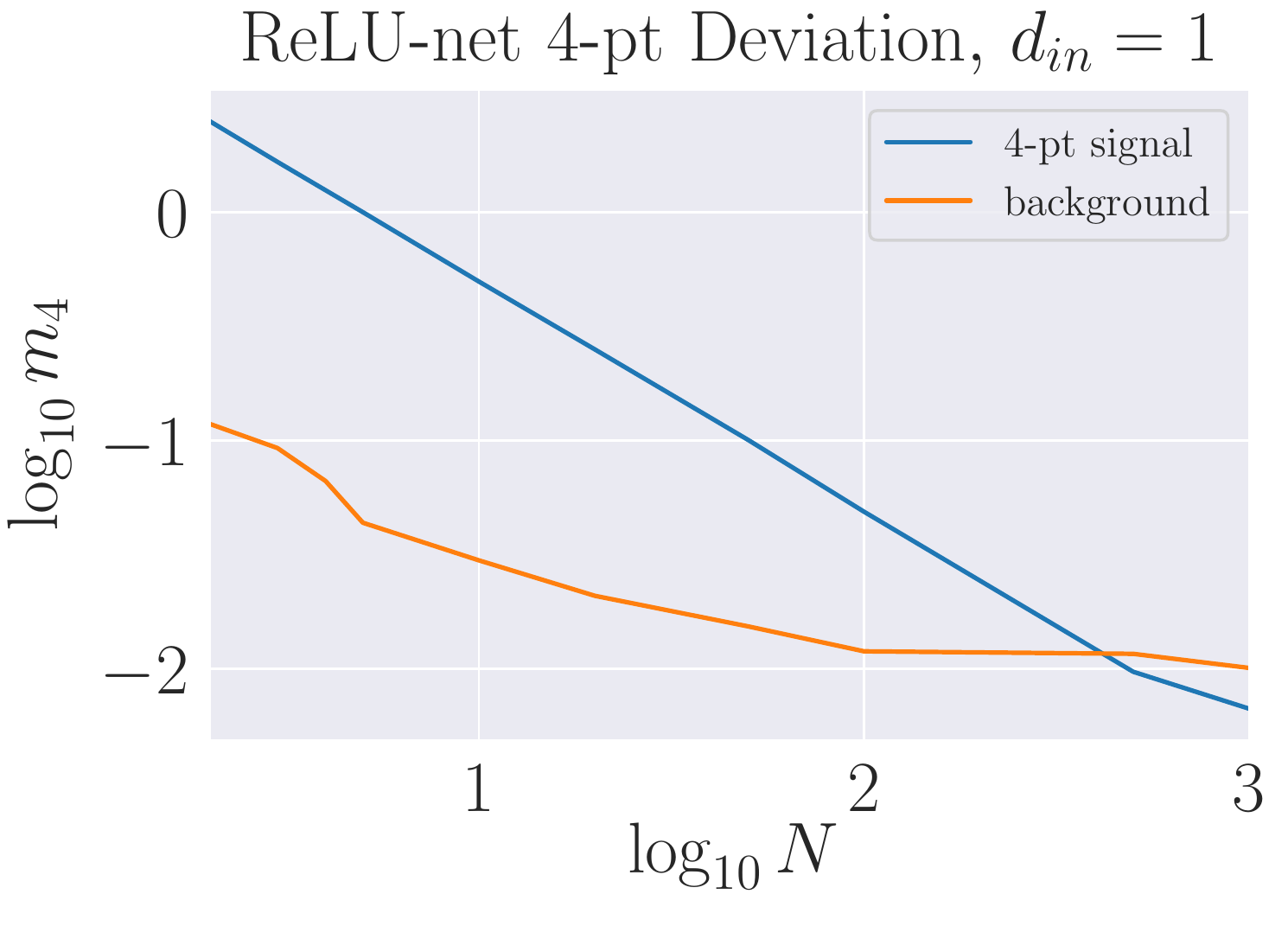}
  \includegraphics[width=0.36\textwidth]{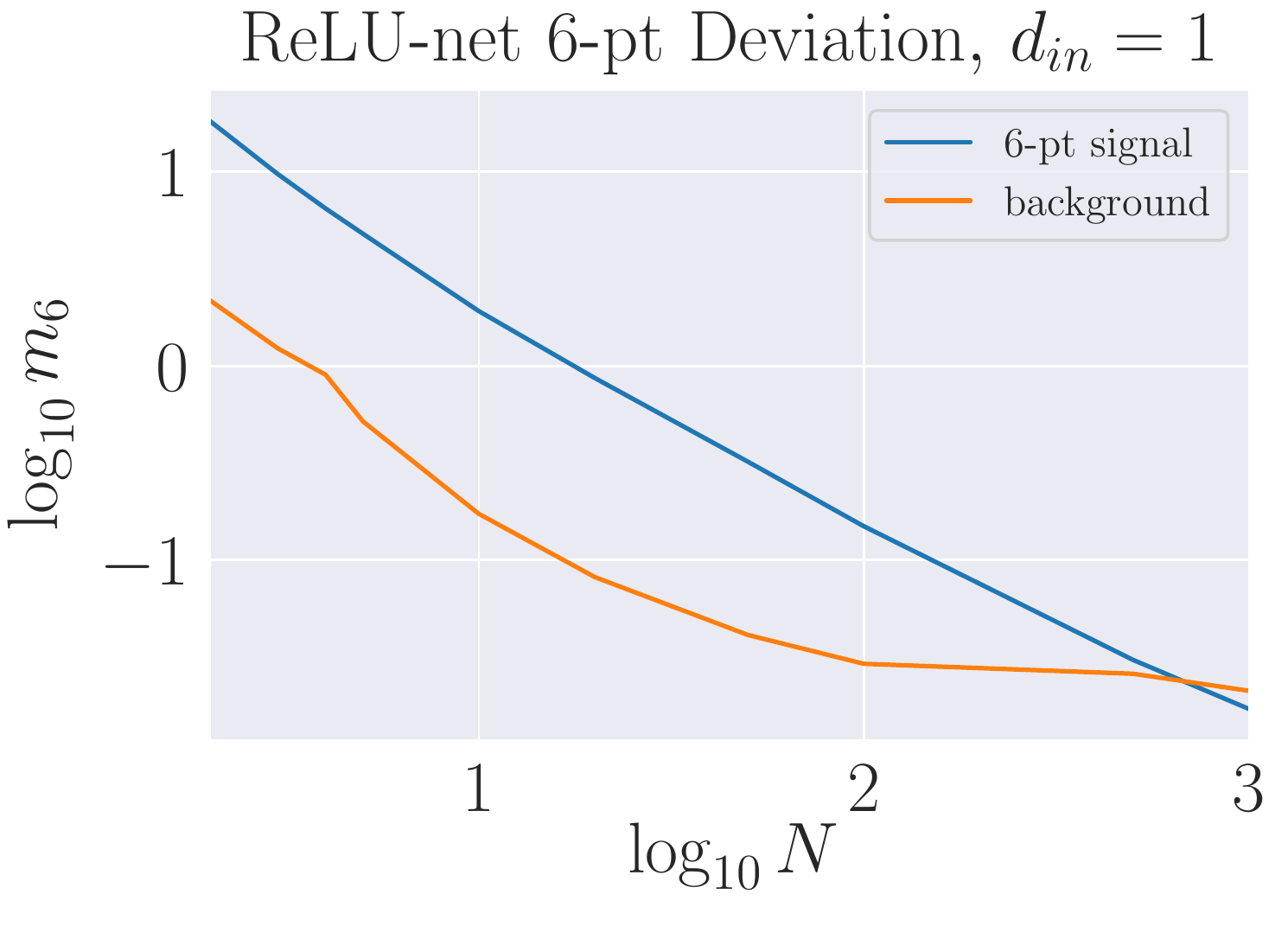}}
  \makebox[\textwidth][c]{
  \includegraphics[width=0.36\textwidth]{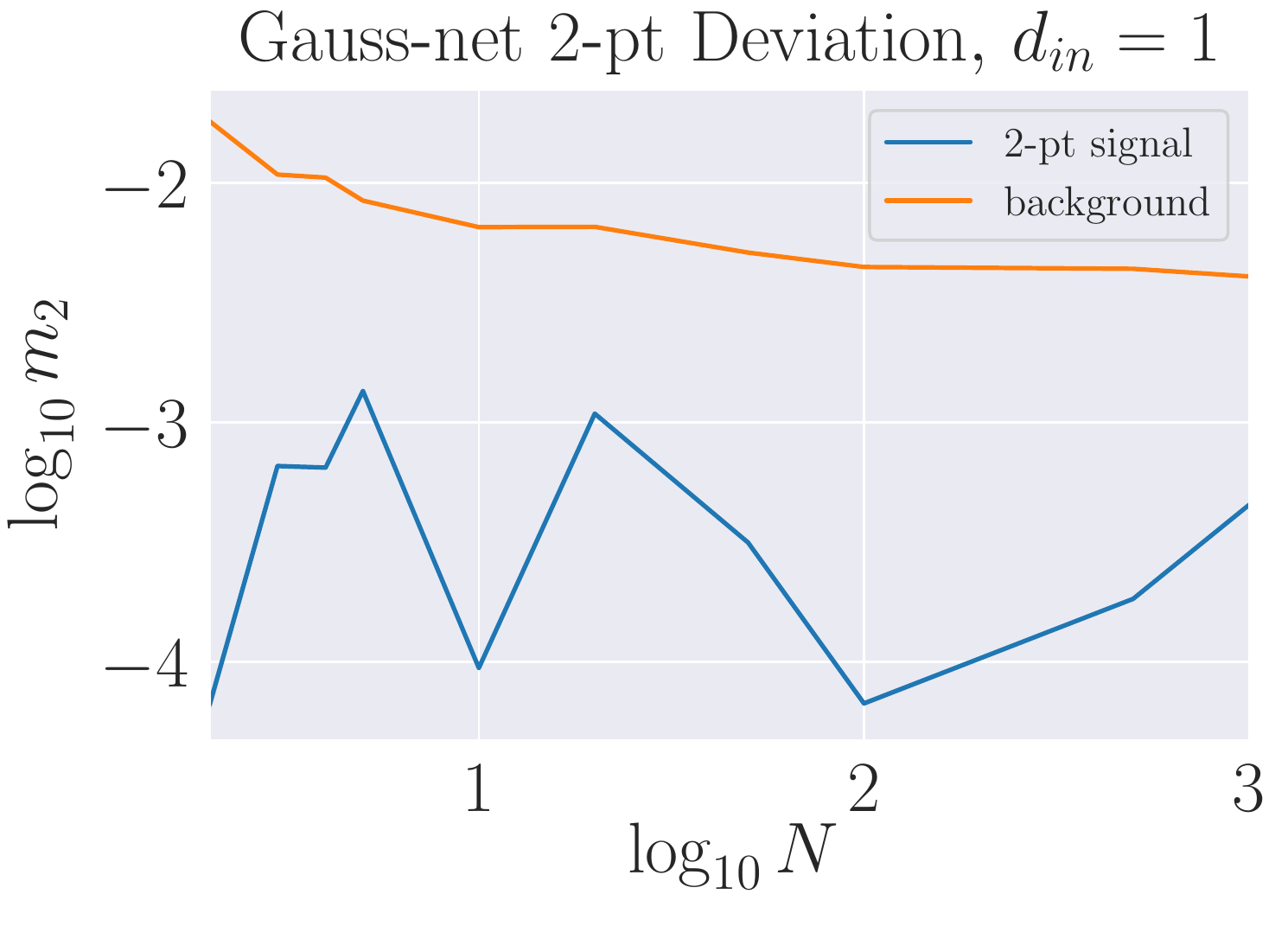}
  \includegraphics[width=0.36\textwidth]{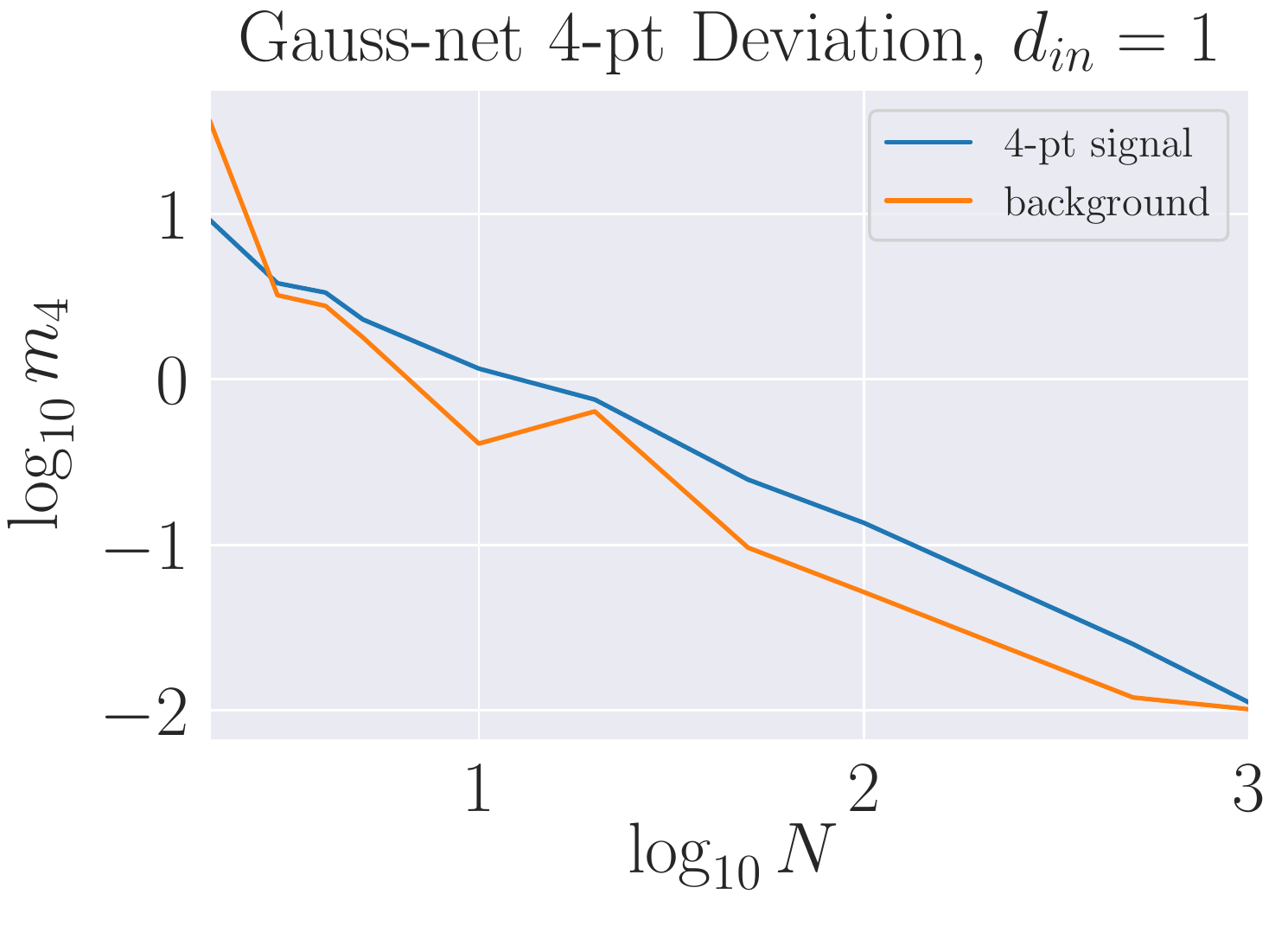}
  \includegraphics[width=0.36\textwidth]{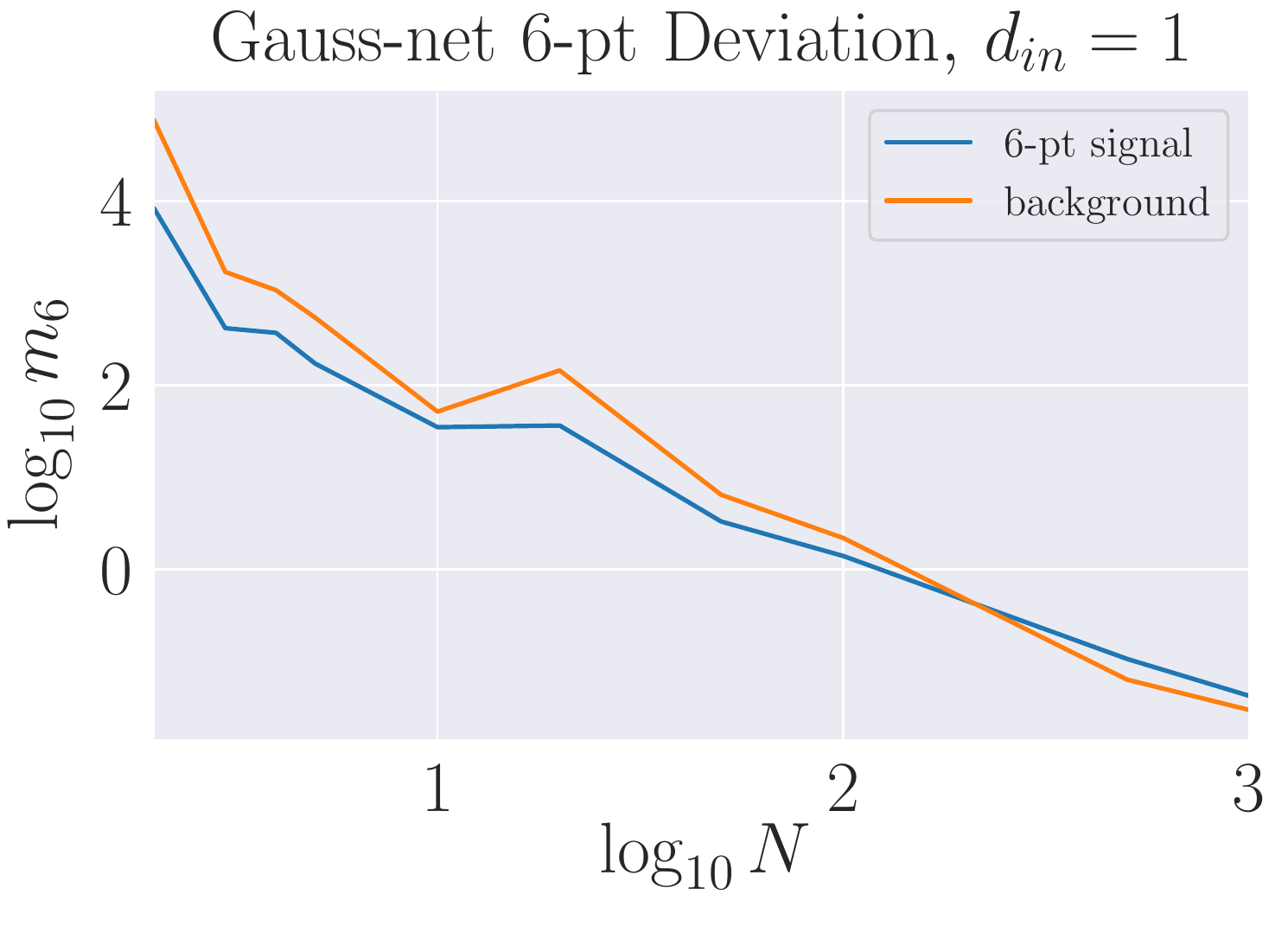}}
  \caption{ Normalized $2$-pt deviation, $4$-pt deviation and $6$-pt deviation of Erf-net, ReLU-net and \gnet. Signals represent the elements of the tensor $m_{n}$, and backgrounds are the average elementwise standard deviation across $100$ experiments of the tensor elements.}
  \label{fig1}
\end{figure}

To study the falloff to GP predictions $\ggp^{(n)}$
as $N\rightarrow\infty$, the $n$-pt deviations $\Delta G^{(n)}$ are normalized by the GP prediction to obtain a measure $m_{n} = \Delta G^{(n)}/\ggp^{(n)}$, with input dependence implicit. These are plotted in Figure \ref{fig1} in comparison to a background defined to be the average elementwise standard deviation (across the $100$ experiments) of the experimental $m_n$, as a function of width $N$, where for each width the solid blue line is the mean, and error bars are the 
95\
\amnew{We expect $\Delta G^{(n)} = G^{(n)} - G^{(n)}_{GP}$ to show a $1/N$ falloff for all $n$; we will prove this in Section \ref{sec:nscaling}.} Figure \ref{fig1} shows that $2$-pt deviations $\Delta G^{(2)}$ are below the background level and therefore consistent with zero for all three architectures, indicating that the kernel is an exact measure of the $2$-pt correlation function even away from the GP. This is expected, since in Appendix \ref{kernel_derivation_appendix} it is shown that the GP kernels associated to the architectures we study are the exact two-point functions at all widths; from here on we therefore use kernel and $2$-pt function interchangeably. The $4$-pt and $6$-pt signals are linearly decreasing (on a log-log scale) with increasing width in the region above the background level, falling below the background level at higher widths. The slope of the line in the region above the background is $\simeq -1$, indicating in our experiments that
\begin{equation}
\Delta G^{(n)} \propto N^{-1}
\end{equation}
for $n = 4, 6$, \amnew{up to experimental errors, shown by the flattening of signal well below the background}. This explicitly demonstrates the falloff to GP as the width increases. 

At large width, therefore,  we see that these networks are drawn from GPs, with their statistics entirely determined by Wick contractions of the appropriate kernels, yielding $\ggp^{(n)}$. At small width, the GP prediction no longer correctly predicts the experimental $n$-pt function; the neural networks are not drawn from a GP, but instead an NGP. Our goal is to develop a method for capturing non-Gaussian
corrections to the log-likelihood.

\subsection{$N$-scaling of Correlation Functions of Fully-Connected Networks}
\label{sec:nscaling}

\jimnew{
Having seen that $\Delta G^{(n)} \propto N^{-1}$, we would like to understand this and related results in more theoretical detail. For convenience, we recall that linear output layer acts on the last post-activation as 
	\begin{equation}
		f_{\theta,N}(x) = z_{k,1} = \sum_{j = 1}^{N} W_{k,j,1} x_{j,1} + b_{k,1},
	\end{equation}
where the post-activation is labeled $x_1$ since we are considering networks with a single hidden layer. The following results also hold for deep fully-connected networks with $l$ layers, with $x_1$ replaced by $x_l$ and $W_1$ by $W_l$. Henceforth we will use $x_l$ and $W_l$, for generality, and suppress indices associated with the output and input dimensions, since they do not affect the result (or alternatively take $d_\text{in}=d_\text{out}=1$). By $x_{l}^i$ we denote the last post-activation of the $i^{\text{th}}$ input.

We are interested in the $N$-scaling properties of the correlation functions in these networks. 
Recall that a neural network with linear output layer is the sum of two functions $f_b$ and $f_W$ drawn from independent processes, $f_b\sim \mathcal{P}_b$ and $f_W\sim \mathcal{P}_W$, and that the bias term $f_b$ does not depend on $N$. When its entries are drawn from Gaussian, it is an ultra-local GP at all $N$. Let $\mathcal{P}$ be the process from which $f=f_b+f_W$ is drawn. The correlation functions of $\mathcal{P}$ decompose into sums of products of correlation functions from $\mathcal{P}_b$ and $\mathcal{P}_W$, and since $\mathcal{P}_b$ does not depend on $N$ and is Gaussian, all $N$-dependence and non-Gaussianities arise from $\mathcal{P}_W$. 

We first wish to determine the leading $N$-scaling of the $2k$-pt correlator $G^{(2k)}(x_1,\dots,x_{2k})$. In our normalizations which following, e.g., \cite{lee}, the $N$-dependence of the $k=1$ case drops out,
\begin{align}
	G^{(2)}(x_1,x_2) &= \mathbb{E}[b_1^2] + \sum_{i,j} \bE[W_i W_j]\bE[x_{i}^1 x_{j}^2 ] \nonumber \\
	&= \mathbb{E}[b_1^2] + \frac{\sigma_W^2}{N} \sum_{i} \bE[x_{i}^1 x_{i}^2 ],
\end{align}
where the factor of $N$ is canceled by the $N$ from the sum. $W_i = W_{l,i}$ and $x_{i,j} = x_{l,i}^j$ are respectively the weight of the  $l^{\text{th}}$ layer and the post-activation of the $l^{\text{th}}$ layer of $j^{\text{th}}$ input; we will utilize this notation for rest of this subsection. The leading $N$-dependence must come from the $k=2$ case, then, where the relevant term is (with Einstein summation implied)
\begin{align}
	G^{(4)}(x_1,x_2,x_3,x_4) &= \bE[W_i W_jW_k W_m]\bE[x_{i}^1x_{j}^2x_{k}^3 x_{m}^4 ] + \dots \nonumber \\
	&= \frac{3\sigma_W^4}{N^2}\sum_{i} \bE[x_{i}^1 x_{i}^2 x_{i}^3 x_{i}^4 ] + \frac{\sigma_W^4}{N^2}\sum_{i \neq j} \Big[ \bE[x_{i}^1x_{i}^2x_{j}^3x_{j}^4] \nonumber \\
	&+ \bE[x_{i}^1x_{j}^2x_{j}^3x_{i}^4] + \bE[x_{i}^1x_{j}^2x_{i}^3x_{j}^4] + \dots \Big],
\end{align}
where the terms that survive in the sum have either two pairs of indices equal or all indices the same. The former correspond to contributions from products of $G_2^{(2)}$'s and come with a factor of $N(N-1)$ that results in a mixed scaling of the $(1-\frac{1}{N})$, whereas the latter is a contribution to $G^{(4)}$ that is not a product of terms in lower-point functions and  comes with a factor of $N$, leaving a $1/N$-dependent factor.} 
\amnew{Thus we expect that we might have $\Delta G^{(4)}$, the deviation from GP results, scales as $1/N$. Being more precise, $\Delta G^{(4)}$ is the same as the \emph{connected} piece of $G^{(4)}$ (see, e.g., QFT textbooks such as \cite{Schwartz}),
\begin{eqnarray}
&~&G^{(4)}(x_1, x_2, x_3, x_4)\bigg|_{\text{connected}} 
= G^{(4)}(x_1, x_2, x_3, x_4) - G^{(2)}(x_1, x_2)G^{(2)}(x_3, x_4) -  G^{(2)}(x_1, x_3)G^{(2)}(x_2, x_4)  \nonumber \\
&-& G^{(2)}(x_1, x_4)G^{(2)}(x_2, x_3), \label{G4con} \\
&=& \sum_i  \langle  W_i^4 \rangle \mathbb{E}[x_i^1 x_i^2 x_i^3 x_i^4] + \sum_{i \neq j} \langle  W_i^2 \rangle \langle  W_j^2 \rangle \Bigg( \mathbb{E}[x_i^1 x_i^2]\mathbb{E}[ x_j^3 x_j^4] 
+ \mathbb{E}[x_i^1 x_i^3]\mathbb{E}[ x_j^2 x_j^4] \nonumber  \\  
&+& \mathbb{E}[x_i^1 x_i^4]\mathbb{E}[ x_j^2 x_j^3]  \Bigg) - \sum_{i ,  j} \langle  W_i^2 \rangle \langle  W_j^2 \rangle \Bigg( \mathbb{E}[x_i^1 x_i^2]\mathbb{E}[ x_j^3 x_j^4] + \mathbb{E}[x_i^1 x_i^3]\mathbb{E}[ x_j^2 x_j^4] + \mathbb{E}[x_i^1 x_i^4]\mathbb{E}[ x_j^2 x_j^3]  \Bigg) , \nonumber \\
&=&  \sum_i \langle  W_i^4 \rangle \mathbb{E}[x_i^1 x_i^2 x_i^3 x_i^4] - \sum_{i} \langle  W_i^2 \rangle^2 \Bigg( \mathbb{E}[x_i^1 x_i^2]\mathbb{E}[ x_i^3 x_i^4]
+ \mathbb{E}[x_i^{1} x^{3}_i]\mathbb{E}[ x^{2}_i x^{4}_i]   + \mathbb{E}[x^{1}_i x^{4}_i]\mathbb{E}[ x^{2}_i x^{3}_i]  \Bigg), \nonumber \\
&=& \frac{\sigma_W^4}{N^2} \sum_i \Bigg( 3\cdot \mathbb{E}[x^{1}_i x^{2}_i x^{3}_i x^{4}_i] -  \mathbb{E}[x^{1}_i x^{2}_i]\mathbb{E}[ x^{3}_i x^{4}_i]
- \mathbb{E}[x^{1}_i x^{3}_i]\mathbb{E}[ x^{2}_i x^{4}_i]  - \mathbb{E}[x^{1}_i x^{4}_i]\mathbb{E}[ x^{2}_i x^{3}_i] \Bigg) \propto \frac{1}{N},
\end{eqnarray}   
where $x_{i}^{k} = \sigma(z^{k}_{i, l-1}) = \sigma\left(\sum_{j=1}^{N^\prime} W^{k}_{j,l-1}x^{j}_{i,l-1} + b^k_{l-1}\right)$ is the postactivation of the $l^{\text{th}}$ layer; $N^\prime$ and $x_{i,l-1}^j$ are respectively the length and postactivation of $(l-2)^{\text{th}}$ layer. If $(l-1)^{\text{th}}$ layer is the input layer, then $N^\prime = \din$ and $x_{i,l-1} = x_i$.

Similarly, $N$-dependence of the $6$-pt function arises from 
\begin{equation}
	G^{(6)}(x_1,x_2,x_3,x_4,x_5,x_6) = \sum_{i,j,k,m,n,o} \bE[W_i W_jW_k W_m W_n W_o]\bE[x_{1}^i x_{2}^j x_{3}^k x_{4}^m x_{5}^n x_{6}^o ].
\end{equation}
}
\jimnew{The terms with all indices the same in pairs but different from one another results in $N(N-1)(N-2)$, leading to an overall scaling of $(1-\frac{3}{N} + \frac{2}{N^2})$ once the $N$-dependence of weight variances are taken into account. When four of the indices are the same and the remaining two the same, but different from the four, we have a contribution of $N(N-1)$ that results in a scaling of $(\frac{1}{N} - \frac{1}{N^2})$ after weight variances. We thus see that the leading non-trivial $N$-dependence is $\propto 1/N$, which can be interpreted as the contribution of the connected part of the $4$-pt function to the $6$-pt function, via multiplication with a $2$-pt function,. We have $\Delta G^{(6)}\propto 1/N$, in agreement with experiments. }

\amnew{
However, we are also interested in the scaling of the \emph{connected} contributions to the $6$-pt function. It is given by
\begin{eqnarray}
&&G^{(6)}(x_1,x_2,x_3,x_4,x_5,x_6)\bigg|_{\text{connected}} = G^{(6)}(x_1,x_2,x_3,x_4,x_5,x_6) 
- \sumabcdef \bigg( \nonumber \\
&&
 G^{(4)}(x_a,x_b,x_c,x_d)G^{(2)}(x_e,x_f)  - 2 \cdot
G^{(2)}(x_a,x_b) G^{(2)}(x_c,x_d) G^{(2)}(x_e,x_f)  \bigg), \label{G6con} \\
&=&  G^{(6)}(x_1,x_2,x_3,x_4,x_5,x_6) -  \sumabcdef \bigg(
 G^{(4)}(x_a,x_b,x_c,x_d)\bigg|_{\text{connected}}G^{(2)}(x_e,x_f)  \nonumber \\ &+& 
G^{(2)}(x_a,x_b) G^{(2)}(x_c,x_d) G^{(2)}(x_e,x_f)  \bigg), \nonumber \\
&=& \sum_i  \langle  W^6_i \rangle \mathbb{E}[x^{1}_i x^{2}_i x^{3}_i x^{4}_i x^{5}_i x^{6}_i] +\sumabcdef \Bigg(  \sum_{i \neq j} \langle  W^4_i \rangle \langle  W_j^2 \rangle \mathbb{E}[x^{a}_i x^{b}_i x^{c}_i x^{d}_i]\mathbb{E}[ x^{e}_j x^{f}_j] \nonumber \\
&+& \sum_{i \neq j \neq k} \langle  W^2_i \rangle \langle  W_j^2 \rangle \langle W^2_k \rangle \mathbb{E}[x^{a}_i x^{b}_i] \mathbb{E} [x^{c}_j x^{d}_j]\mathbb{E}[ x^{e}_k x^{f}_k] - \Bigg[ \sum_{i, j} \langle  W^4_i \rangle \langle  (W_j^2 \rangle \mathbb{E}[x^{a}_i x^{b}_i x^{c}_i x^{d}_i]\mathbb{E}[ x^{e}_j x^{f}_j]  \nonumber \\
 &-& \sum_{i, j } \langle  W^2_i \rangle^2 \langle  W_j^2 \rangle \mathbb{E}[x^{a}_i x^{b}_i] \mathbb{E} [x^{c}_i x^{d}_i]\mathbb{E}[ x^{e}_j x^{f}_j] +  \sum_{i, j, k} \langle  W^2_i \rangle \langle  W_j^2 \rangle \langle W^2_k \rangle \mathbb{E}[x^{a}_i x^{b}_i] \mathbb{E} [x^{c}_j x^{d}_j]\mathbb{E}[ x^{e}_k x^{f}_k] \Bigg] \Bigg), \nonumber \\
 &=&  \sum_i  \langle  W^6_i \rangle \mathbb{E}[x^{1}_i x^{2}_i x^{3}_i x^{4}_i x^{5}_i x^{6}_i]  - \sumabcdef  \sum_{i} \langle  W^4_i \rangle \langle  W_i^2 \rangle \mathbb{E}[x^{a}_i x^{b}_i x^{c}_i x^{d}_i]\mathbb{E}[ x^{e}_i x^{f}_i] + \text{terms}~, \nonumber \\
\end{eqnarray} 
and 
\begin{eqnarray}
\text{terms} &=& \sumabcdef \Bigg( \sum_{i \neq j \neq k} \langle  W^2_i \rangle \langle  W_j^2 \rangle \langle W^2_k \rangle \mathbb{E}[x^{a}_i x^{b}_i] \mathbb{E} [x^{c}_j x^{d}_j]\mathbb{E}[ x^{e}_k x^{f}_k] \nonumber \\
&+& \sum_{i, j } \langle  W^2_i \rangle^2 \langle  W_j^2 \rangle  \mathbb{E}[ x^{e}_j x^{f}_j] \bigg(\mathbb{E}[x^{a}_i x^{b}_i]\mathbb{E} [x^{c}_i x^{d}_i] + \mathbb{E}[x^{a}_i x^{c}_i]\mathbb{E} [x^{b}_i x^{d}_i] + \mathbb{E}[x^{a}_i x^{d}_i]\mathbb{E} [x^{b}_i x^{c}_i] \bigg) \nonumber \\
&-& \sum_{i, j, k} \langle  W^2_i \rangle \langle  W_j^2 \rangle \langle W^2_k \rangle \mathbb{E}[x^{a}_i x^{b}_i] \mathbb{E} [x^{c}_j x^{d}_j]\mathbb{E}[ x^{e}_k x^{f}_k] \Bigg),
\end{eqnarray}
where $\mathcal{P}(abcdef)$ indicates that the sum is over all unique terms made by choices of $\{a, b, c, d, e, f\} \in \{1, 2, 3, 4, 5, 6\}$, where no two elements are the same. Using the expansion,
\begin{eqnarray}
&&\sum_{i, j, k} \langle  W^2_i \rangle \langle  W_j^2 \rangle \langle W^2_k \rangle \mathbb{E}[x^{a}_i x^{b}_i] \mathbb{E} [x^{c}_j x^{d}_j]\mathbb{E}[ x^{e}_k x^{f}_k]  = -2 \cdot \sum_{i} \langle  W^2_i \rangle^3  \mathbb{E}[x^{a}_i x^{b}_i] \mathbb{E} [x^{c}_i x^{d}_i]\mathbb{E}[ x^{e}_i x^{f}_i] + \nonumber \\
&+& \sum_{i, j } \langle  W^2_i \rangle^2 \langle  W_j^2 \rangle  \mathbb{E}[ x^{e}_j x^{f}_j] \bigg(\mathbb{E}[x^{a}_i x^{b}_i]\mathbb{E} [x^{c}_i x^{d}_i] + \mathbb{E}[x^{a}_i x^{c}_i]\mathbb{E} [x^{b}_i x^{d}_i] + \mathbb{E}[x^{a}_i x^{d}_i]\mathbb{E} [x^{b}_i x^{c}_i] \bigg) \nonumber \\
&-& \sum_{i \neq j \neq k} \langle  W^2_i \rangle \langle  W_j^2 \rangle \langle W^2_k \rangle \mathbb{E}[x^{a}_i x^{b}_i] \mathbb{E} [x^{c}_j x^{d}_j]\mathbb{E}[ x^{e}_k x^{f}_k],
\end{eqnarray}
the \emph{connected} part of $G^{(6)}$ is obtained as 
\begin{eqnarray}
&& G^{(6)}(x_1,x_2,x_3,x_4,x_5,x_6)\bigg|_{\text{connected}} =  \sum_i \Bigg[ \langle  W^6_i \rangle \mathbb{E}[x^{1}_i x^{2}_i x^{3}_i x^{4}_i x^{5}_i x^{6}_i] \nonumber \\
 &-& \sumabcdef \Bigg( \sum_{i} \langle  W^4_i \rangle \langle  W_i^2 \rangle \mathbb{E}[x^{a}_i x^{b}_i x^{c}_i x^{d}_i]\mathbb{E}[ x^{e}_i x^{f}_i] -2 \cdot \sum_{i} \langle  W^2_i \rangle^3  \mathbb{E}[x^{a}_i x^{b}_i] \mathbb{E} [x^{c}_i x^{d}_i]\mathbb{E}[ x^{e}_i x^{f}_i] \Bigg) \Bigg] \nonumber \\
 &=& \frac{\sigma_W^6}{N^3} \sum_i \Bigg[15\cdot\mathbb{E}[x^{1}_i x^{2}_i x^{3}_i x^{4}_i x^{5}_i x^{6}_i] -  3\cdot\sumabcdef  \mathbb{E}[x^{a}_i x^{b}_i x^{c}_i x^{d}_i]\mathbb{E}[ x^{e}_i x^{f}_i] \nonumber \\
 &+& 2\cdot\sumabcdef  \mathbb{E}[x^{a}_i x^{b}_i] \mathbb{E} [x^{c}_i x^{d}_i]\mathbb{E}[ x^{e}_i x^{f}_i] 
  \Bigg] .
\end{eqnarray}
That is, we have
\begin{equation}
	G^{(6)}(x_1,\dots,x_{6})|_\text{connected} \propto \frac{1}{N^{2}},
\end{equation}
which is $1/N$-suppressed relative to the connected part of the $4$-pt function. 

Based on the structure of the two examples we have computed, it is natural to make a conjecture regarding the 
connected contributions to the $2k$-pt function. In both cases, when representing the connected correlator in terms of a sum of full correlators, with some terms involving lower correlators, we saw that the only terms that didn't cancel were the ones with all indices the same. It is natural to expect this to be general, in which case we have
\begin{align}
G^{(2k)}(x_1,\dots, x_{2k})|_\text{connected} = \left[ G^{(2k)}(x_1,\dots, x_{2k}) - S(x_1,\dots,x_{2k})\right]|_\text{internal indices same},
\end{align}
where the terms in brackets are the general result from consideration of the connected generating  functional $W[J]$, which yields an expression for the connected correlator as the full correlator minus sums of products of lower point functions, the latter represented by $S(x_1,\dots, x_{2k})$. The restriction to all internal indices
is specific to the linear output layer we have utilized. For instance, in the connected $4$-pt function 
\begin{equation}
	S(x_1,\dots,x_4)
=G^{(2)}(x_1,x_2)G^{(2)}(x_3,x_4)+G^{(2)}(x_1,x_3)G^{(2)}(x_2,x_4)+G^{(2)}(x_1,x_4)G^{(2)}(x_2,x_3),
\end{equation} 
and the associated restriction in the case of linear output layer is
\begin{align}
\left[G^{(2)}(x_1,x_2)G^{(2)}(x_3,x_4)+G^{(2)}(x_1,x_3)G^{(2)}(x_2,x_4)+G^{(2)}(x_1,x_4)G^{(2)}(x_2,x_3)\right]|_\text{internal indices same} \nonumber \\
= \sum_{i} \langle  W_i^2 \rangle^2 \Bigg( \mathbb{E}[x_i^1 x_i^2]\mathbb{E}[ x_i^3 x_i^4]
+ \mathbb{E}[x_i^{1} x^{3}_i]\mathbb{E}[ x^{2}_i x^{4}_i]   + \mathbb{E}[x^{1}_i x^{4}_i]\mathbb{E}[ x^{2}_i x^{3}_i]  \Bigg)), 
\end{align}
appears directly in the result for the connected $4$-pt function. For our weight distributions, the conjecture implies
\begin{equation}
	G^{(2k)}(x_1,\dots,x_{2k})|_\text{connected} \propto \frac{1}{N^{k-1}},
\end{equation}
which explicitly holds for the two cases we computed, $k=1$ and $k=2$.

Foreshadowing our non-Gaussian process treatment, our explicit computations have an important implication: the simplest way to have the connected $4$-pt and $6$-pt correlators scale this way is to have the $4$-pt and $6$-point couplings scale as $1/N$ and $1/N^2$, respectively, since they generate tree-level contributions to the associated connected correlators.

\begin{figure}
    \centering
  \includegraphics[width=0.48\textwidth]{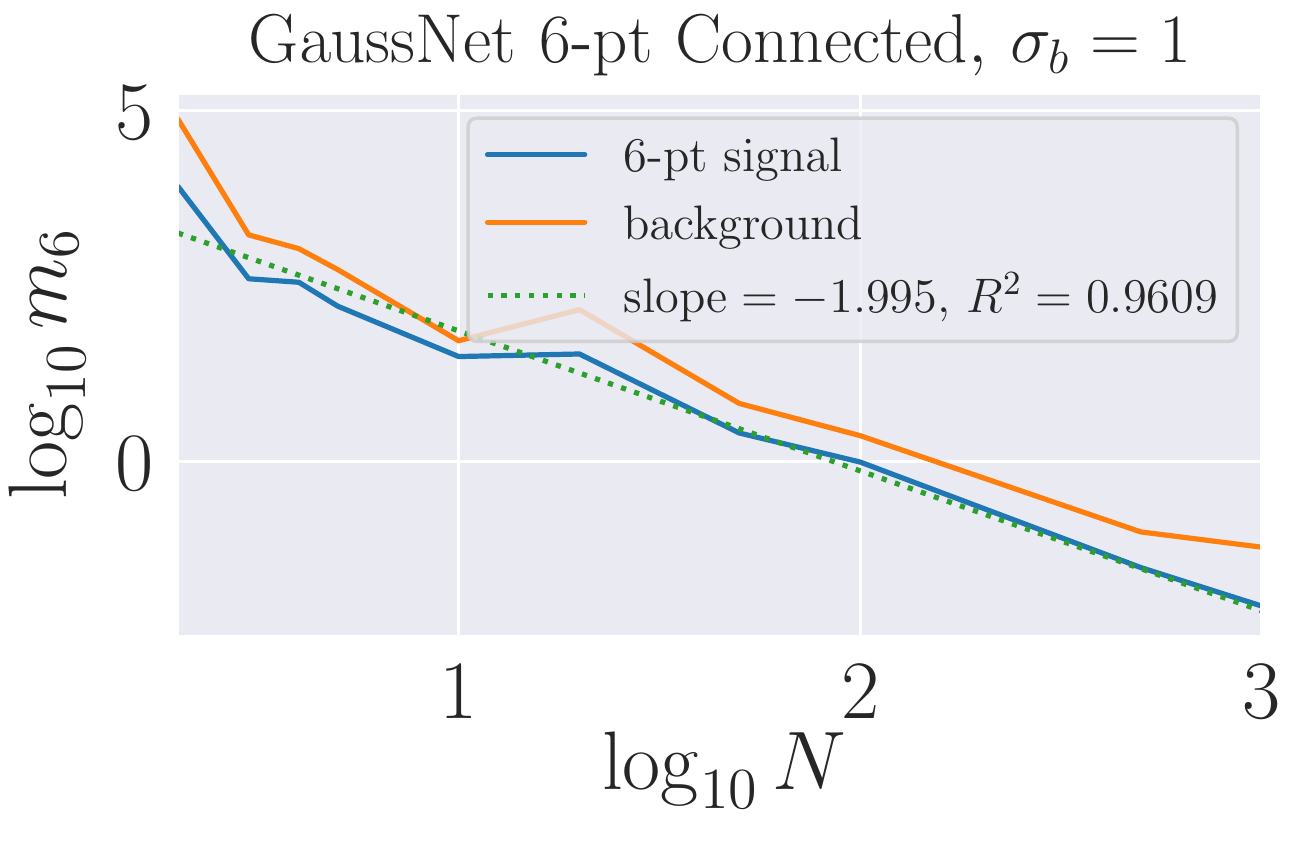}
  \includegraphics[width=0.48\textwidth]{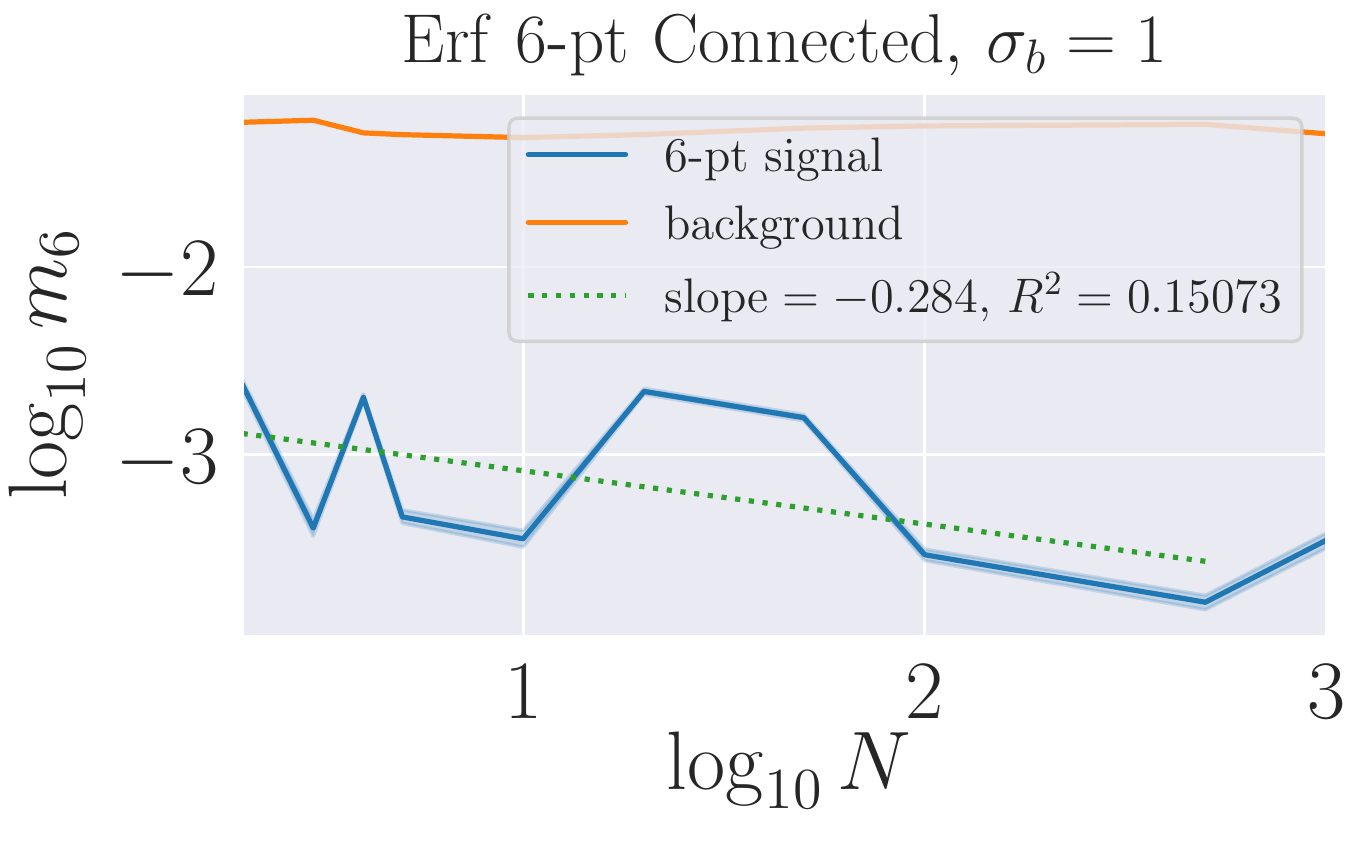}
  \includegraphics[width=0.48\textwidth]{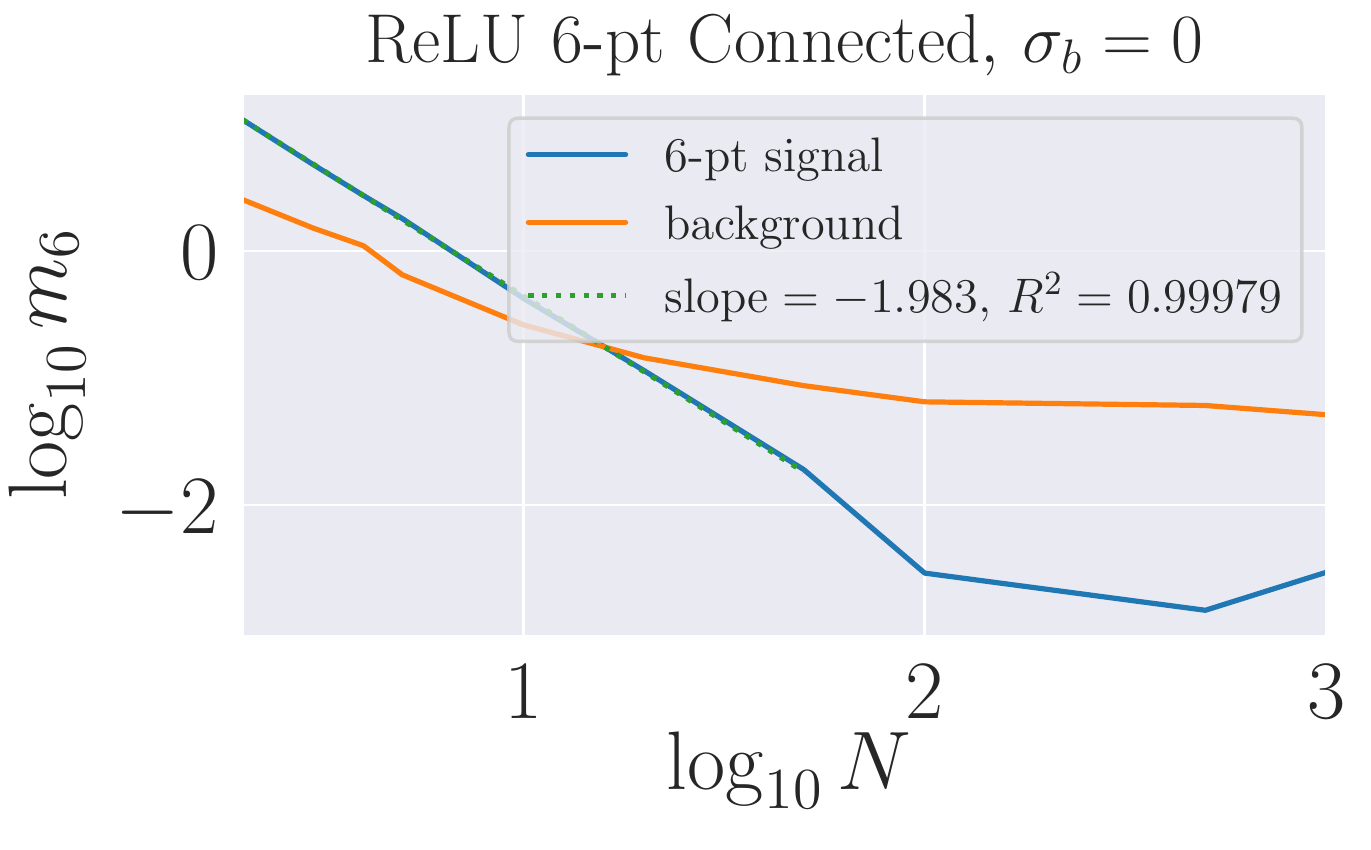}
  \caption{\amnew{Normalized connected part of $6$-pt function of Erf-net, ReLU-net and \gnet. Signals represent the elements of the normalized connected-$G^{(6)}$ tensor, and backgrounds are obtained using \eqref{eq:G6conback}. Slopes are about $-2$, in agreement with theory, whenever signal is not well below background; for Erf-net, it is for all $N$.}}
  \label{figG6con}
\end{figure}
The background levels in Figure \ref{figG6con} is determined by the average of element-wise standard deviations of $100$ experiments for each of the $n$-pt functions at each width, and by further propagating the error to obtain the error in the connected $6$-pt function, 
\begin{equation}
	\delta G^{(6)}\Bigg|_{\text{connected}} = \sqrt{ (\delta G^{(6)})^2 + (G^{(2)}\delta G^{(4)})^2 }. \label{eq:G6conback}
\end{equation}
No error for $G^{(2)}$ is included because we use the exact theory expression. Erf-net connected-$G^{(6)}$ is well below background at all widths, and does not show the expected $1/N^2$ dependence. ReLU-net shows expected $1/N^2$ falloff until it is well below the background level. \gnet~also shows expected $1/N^2$ dependence. Thus, up to statistical errors $G^{(6)}\bigg|_{\text{connected}}$ shows $1/N^2$ dependence. 
}

\section{Neural Networks and Non-Gaussian Processes with Effective Field Theory \label{sec:EFT}}

In this Section we propose using an approach to quantum field theory known
as Wilsonian effective field theory (EFT) to understand and analyze neural networks away from their
GP limits. From an ML point of view, it can be thought of as a useful way to determine minimal log-likelihoods
of NGPs by determining the most relevant non-Gaussian corrections to the GP.

The essential idea is that the GP action $\sgp$ does not suffice to determine correlation
functions $G^{(n)}$ away from the GP limit, as demonstrated experimentally in Section \ref{sec:gpexperiments}, 
but the principles of EFT allows for the determination
of corrections $\Delta S$ to the GP log-likelihood, yielding an effective log-likelihood for the NGP\footnote{Henceforth, the NGP effective action.} associated
to the finite-width neural networks,
\begin{equation}
  S = \sgp + \Delta S.
\end{equation}
The organizing principles of EFT are symmetries and scales, which together allow for the
determination of an appropriate $\Delta S$ that may have its parameters fixed by experiments.
It may then be used to make effective (correct) predictions
for other experiments. 

For the ML reader, let us briefly tour these ideas in physics, where effective field theories correctly describe a vast array of systems,
from superfluids and superconductors in condensed matter physics to beta decay and elementary
particle interactions\footnote{The Standard Model of particle physics is an effective
field theory, and for some measurements it yields perhaps the highest precision agreement between theory and
experiment in all of science.} in high energy physics. As a concrete example, consider beta decay, a 
process by which a neutron decays into a proton, electron, and anti-neutrino,
\begin{equation}
n \to p + e^- +\bar \nu_e.
\end{equation}
From the perspective of the Standard Model, we have a deep knowledge of this process:
the neutrons and protons are made up of quarks and the decay process
is mediated by an intermediate W-boson. However, this detailed 
knowledge did not keep Fermi from arriving at an EFT description  of the 
process in $1933$ \cite{fermi},
long before quarks and W-bosons were even theorized, let alone discovered over
$35$ years later. Though symmetries permeate both Fermi's theory and the Standard Model,
the crucial insight is one of energy scale: Fermi's interaction that effectively
describes beta decay is obtained from the Standard Model by going to lower and lower
scattering energies, at which point the contribution from the intermediate W-boson is negligible and
an effective four-fermion interaction suffices to describe the process.

\bigskip

In this spirit, we propose understanding neural networks away from GP limits 
in terms of effective field theory. Specifically, EFT allows for the determination of
an NGP effective action, with its parameters fixed by experiments, which is then used to 
make verifiable predictions. \amnew{To that end, the correspondence in Table \ref{tab:NGPQFT} identifies the NN input $x \in \bR^{\din}$ with a point in space or momentum space in field theory, the NN kernel with free or exact propagator in QFT, depending on the NNGP in infinite width limit. The NN output $f(x)$ and log-likelihood now corresponds to the interacting field and the EFT action respectively.}

Rather than organizing according to
length or momentum scale, the NN-QFT correspondence that we are developing replaces fields as a function of space
with neural networks as a function of input, suggesting organization of the problem according to input scale.
The interpretation of this in ML depends on the problem, but an example of ``input scale'' in images would
be brightness.

As in physics, we utilize the
quadratic terms to determine classical scaling dimensions, which here are 
given by
the probability of an infinite-width neural net (GP draw) $f$
\begin{equation}
P[f] = e^{-\sgp[f]} = \text{exp}\left[-\frac12 \int d^{d_\text{in}}x \, d^{d_\text{in}} y \, f(x) \Xi(x,y) \, f(y)\right].
\end{equation}
Specifically,
$\sgp$ must be dimensionless; it scales as $x^0$, and we write $[\sgp]=0$. Since $d^{d_\text{in}}x$ scales as $x^{d_\text{in}}$ it has
input dimensions of $d_\text{in}$, $[d^{d_\text{in}} x] = d_\text{in}$, and similarly $[d^{d_\text{in}}y]=d_\text{in}$. $[S]=0$ then  determines
a relation between dimensions of $f$ and $\Xi$, $[S] = 2d_\text{in} + 2[f] + [\Xi] = 0$,
which determines the classical scaling dimension of the neural network $f$,
\begin{equation}
[f] = -\frac{2d_\text{in} + [\Xi]}{2}.
\end{equation}
This in turn may be used to determine the dimensions of the coefficients of operators that might appear in $\Delta S$. For instance, consider operators
\begin{equation}
  \mathcal{O}_k := g_k \, f(x)^k
\end{equation}
appearing in $\Delta S$ as $\int d^{\din} x \, \cO_k$. Then $[\Delta S] = 0$
requires $d_\text{in} + k[f] + [g_k] = 0$ and we have
\begin{equation}
	[g_k] = -d_\text{in} + \frac{k(2d_\text{in}+[\Xi])}{2}.
	\end{equation}
By \eqref{eq:functionalinverseK} and the fact\footnote{This follows from the natural $d$-dimensional extension of the identity  $\delta(ax)=\delta(x)/|a|$.}
that $[\delta^{(d_\text{in})}(x)] = -d_\text{in}$
we have $d_\text{in} + [\Xi] + [K] = -d_\text{in}$, and therefore can rewrite $[g_k]$ in terms
of the scaling dimension of the kernel
\begin{equation} \label{eq:gdim}
	[g_k] = -d_\text{in} - \frac{k \,[K]}{2}.
	\end{equation}
Unlike in many physical cases, for $[K] \ge 0$
the couplings $g_k$ have dimensions of the same sign $\forall \, k$. In Section \ref{sec:rg}
we will use arguments from Wilson's picture of the renormalization group to argue
that operators $\cO_k$ can be safely ignored for sufficiently large $k$.

\bigskip
How does one construct the effective action of an NGP associated to a neural network architecture that admits a known GP limit? 
Wilsonian EFT dictates the following rules:
\begin{itemize}
\item Determine the symmetries (or desired symmetries) respected by the system of interest.
\item Fix an upper bound $k$ on the dimension of any operator appearing in $\Delta S$.
\item Define $\Delta S$ to contain all operators of dimension $\leq k$ that respect the symmetries.
\end{itemize}
Since the GP limit is known, the NGP is defined by $S=\sgp + \Delta S$. As we will see in a moment, this allows for the
determination of correlation functions. By experimentally measuring them, one may fix coefficients of terms in $\Delta S$ and make subsequent predictions.

The desired symmetries may be determined by architecture considerations or, e.g., by demanding that the NGP
respects the same symmetries as its GP limit, in which case the NGP symmetries are the symmetries of the GP kernel
$K(x_1,x_2)$. The choice of a relevant value of $k$ is sometimes dictated by the system of study. For instance,
Fermi knew that his theory must have only spin-$1/2$ particles\footnote{Since the W-boson mediator had not been observed yet.} 
and that four of them must interact to describe beta decay. In QFT this requires a term schematically of the form $\psi \psi \psi \psi$, where $\psi$ is a field associated with a spin-$1/2$ particle.
Since $[\psi]=3/2$ in four dimensions, Fermi's theory needed $k \geq6$. In the examples we study in this paper,
we will see that it is crucial it have $k\geq 4$.

 \begin{table}[t]
	\centering 
	\begin{tabular}{|c|c|}
	\hline 
	NGP / finite NN& Interacting QFT \\ \hline
	input $x$ & external space or \jimnew{momentum space} point \\
	kernel $K(x_1,x_2)$ & free or exact propagator \\ 
	network output $f(x)$ & interacting field \\
	\amnew{non-Gaussianities} & \amnew{interactions} \\
	\amnew{non-Gaussian coefficients} & \amnew{coupling strengths} \\ 
	log probability & effective action $S$ \\  \hline
	\end{tabular}
	\caption{Correspondence between quantities in the NGP / finite-width neural network and QFT. See text for
	discussion on whether the kernel is the free or exact propagator.}
	\label{tab:NGPQFT}
\end{table}

\subsection{Correlation Functions in NGPs with Interacting Feynman Diagrams} \label{eftsec}

Having introduced EFT rules that allow for the determination of $\Delta S$, we must introduce a method for computing
NGP correlation functions. Before doing this explicitly in the cases of interest, 
we must briefly introduce the basics of cutoffs and perturbation theory.

First, perturbation theory: consider an NGP associated with a finite-width neural network architecture, with
associated effective action $S = \sgp + \Delta S$. The GP correlation functions $\ggp^{(n)}$
were exactly computable precisely because the action $\sgp$ was Gaussian, so non-Gaussian
corrections in $\Delta S$ prevent the NGP $n$-pt correlation functions
\begin{equation}
  G^{(n)}(x_{1},\dots, x_{n}) = \frac{\int df \,\, f(x_{1})\dots f(x_{n})\, e^{-S}}{Z_0},~~~~~\text{where}~~Z_0 = \int df e^{-S},
\end{equation}from 
being computed exactly. However, if the coefficients of operators in $\Delta S$
are appropriately small, approximating the $n$-pt functions using perturbation
theory is possible. In QFT, this corresponds to the existence of non-trivial interactions, where
the interaction strength being small yields a correction to the leading process. As an example,
consider corrections to the $n$-pt function arising from $\Delta S = \int d^{d_\text{in}}x \, g_k \, f(x)^k$, with $k>2$. Multiplying the numerator and denominator by $1/\zgpzero$ and expanding 
under the assumption of small $g_k$, we have
\begin{equation}
  G^{(n)}(x_{1},\dots, x_{n}) = \frac{\int df \,\, f(x_{1})\dots f(x_{n}) \left[1 - \int d^{d_\text{in}}x \, g_k f(x)^k + O(g_k^2)\right]\, e^{-\sgp }/\zgpzero}{\int df \,\, \left[1 - \int d^{d_\text{in}}x \, g_k f(x)^k + O(g_k^2)\right]e^{-\sgp }/\zgpzero}.
\end{equation}
Truncating at a desired order in $g_k$ (here just the leading correction), one obtains an
approximation for the $n$-pt function where the numerator
and denominator may both be computed via Wick's theorem, and both may be represented diagrammatically\footnote{\label{footnote:nobubbles}Some (all) diagrams in the numerator (denominator) contain vacuum bubbles. A vacuum bubble is a diagram that is not connected to any external points. They may arise as disconnected components of more complicated diagrams, as in the numerator. However, the vacuum bubbles that arise in the denominator precisely cancel those arising in the numerator, so that the final expression for $G^{(n)}$ does not contain any diagrams with vacuum bubbles.}. The $O(g_k^0)$ contribution in the numerator is precisely the GP $n$-pt function $\ggp^{(n)}$. The $O(g_k^1)$ term in the numerator may be computed via Wick's theorem, with one of the associated diagrams being a tree-level (no-loops) diagram by which the $n$ external points $\{x_1,\dots, x_n\}$ connect to the point $x$ that is integrated over; the latter is known as the
interaction vertex, and is referred to as an internal point. These calculations and vacuum
bubble cancellation is reviewed in Appendix \ref{app:gaussian}.

We now introduce cutoffs. Computing NGP correlation functions via perturbation theory can lead to
integrals over input space that naively yield divergences.
A simple way to treat the divergence is to cut them off by the replacement
\begin{equation}
S \to S_\Lambda,
\end{equation}
where $S_\Lambda$ differs from $S$ only in the fact that all integrals over input space
are bounded from below by $-\Lambda$ and above by $\Lambda$, where $\Lambda$ is positive and known as the cutoff.
In QFT it is usually integrals over momenta \amnew{(of the virtual particles  created and annihilated during particle interactions)} that are cut off, with the justification
that theories have finite regimes of validity; e.g., theories describing scattering done at one momentum
scale should not be valid up to arbitrarily high momenta. 
Accordingly, low energy
experiments with momenta $|p| \ll \Lambda$ should be insensitive to the choice of $\Lambda$.
This requirement imposes that the coefficients of operators in $S$ must obey a differential equation
known as the Wilsonian renormalization group equations (RGEs). We will discuss this at
length in Section \ref{sec:rge}, and demonstrate that, the NGP associated to finite-width ReLU-net
satisfies appropriate Wilsonian RGE.

Our discussion of perturbation theory focused on a particular term in $\Delta S$, however, it is of course 
possible to have many terms in $\Delta S$, each with its own coefficient to expand in, if it is small. This 
gives a general prescription for approximating correlation functions in neural network NGPs, which may be written as Feynman diagrams via the development of appropriate Feynman rules.

\bigskip

Of course, we are interested in doing neural net experiments that validate theoretical predictions, and so we again focus on the finite width single-layer 
networks introduced in Section \ref{sec:networksetup}. We have a single output, $d_\text{out}=1$, and therefore one might consider non-Gaussian
terms of the form
\begin{equation}
\Delta S = \int  \, d^{\din}x \, \left[ g \, f(x)^3 + \lambda \, f(x)^4 + \alpha \, f(x)^5 + \kappa \, f(x)^6 + \dots \right].
\end{equation}
However, all odd-point functions in our experiments must be zero, since the means of the weights and biases are zero. 
This motivates $g=\alpha=0$, since when expanded to linear order those
terms yield non-trivial contributions to the $3$-pt and $5$-pt functions, respectively. An intuitive way to see this is that 
in our randomly initialized neural nets $f$ and $-f$ should be on equal footing, and thus $S$ must have an $f\to -f$ symmetry (i.e., be invariant under this transformation) that would be broken by either $g\ne 0$ or $\alpha \ne 0$. Furthermore,
in Section \ref{sec:rg} we will explain why for sufficiently large $\Lambda$, $\kappa$ must be 
negligible, or \emph{irrelevant} in the sense of Wilsonian RG;
however, we will consider both until Section \ref{sec:rg}. By these considerations, the effective action we will
utilize for the remainder of the paper is
\begin{equation} \label{eq:eft-action}
S = \sgp + \int \, d^{d_{\text{in}}} x \,\, \left[\lambda \, f(x)^4 + \kappa \, f(x)^6\right].
\end{equation}
In our experiments we will also see that $\kappa$ is negligible, and therefore a single quartic correction will be sufficient to explain our finite-width neural net experiments.

With this effective action for the NGP, one may compute correlation functions in perturbation theory. Equivalently,
one may represent the correlation functions diagrammatically by the following Feynman rules. They may be stated in different ways according to the goals
at hand. Here we state them in a way relevant for computing the $O(\lambda^l \kappa^m)$ correction to $G^{(n)}(x_1,\dots,x_n)$, where the ``interaction vertices''
are $y_j$  ($j=1,\dots, l$) and $z_k$ ($k=1,\dots, m$).
\begin{itemize}
  \item[1)] For each of the $n$ external points $x_i$, draw 
 $ \begin{tikzpicture}[line width=1.0 pt,baseline=-2.8*\fontsizeshift]
		\coordinate (C) at (0.75,-1);
	\coordinate (x1) at (0,-1);
	\coordinate (ls) at (0,-.3);
	\draw[scs] (x1) -- (C);
	\draw[fill] (x1) circle (.05);
	\node at ($(x1) + 1.2*(ls)$) {$x_i$};
\end{tikzpicture} \,\,.$ 
  \item[2)] For each $y_j$, draw $\,\,\begin{tikzpicture}[line width=1.0 pt,baseline=-2.8*\fontsizeshift]
		\coordinate (C) at (0.75,-1);
	\coordinate (x1) at (0,-0.5);
	\coordinate (x2) at (1.5,-0.5);
	\coordinate (x5) at (0, -1.5);
	\coordinate (x6) at (1.5, -1.5);
	\coordinate (ls) at (0,-.3);
	\draw[scs] (x1) -- (C);
	\draw[scs] (x2) -- (C);
	\draw[scs] (x5) -- (C);
	\draw[scs] (x6) -- (C);
	\draw[fill] (C) circle (.05);
	\node at ($(C) + 1.2*(ls)$) {$y_j$};
\end{tikzpicture} $ \,\,. For each $z_k$, draw $\,\,\begin{tikzpicture}[line width=1.0 pt,baseline=-1.5*\fontsizeshift]
		\coordinate (C) at (0.75,-0.5);
	\coordinate (x1) at (0,0);
	\coordinate (x2) at (1.5,0);
	\coordinate (x3) at (0,-1);
	\coordinate (x4) at (1.5,-1);
	\coordinate (x5) at (1.5,-0.5);
	\coordinate (x6) at (0,-0.5);
	\coordinate (ls) at (0,-.3);
	\draw[scs] (x1) -- (C);
	\draw[scs] (x2) -- (C);
	\draw[scs] (x3) -- (C);
	\draw[scs] (x4) -- (C);
	\draw[scs] (x5) -- (C);
	\draw[scs] (x6) -- (C);
	\draw[fill] (C) circle (.05);
	\node at ($(C) + 1.2*(ls)$) {$z_k$};
\end{tikzpicture} \,\,.$
  \item[3)] Determine all ways to pair up the loose ends associated to $x_i$'s, $y_j$'s, and $z_k$'s. This will yield some number of topologically distinct diagrams. Draw them with solid lines.
  \item[4)] Write a sum over the diagrams with an appropriate combinatoric factor out front, which is the number of ways to form that diagram. Each diagram corresponds to an analytic term in the sum.
  \item[5)] For each diagram, write $-\int d^{d_\text{in}} y_j \, \lambda$ for 
    each  $\,\,\begin{tikzpicture}[line width=1.0 pt,baseline=-2.8*\fontsizeshift]
		\coordinate (C) at (0.75,-1);
	\coordinate (x1) at (0,-0.5);
	\coordinate (x2) at (1.5,-0.5);
	\coordinate (x5) at (0, -1.5);
	\coordinate (x6) at (1.5, -1.5);
	\coordinate (ls) at (0,-.3);
	\draw[scs] (x1) -- (C);
	\draw[scs] (x2) -- (C);
	\draw[scs] (x5) -- (C);
	\draw[scs] (x6) -- (C);
	\draw[fill] (C) circle (.05);
	\node at ($(C) + 1.2*(ls)$) {$y_j$};
\end{tikzpicture} \,\,,$ and $-\int d^{d_\text{in}} z_k \, \kappa$ for each $\,\,\begin{tikzpicture}[line width=1.0 pt,baseline=-1.5*\fontsizeshift]
		\coordinate (C) at (0.75,-0.5);
	\coordinate (x1) at (0,0);
	\coordinate (x2) at (1.5,0);
	\coordinate (x3) at (0,-1);
	\coordinate (x4) at (1.5,-1);
	\coordinate (x5) at (1.5,-0.5);
	\coordinate (x6) at (0,-0.5);
	\coordinate (ls) at (0,-.3);
	\draw[scs] (x1) -- (C);
	\draw[scs] (x2) -- (C);
	\draw[scs] (x3) -- (C);
	\draw[scs] (x4) -- (C);
	\draw[scs] (x5) -- (C);
	\draw[scs] (x6) -- (C);
	\draw[fill] (C) circle (.05);
	\node at ($(C) + 1.2*(ls)$) {$z_k$};
\end{tikzpicture} \,\,.$
  \item[6)] Write $K(u,v)$ for each $\,\, \begin{tikzpicture}[line width=1.0 pt,baseline=-2.8*\fontsizeshift]
		\coordinate (C) at (0.75,-1);
	\coordinate (x1) at (0,-1);
	\coordinate (ls) at (0,-.3);
	\draw[scs] (x1) -- (C);
	\draw[fill] (x1) circle (.05);
	\draw[fill] (C) circle (.05);
	\node at ($(x1) + 1.2*(ls)$) {$u$};
	\node at ($(C) + 1.2*(ls)$) {$v$};
\end{tikzpicture}$ .
  \item[7)] Throw away any terms containing vacuum bubbles; see Footnote \ref{footnote:nobubbles}.
\end{itemize}
As a non-trivial check, after step $2)$ there are $6m$ $z_k$ loose
ends, $4l$ $y_j$ loose ends, and $n$ $x_i$
loose ends, and there are $(n+4l+6m-1)!!$ ways of connecting these in pairs, which
must be the sum of the coefficients of the topologically distinct diagrams,
including vacuum bubbles.

\bigskip

We \emph{strongly emphasize} that the cases relevant for our experiments
require an important modification to what we have 
presented thus far, which is correct when $S = \sgp + \Delta S$, and 
$\Delta S$ is comprised of only non-Gaussian corrections, i.e. $\sgp$ is the
only Gaussian part of the action. In that case the two-point function is 
\begin{equation}
G^{(2)}(x_1,x_2) = K(x_1,x_2) + \text{$\lambda$- and $\kappa$- corrections}.
\end{equation}
In particular, $K(x_1,x_2)$ is the analog of the free-theory propagator in 
QFT, and it is only the leading piece in the $2$-pt function, which receives
corrections from interactions.

In the architectures used in our experiments, however, we remind the
reader that the GP kernel
is the exact $2$-pt function for the NGP as well
\begin{equation}
  \label{eqn:exact2pt}
  G^{(2)}(x_1,x_2) = K(x_1,x_2),
\end{equation}
as experimentally demonstrated in Section \ref{sec:gpexperiments} and theoretically derived in
Appendix \ref{kernel_derivation_appendix}. In particular this means that $S\neq S_{GP} + \Delta S$, but instead 
$S = S_G + \Delta S$ for some other Gaussian action $S_G$, defined to be the 
action such that \eqref{eqn:exact2pt} holds. One can in principle compute $S_G$,
but a simpler modification that we employ is to modify the Feynman rules for
computing the $O(\lambda^l \kappa^m)$ correction to $G^{(n)}(x_1,\dots,x_n)$.
The complete new Feynman rules in this case are 
\begin{itemize}
    \item[1)] For each of the $n$ external points $x_i$, draw 
   $ \begin{tikzpicture}[line width=1.0 pt,baseline=-2.8*\fontsizeshift]
            \coordinate (C) at (0.75,-1);
      \coordinate (x1) at (0,-1);
      \coordinate (ls) at (0,-.3);
      \draw[scd] (x1) -- (C);
      \draw[fill] (x1) circle (.05);
      \node at ($(x1) + 1.2*(ls)$) {$x_i$};
  \end{tikzpicture} \,\,.$ 
    \item[2)] For each $y_j$, draw $\,\,\begin{tikzpicture}[line width=1.0 pt,baseline=-2.8*\fontsizeshift]
            \coordinate (C) at (0.75,-1);
      \coordinate (x1) at (0,-0.5);
      \coordinate (x2) at (1.5,-0.5);
      \coordinate (x5) at (0, -1.5);
      \coordinate (x6) at (1.5, -1.5);
      \coordinate (ls) at (0,-.3);
      \draw[scd] (x1) -- (C);
      \draw[scd] (x2) -- (C);
      \draw[scd] (x5) -- (C);
      \draw[scd] (x6) -- (C);
      \draw[fill] (C) circle (.05);
      \node at ($(C) + 1.2*(ls)$) {$y_j$};
  \end{tikzpicture} $ \,\,. For each $z_k$, draw $\,\,\begin{tikzpicture}[line width=1.0 pt,baseline=-1.5*\fontsizeshift]
            \coordinate (C) at (0.75,-0.5);
      \coordinate (x1) at (0,0);
      \coordinate (x2) at (1.5,0);
      \coordinate (x3) at (0,-1);
      \coordinate (x4) at (1.5,-1);
      \coordinate (x5) at (1.5,-0.5);
      \coordinate (x6) at (0,-0.5);
      \coordinate (ls) at (0,-.3);
      \draw[scd] (x1) -- (C);
      \draw[scd] (x2) -- (C);
      \draw[scd] (x3) -- (C);
      \draw[scd] (x4) -- (C);
      \draw[scd] (x5) -- (C);
      \draw[scd] (x6) -- (C);
      \draw[fill] (C) circle (.05);
      \node at ($(C) + 1.2*(ls)$) {$z_k$};
  \end{tikzpicture} \,\,.$
    \item[3)] Determine all ways to pair up the loose ends associated to $x_i$'s, $y_j$'s, and $z_k$'s. This will yield some number of topologically distinct diagrams. Draw them with dashed lines.
    \item[4)] Write a sum over the diagrams with an appropriate combinatoric factor out front, which is the number of ways to form that diagram. Each diagram corresponds to an analytic term in the sum.
    \item[4.5)] Throw away any diagram that has a component with a $\lambda$- or $\kappa$ correction to the $2$-pt function.  
    \item[5)] For each diagram, write $-\int d^{d_\text{in}} y_j \, \lambda$ for 
      each  $\,\,\begin{tikzpicture}[line width=1.0 pt,baseline=-2.8*\fontsizeshift]
            \coordinate (C) at (0.75,-1);
      \coordinate (x1) at (0,-0.5);
      \coordinate (x2) at (1.5,-0.5);
      \coordinate (x5) at (0, -1.5);
      \coordinate (x6) at (1.5, -1.5);
      \coordinate (ls) at (0,-.3);
      \draw[scd] (x1) -- (C);
      \draw[scd] (x2) -- (C);
      \draw[scd] (x5) -- (C);
      \draw[scd] (x6) -- (C);
      \draw[fill] (C) circle (.05);
      \node at ($(C) + 1.2*(ls)$) {$y_j$};
  \end{tikzpicture} \,\,,$ and $-\int d^{d_\text{in}} z_k \, \kappa$ for each $\,\,\begin{tikzpicture}[line width=1.0 pt,baseline=-1.5*\fontsizeshift]
            \coordinate (C) at (0.75,-0.5);
      \coordinate (x1) at (0,0);
      \coordinate (x2) at (1.5,0);
      \coordinate (x3) at (0,-1);
      \coordinate (x4) at (1.5,-1);
      \coordinate (x5) at (1.5,-0.5);
      \coordinate (x6) at (0,-0.5);
      \coordinate (ls) at (0,-.3);
      \draw[scd] (x1) -- (C);
      \draw[scd] (x2) -- (C);
      \draw[scd] (x3) -- (C);
      \draw[scd] (x4) -- (C);
      \draw[scd] (x5) -- (C);
      \draw[scd] (x6) -- (C);
      \draw[fill] (C) circle (.05);
      \node at ($(C) + 1.2*(ls)$) {$z_k$};
  \end{tikzpicture} \,\,.$
    \item[6)] Write $K(u,v)$ for each $\,\, \begin{tikzpicture}[line width=1.0 pt,baseline=-2.8*\fontsizeshift]
            \coordinate (C) at (0.75,-1);
      \coordinate (x1) at (0,-1);
      \coordinate (ls) at (0,-.3);
      \draw[scd] (x1) -- (C);
      \draw[fill] (x1) circle (.05);
      \draw[fill] (C) circle (.05);
      \node at ($(x1) + 1.2*(ls)$) {$u$};
      \node at ($(C) + 1.2*(ls)$) {$v$};
  \end{tikzpicture}$ .
    \item[7)] Throw away any terms containing vacuum bubbles.
  \end{itemize}
Despite having presented the complete list, note that the only changes are the dashed lines in steps $3$ and $7$, and
the new step $4.5$. 

Diagrams thrown out at that step should not be included in this case because in writing $K(x_1,x_2)$ for \begin{tikzpicture}[line width=1.0 pt,baseline=-3*\fontsizeshift]
		\coordinate (C) at (0.75,-1);
	\coordinate (x1) at (0,-1);
	\coordinate (ls) at (0,-.3);
	\draw[scd] (x1) -- (C);
	\draw[fill] (x1) circle (.05);
	\draw[fill] (C) circle (.05);
	\node at ($(x1) + 1.2*(ls)$) {$x_1$};
	\node at ($(C) + 1.2*(ls)$) {$x_2$};
\end{tikzpicture}
in Step $6$, the $\lambda$- and $\kappa$- corrections to the $2$-pt function are already included, since the exact
two-point function is the GP kernel.
An example of such a diagram is of the form
\begin{equation}
 \lambda \, \begin{tikzpicture}[line width=1.0 pt,baseline=-2.8*\fontsizeshift]
		\coordinate (C) at (0.75,-1);
	\coordinate (x1) at (0,-1);
	\coordinate (x2) at (1.5,-1);
	\coordinate (x5) at (0, -1.5);
	\coordinate (x6) at (1.5, -1.5);
	\coordinate (ls) at (0,-.3);
	\draw[scs] (x1) -- (C);
	\draw[scs] (x2) -- (C);
	\draw[scs] (x5) -- (x6);
	\draw[fill] (C) circle (.05);
	\draw[fill] (x1) circle (.05);
	\draw[fill] (x2) circle (.05);
	\draw[fill] (x5) circle (.05);
	\draw[fill] (x6) circle (.05);
	\node at ($(C) + 1.0*(ls)$) {$y$};
	\node at ($(x1) + 1.2*(ls)$) {};
	\node at ($(x2) + 1.2*(ls)$) {};
	\node at ($(x5) + 1.2*(ls)$) {};
	\node at ($(x6) + 1.2*(ls)$) {};
	\begin{scope}[shift={(0,.27)}]\draw[scs] (0.75,-1) circle (.26);\end{scope}
\end{tikzpicture}~~~~~~
\text{in}~~G^{(4)}(x_1,x_2, x_3, x_4)
\end{equation}
which is implicitly included in the term of the form
\begin{equation}
    \begin{tikzpicture}[line width=1.0 pt,baseline=-2.8*\fontsizeshift]
              \coordinate (C) at (0.75,-1);
       \coordinate (x1) at (0,-1);
       \coordinate (x2) at (1.5,-1);
       \coordinate (x5) at (0, -1.5);
       \coordinate (x6) at (1.5, -1.5);
       \coordinate (ls) at (0,-.3);
       \draw[scd] (x1) -- (x2);
              \draw[scd] (x5) -- (x6);
       \draw[fill] (x1) circle (.05);
       \draw[fill] (x2) circle (.05);
       \draw[fill] (x5) circle (.05);
       \draw[fill] (x6) circle (.05);
              \node at ($(x1) + 1.2*(ls)$) {};
       \node at ($(x2) + 1.2*(ls)$) {};
       \node at ($(x5) + 1.2*(ls)$) {};
       \node at ($(x6) + 1.2*(ls)$) {};
   \end{tikzpicture}~~~~~~
   \text{in}~~G^{(4)}(x_1,x_2, x_3, x_4)
   \end{equation}
when the exact $2$-pt function is used.
It is easy to identify similar diagrams that must be thrown away as they arise.
Of course, this type of modification arises for any NGP in which  \eqref{eqn:exact2pt} holds.
The $(n+4l+6m-1)!!$ combinatorics may still be used as a non-trivial check prior to discarding
diagrams in steps 4.5) and 7). Finally, if we draw \begin{tikzpicture}[line width=1.0 pt,baseline=-2.7*\fontsizeshift]
		\coordinate (C) at (0.75,-1);
	\coordinate (x1) at (0,-1);
	\coordinate (ls) at (0,-.3);
	\draw[scs] (x1) -- (C);
	\draw[fill] (x1) circle (.05);
	\draw[fill] (C) circle (.05);
		\end{tikzpicture} in a case
where $\eqref{eqn:exact2pt}$ holds, it means the propagator / kernel of $S_G$.

\subsection{Four-point and Six-point NGP Correlation Functions}

We now turn to the computation of the $4$-pt and $6$-pt functions,
since they will be studied in our experiments. We emphasize,
however, that the calculations here are valid for the stated
orders in $\lambda$ and $\kappa$ for broader classes of NGPs
associated with neural network architectures.

Since in our experiments the NGP two-point function $G^{(2)}$ is 
exactly the kernel in the GP limit, i.e. \eqref{eqn:exact2pt} holds, 
we are in the case where $S = S_G + \Delta S$.
Accordingly we use the second set of Feynman rules in which \begin{tikzpicture}[line width=1.0 pt,baseline=-2.7*\fontsizeshift]
		\coordinate (C) at (0.75,-1);
	\coordinate (x1) at (0,-1);
	\coordinate (ls) at (0,-.3);
	\draw[scs] (x1) -- (C);
	\draw[fill] (x1) circle (.05);
	\draw[fill] (C) circle (.05);
		\end{tikzpicture} 
represents the propagator of $S_G$ and \begin{tikzpicture}[line width=1.0 pt,baseline=-2.7*\fontsizeshift]
		\coordinate (C) at (0.75,-1);
	\coordinate (x1) at (0,-1);
	\coordinate (ls) at (0,-.3);
	\draw[scd] (x1) -- (C);
	\draw[fill] (x1) circle (.05);
	\draw[fill] (C) circle (.05);
		\end{tikzpicture} represents the exact two-point 
function. We may represent the two-point function in perturbation theory as
\begin{eqnarray}
\label{2ptngp}
G^{(2)}(x_1,x_2) &=& 
\begin{tikzpicture}[line width=1.0 pt,baseline=-0.5*\fontsizeshift]
		\coordinate (x1) at (0,0);
	\coordinate (x2) at (2,0);
	\draw[scs] (x1) -- (x2);
	\draw[fill] (x1) circle (.05);
	\draw[fill] (x2) circle (.05);
\end{tikzpicture} \,\,
- \lambda \Bigg[ 12\,
\begin{tikzpicture}[line width=1.0 pt,baseline=-3*\fontsizeshift]
		\coordinate (C) at (1.5,-1);
	\coordinate (x1) at (0,-1);
	\coordinate (x2) at (3,-1);
	\coordinate (ys) at (0,-.3);
	\draw[scs] (x1) -- (C);
	\draw[scs] (C) -- (x2);
	\draw[fill] (x1) circle (.05);
	\draw[fill] (x2) circle (.05);
	\draw[fill] (C) circle (.05);
		\begin{scope}[shift={(0,.4)}]\draw[scs] (1.5,-1) circle (.4);\end{scope}
	\node at ($(C) + (ys)$) {$y$};
	\node at ($(x1) + (ys)$) {$x_1$};
	\node at ($(x2) + (ys)$) {$x_2$};
\end{tikzpicture} \Bigg]\, - \kappa \Bigg[ 90\,
\begin{tikzpicture}[line width=1.0 pt,baseline=-3*\fontsizeshift]
		\coordinate (C) at (2,-1);
	\coordinate (x1) at (0,-1);
	\coordinate (x2) at (4,-1);
	\coordinate (ls) at (0,-.3);
	\draw[scs] (x1) -- (C);
	\draw[scs] (C) -- (x2);
	\draw[fill] (x1) circle (.05);
	\draw[fill] (x2) circle (.05);
	\draw[fill] (C) circle (.05);
	\begin{scope}[shift={(0,.41)}]\draw[scs] (2,-1) circle (.4);\end{scope}
	\begin{scope}[shift={(0,-.4)}]\draw[scs] (2,-1) circle (.4);\end{scope}
	\node at ($(C) + 1.2*(ls)$) {$z$};
	\node at ($(x1) + 1.2*(ls)$) {$x_1$};
	\node at ($(x2) + 1.2*(ls)$) {$x_2$};
\end{tikzpicture} \Bigg] \nonumber \\[10pt]
&=& \begin{tikzpicture}[line width=1.0 pt,baseline=-0.5*\fontsizeshift]
		\coordinate (x1) at (0,0);
	\coordinate (x2) at (2,0);
	\draw[scd] (x1) -- (x2);
	\draw[fill] (x1) circle (.05);
	\draw[fill] (x2) circle (.05);
\end{tikzpicture} \nonumber \\[10pt]  &=& \,\, K(x_1,x_2),
\end{eqnarray} 
where the second diagrammatic equation represents the analytic expression
via the second set of Feynman rules, critically relying on step $4.5)$.

The $4$-pt and $6$-pt functions may be computed similarly. To simplify the
Feynman diagrams we add the additional rule that we do not label external
points, which is to be interpreted as summing over all combinations of external
points. The $4$-pt function is
\begin{eqnarray}  \label{full4-ptdiagrams}
G^{(4)}(x_1,x_2, x_3, x_4) &=&  3\, 
\begin{tikzpicture}[line width=1.0 pt,baseline=-2*\fontsizeshift]
		\coordinate (C) at (0,0);
	\coordinate (x3) at (0,-0.5);
	\coordinate (x4) at (1,-0.5);
	\coordinate (x2) at (1,-1);
	\coordinate (x1) at (0,-1);
	\coordinate (ls) at (0,-.3);
	\draw[scs] (x1) -- (x2);
	\draw[scs] (x3) -- (x4);
		\draw[fill] (x1) circle (.05);
	\draw[fill] (x2) circle (.05);
	\draw[fill] (x3) circle (.05);
	\draw[fill] (x4) circle (.05);
			\node at ($(x1) + 1.2*(ls)$) {};
	\node at ($(x2) + 1.2*(ls)$) {};
	\node at ($(x3) + 1.2*(ls)$) {};
	\node at ($(x4) + 1.2*(ls)$) {};
\end{tikzpicture}\, - \lambda \, \Bigg[\,
72\,\, \begin{tikzpicture}[line width=1.0 pt,baseline=-3*\fontsizeshift]
		\coordinate (C) at (0.75,-1);
	\coordinate (x1) at (0,-1);
	\coordinate (x2) at (1.5,-1);
	\coordinate (x5) at (0, -1.5);
	\coordinate (x6) at (1.5, -1.5);
	\coordinate (ls) at (0,-.3);
	\draw[scs] (x1) -- (C);
	\draw[scs] (x2) -- (C);
	\draw[scs] (x5) -- (x6);
	\draw[fill] (C) circle (.05);
	\draw[fill] (x1) circle (.05);
	\draw[fill] (x2) circle (.05);
	\draw[fill] (x5) circle (.05);
	\draw[fill] (x6) circle (.05);
	\node at ($(C) + 1*(ls)$) {$y$};
	\node at ($(x1) + 1.2*(ls)$) {};
	\node at ($(x2) + 1.2*(ls)$) {};
	\node at ($(x5) + 1.2*(ls)$) {};
	\node at ($(x6) + 1.2*(ls)$) {};
	\begin{scope}[shift={(0,.27)}]\draw[scs] (0.75,-1) circle (.26);\end{scope}
\end{tikzpicture}\,
+ 24\,\,\, \begin{tikzpicture}[line width=1.0 pt,baseline=-3*\fontsizeshift]
		\coordinate (C) at (0.75,-1);
	\coordinate (x1) at (0,-0.5);
	\coordinate (x2) at (1.5,-0.5);
	\coordinate (x5) at (0, -1.5);
	\coordinate (x6) at (1.5, -1.5);
	\coordinate (ls) at (0,-.3);
	\draw[scs] (x1) -- (C);
	\draw[scs] (x2) -- (C);
	\draw[scs] (x5) -- (C);
	\draw[scs] (x6) -- (C);
	\draw[fill] (C) circle (.05);
	\draw[fill] (x1) circle (.05);
	\draw[fill] (x2) circle (.05);
	\draw[fill] (x5) circle (.05);
	\draw[fill] (x6) circle (.05);
					\end{tikzpicture} \Bigg] \, \nonumber \\
&-& \kappa \, \Bigg[ \,
540\,\, \begin{tikzpicture}[line width=1.0 pt,baseline=-1.5*\fontsizeshift]
		\coordinate (C) at (0.75,0);
	\coordinate (x1) at (0,0);
	\coordinate (x2) at (1.5,0);
	\coordinate (x3) at (0,-1);
	\coordinate (x4) at (1.5,-1);
	\coordinate (ls) at (0,-.3);
	\draw[scs] (x1) -- (C);
	\draw[scs] (C) -- (x2);
	\draw[scs] (x3) -- (x4);
	\draw[fill] (x1) circle (.05);
	\draw[fill] (x2) circle (.05);
	\draw[fill] (x3) circle (.05);
	\draw[fill] (x4) circle (.05);
	\draw[fill] (C) circle (.05);
	\begin{scope}[shift={(0,.26)}]\draw[scs] (0.75,0) circle (.26);\end{scope}
	\begin{scope}[shift={(0,-.26)}]\draw[scs] (0.75,0) circle (.26);\end{scope}
	\node at ($(C) + 1*(ls)$) {$z$};
				\end{tikzpicture} \, \,+ 360\,\, 
\begin{tikzpicture}[line width=1.0 pt,baseline=-1.5*\fontsizeshift]
		\coordinate (C) at (0.75,-0.5);
	\coordinate (x1) at (0,0);
	\coordinate (x2) at (0,-1);
	\coordinate (x3) at (1.5,0);
	\coordinate (x4) at (1.5,-1);
	\coordinate (ls) at (0,-.3);
	\draw[scs] (x1) -- (C); 
	\draw[scs] (x2) -- (C);
	\draw[scs] (x3) -- (C);
	\draw[scs] (x4) -- (C);
	\draw[fill] (C) circle (.05);
	\draw[fill] (x1) circle (.05);
	\draw[fill] (x2) circle (.05);
	\draw[fill] (x3) circle (.05);
	\draw[fill] (x4) circle (.05);
	\node at ($(C) + 1.2*(ls)$) {$z$};
					\begin{scope}[shift={(0,.34)}]\draw[scs] (0.75,-0.5) circle (.3);\end{scope}
\end{tikzpicture} \Bigg]  \nonumber \\[10pt]
&=& 3\, 
\begin{tikzpicture}[line width=1.0 pt,baseline=-2*\fontsizeshift]
		\coordinate (C) at (0,0);
	\coordinate (x3) at (0,-0.5); 
	\coordinate (x4) at (1,-0.5);
	\coordinate (x2) at (1,-1);
	\coordinate (x1) at (0,-1);
	\coordinate (ls) at (0,-.3);
	\draw[scd] (x1) -- (x2);
	\draw[scd] (x3) -- (x4);
		\draw[fill] (x1) circle (.05);
	\draw[fill] (x2) circle (.05);
	\draw[fill] (x3) circle (.05);
	\draw[fill] (x4) circle (.05);
			\node at ($(x1) + 1.2*(ls)$) {};
	\node at ($(x2) + 1.2*(ls)$) {};
	\node at ($(x3) + 1.2*(ls)$) {};
	\node at ($(x4) + 1.2*(ls)$) {};
\end{tikzpicture}
- 24\,\,\lambda \,\, \begin{tikzpicture}[line width=1.0 pt,baseline=-2.8*\fontsizeshift]
		\coordinate (C) at (0.75,-1);
	\coordinate (x1) at (0,-0.5);
	\coordinate (x2) at (1.5,-0.5);
	\coordinate (x5) at (0, -1.5);
	\coordinate (x6) at (1.5, -1.5);
	\coordinate (ls) at (0,-.3);
	\draw[scd] (x1) -- (C);
	\draw[scd] (x2) -- (C);
	\draw[scd] (x5) -- (C);
	\draw[scd] (x6) -- (C);
	\draw[fill] (C) circle (.05);
	\draw[fill] (x1) circle (.05);
	\draw[fill] (x2) circle (.05);
	\draw[fill] (x5) circle (.05);
	\draw[fill] (x6) circle (.05);
	\node at ($(C) + 1.2*(ls)$) {$y$};
	\node at ($(x1) + 1.2*(ls)$) {};
	\node at ($(x2) + 1.2*(ls)$) {};
	\node at ($(x5) + 1.2*(ls)$) {};
	\node at ($(x6) + 1.2*(ls)$) {};
\end{tikzpicture}\,\,
-  360\,\, \kappa \,\,
\begin{tikzpicture}[line width=1.0 pt,baseline=-1.7*\fontsizeshift]
		\coordinate (C) at (0.75,-0.5);
	\coordinate (x1) at (0,0);
	\coordinate (x2) at (0,-1);
	\coordinate (x3) at (1.5,0);
	\coordinate (x4) at (1.5,-1);
	\coordinate (ls) at (0,-.3);
	\draw[scd] (x1) -- (C);
	\draw[scd] (x2) -- (C);
	\draw[scd] (x3) -- (C);
	\draw[scd] (x4) -- (C);
	\draw[fill] (C) circle (.05);
	\draw[fill] (x1) circle (.05);
	\draw[fill] (x2) circle (.05);
	\draw[fill] (x3) circle (.05);
	\draw[fill] (x4) circle (.05);
	\node at ($(C) + 1.2*(ls)$) {$z$};
					\begin{scope}[shift={(0,.34)}]\draw[scd] (0.75,-0.5) circle (.3);\end{scope}
\end{tikzpicture} \nonumber \\[10pt]
&=& K(x_1,x_2)K(x_3,x_4) + K(x_1,x_3)K(x_2,x_4) + K(x_1,x_4)K(x_2,x_3) \nonumber \\[10pt]
&-& 24 \,  \int d^{d_\text{in}}y\, \lambda \, K(x_1,y)K(x_2,y)K(x_3,y)K(x_4,y) \nonumber \\[10pt]
&-& 360 \, \int d^{d_\text{in}}z\, \kappa \, K(x_1,z)K(x_2,z)K(x_3,z)K(x_4,z)K(z,z) 
\end{eqnarray}
and the $6$-pt function is
\begin{eqnarray} \label{6-ptlambda}
G^{(6)}(x_1,x_2, x_3, x_4, x_5, x_6) &=& 15\,\, 
\begin{tikzpicture}[line width=1.0 pt,baseline=-1.6*\fontsizeshift]
		\coordinate (C) at (0,0);
	\coordinate (x3) at (0,-0.5);
	\coordinate (x5) at (0,-1);
	\coordinate (x4) at (1,-0.5);
	\coordinate (x6) at (1,-1);
	\coordinate (x2) at (1,0);
	\coordinate (x1) at (0,0);
	\coordinate (ls) at (0,-.3);
	\draw[scs] (x1) -- (x2);
	\draw[scs] (x3) -- (x4);
	\draw[scs] (x5) -- (x6);
		\draw[fill] (x1) circle (.05);
	\draw[fill] (x2) circle (.05);
	\draw[fill] (x3) circle (.05);
	\draw[fill] (x4) circle (.05);
	\draw[fill] (x5) circle (.05);
	\draw[fill] (x6) circle (.05);
			\node at ($(x1) + 1.2*(ls)$) {};
	\node at ($(x2) + 1.2*(ls)$) {};
	\node at ($(x3) + 1.2*(ls)$) {};
	\node at ($(x4) + 1.2*(ls)$) {};
	\node at ($(x5) + 1.2*(ls)$) {};
	\node at ($(x6) + 1.2*(ls)$) {};
\end{tikzpicture}\, - \lambda \, \Bigg[\,
540\, \, \begin{tikzpicture}[line width=1.0 pt,baseline=-3.75*\fontsizeshift]
		\coordinate (C) at (0.75,-1);
	\coordinate (x1) at (0,-1);
	\coordinate (x2) at (1.5,-1);
	\coordinate (x5) at (0, -1.5);
	\coordinate (x6) at (1.5, -1.5);
	\coordinate (x3) at (0, -2);
	\coordinate (x4) at (1.5, -2);
	\coordinate (ls) at (0,-.3);
	\draw[scs] (x1) -- (C);
	\draw[scs] (x2) -- (C);
	\draw[scs] (x5) -- (x6);
	\draw[scs] (x3) -- (x4);
	\draw[fill] (C) circle (.05);
	\draw[fill] (x1) circle (.05);
	\draw[fill] (x2) circle (.05);
	\draw[fill] (x5) circle (.05);
	\draw[fill] (x6) circle (.05);
	\draw[fill] (x3) circle (.05);
	\draw[fill] (x4) circle (.05);
	\node at ($(C) + 1*(ls)$) {$y$};
	\node at ($(x1) + 1.2*(ls)$) {};
	\node at ($(x2) + 1.2*(ls)$) {};
	\node at ($(x5) + 1.2*(ls)$) {};
	\node at ($(x6) + 1.2*(ls)$) {};
	\node at ($(x4) + 1.2*(ls)$) {};
	\node at ($(x3) + 1.2*(ls)$) {};
	\begin{scope}[shift={(0,.27)}]\draw[scs] (0.75,-1) circle (.26);\end{scope}
\end{tikzpicture}\,
+ 360\,\,\, \begin{tikzpicture}[line width=1.0 pt,baseline=-3.75*\fontsizeshift]
		\coordinate (C) at (0.75,-1);
	\coordinate (x1) at (0,-0.5);
	\coordinate (x2) at (1.5,-0.5);
	\coordinate (x5) at (0, -1.5);
	\coordinate (x6) at (1.5, -1.5);
	\coordinate (x3) at (0, -2);
	\coordinate (x4) at (1.5, -2);
	\coordinate (ls) at (0,-.3);
	\draw[scs] (x1) -- (C);
	\draw[scs] (x2) -- (C);
	\draw[scs] (x5) -- (C);
	\draw[scs] (x6) -- (C);
	\draw[scs] (x3) -- (x4);
	\draw[fill] (C) circle (.05);
	\draw[fill] (x1) circle (.05);
	\draw[fill] (x2) circle (.05);
	\draw[fill] (x5) circle (.05);
	\draw[fill] (x6) circle (.05);
	\draw[fill] (x3) circle (.05);
	\draw[fill] (x4) circle (.05);
	\node at ($(C) + 1.2*(ls)$) {$y$};
	\node at ($(x1) + 1.2*(ls)$) {};
	\node at ($(x2) + 1.2*(ls)$) {};
	\node at ($(x5) + 1.2*(ls)$) {};
	\node at ($(x6) + 1.2*(ls)$) {};
	\node at ($(x4) + 1.2*(ls)$) {};
	\node at ($(x3) + 1.2*(ls)$) {};
\end{tikzpicture}\,\, \Bigg] \nonumber \\
&-&\kappa\,\Bigg[ \, 720\,\,
\begin{tikzpicture}[line width=1.0 pt,baseline=-1.6*\fontsizeshift]
		\coordinate (C) at (0.75,-0.5);
	\coordinate (x1) at (0,0);
	\coordinate (x2) at (1.5,0);
	\coordinate (x3) at (0,-1);
	\coordinate (x4) at (1.5,-1);
	\coordinate (x5) at (1.5,-0.5);
	\coordinate (x6) at (0,-0.5);
	\coordinate (ls) at (0,-.3);
	\draw[scs] (x1) -- (C);
	\draw[scs] (x2) -- (C);
	\draw[scs] (x3) -- (C);
	\draw[scs] (x4) -- (C);
	\draw[scs] (x5) -- (C);
	\draw[scs] (x6) -- (C);
	\draw[fill] (C) circle (.05);
	\draw[fill] (x1) circle (.05);
	\draw[fill] (x2) circle (.05);
	\draw[fill] (x3) circle (.05);
	\draw[fill] (x4) circle (.05);
	\draw[fill] (x5) circle (.05);
	\draw[fill] (x6) circle (.05);
		\node at ($(C) + 1.2*(ls)$) {$z$};
	\node at ($(x1) + 1.2*(ls)$) {};
	\node at ($(x2) + 1.2*(ls)$) {};
	\node at ($(x3) + 1.2*(ls)$) {};
	\node at ($(x4) + 1.2*(ls)$) {};
	\node at ($(x5) + 1.2*(ls)$) {};
	\node at ($(x6) + 1.2*(ls)$) {};
\end{tikzpicture}\, + 5400\,\,
\begin{tikzpicture}[line width=1.0 pt,baseline=-2.6*\fontsizeshift]
		\coordinate (C) at (0.75,-0.5);
	\coordinate (x1) at (0,0);
	\coordinate (x2) at (0,-1);
	\coordinate (x3) at (1.5,0);
	\coordinate (x4) at (1.5,-1);
	\coordinate (ls) at (0,-.3);
	\coordinate (x5) at (0,-1.5);
	\coordinate (x6) at (1.5,-1.5);
	\draw[scs] (x1) -- (C);
	\draw[scs] (x2) -- (C);
	\draw[scs] (x3) -- (C);
	\draw[scs] (x4) -- (C);
	\draw[scs] (x5) -- (x6);
	\draw[fill] (C) circle (.05);
	\draw[fill] (x1) circle (.05);
	\draw[fill] (x2) circle (.05);
	\draw[fill] (x3) circle (.05);
	\draw[fill] (x4) circle (.05);
	\draw[fill] (x5) circle (.05);
	\draw[fill] (x6) circle (.05);
	\node at ($(C) + 1.2*(ls)$) {$z$};
	\node at ($(x1) + 1.2*(ls)$) {};
	\node at ($(x2) + 1.2*(ls)$) {};
	\node at ($(x3) + 1.2*(ls)$) {};
	\node at ($(x4) + 1.2*(ls)$) {};
	\node at ($(x5) + 1.2*(ls)$) {};
	\node at ($(x6) + 1.2*(ls)$) {};
	\begin{scope}[shift={(0,.34)}]\draw[scs] (0.75,-0.5) circle (.3);\end{scope}
\end{tikzpicture}\,
+ 4050\,\,\, \begin{tikzpicture}[line width=1.0 pt,baseline=-2.3*\fontsizeshift]
		\coordinate (C) at (0.75,0);
	\coordinate (x1) at (0,0);
	\coordinate (x2) at (1.5,0);
	\coordinate (x3) at (0,-1);
	\coordinate (x4) at (1.5,-1);
	\coordinate (x5) at (0,-1.5);
	\coordinate (x6) at (1.5,-1.5);
	\coordinate (ls) at (0,-.3);
	\draw[scs] (x1) -- (C);
	\draw[scs] (C) -- (x2);
	\draw[scs] (x3) -- (x4);
	\draw[scs] (x5) -- (x6);
	\draw[fill] (x1) circle (.05);
	\draw[fill] (x2) circle (.05);
	\draw[fill] (x3) circle (.05);
	\draw[fill] (x4) circle (.05);
	\draw[fill] (x5) circle (.05);
	\draw[fill] (x6) circle (.05);
	\draw[fill] (C) circle (.05);
	\begin{scope}[shift={(0,.26)}]\draw[scs] (0.75,0) circle (.26);\end{scope}
	\begin{scope}[shift={(0,-.26)}]\draw[scs] (0.75,0) circle (.26);\end{scope}
	\node at ($(C) + 1*(ls)$) {$z$};
	\node at ($(x1) + 1.2*(ls)$) {};
	\node at ($(x2) + 1.2*(ls)$) {};
	\node at ($(x3) + 1.2*(ls)$) {};
	\node at ($(x4) + 1.2*(ls)$) {};
	\node at ($(x5) + 1.2*(ls)$) {};
	\node at ($(x6) + 1.2*(ls)$) {};
\end{tikzpicture}\,\, \Bigg] \nonumber \\
= 15\,\, 
\begin{tikzpicture}[line width=1.0 pt,baseline=-1.6*\fontsizeshift]
		\coordinate (C) at (0,0);
	\coordinate (x3) at (0,-0.5);
	\coordinate (x5) at (0,-1);
	\coordinate (x4) at (1,-0.5);
	\coordinate (x6) at (1,-1);
	\coordinate (x2) at (1,0);
	\coordinate (x1) at (0,0);
	\coordinate (ls) at (0,-.3);
	\draw[scd] (x1) -- (x2);
	\draw[scd] (x3) -- (x4);
	\draw[scd] (x5) -- (x6);
		\draw[fill] (x1) circle (.05);
	\draw[fill] (x2) circle (.05);
	\draw[fill] (x3) circle (.05);
	\draw[fill] (x4) circle (.05);
	\draw[fill] (x5) circle (.05);
	\draw[fill] (x6) circle (.05);
			\node at ($(x1) + 1.2*(ls)$) {};
	\node at ($(x2) + 1.2*(ls)$) {};
	\node at ($(x3) + 1.2*(ls)$) {};
	\node at ($(x4) + 1.2*(ls)$) {};
	\node at ($(x5) + 1.2*(ls)$) {};
	\node at ($(x6) + 1.2*(ls)$) {};
\end{tikzpicture}\,
&-& 360\, \lambda \,\, \begin{tikzpicture}[line width=1.0 pt,baseline=-3.5*\fontsizeshift]
		\coordinate (C) at (0.75,-1);
	\coordinate (x1) at (0,-0.5);
	\coordinate (x2) at (1.5,-0.5);
	\coordinate (x5) at (0, -1.5);
	\coordinate (x6) at (1.5, -1.5);
	\coordinate (x3) at (0, -2);
	\coordinate (x4) at (1.5, -2);
	\coordinate (ls) at (0,-.3);
	\draw[scd] (x1) -- (C);
	\draw[scd] (x2) -- (C);
	\draw[scd] (x5) -- (C);
	\draw[scd] (x6) -- (C);
	\draw[scd] (x3) -- (x4);
	\draw[fill] (C) circle (.05);
	\draw[fill] (x1) circle (.05);
	\draw[fill] (x2) circle (.05);
	\draw[fill] (x5) circle (.05);
	\draw[fill] (x6) circle (.05);
	\draw[fill] (x3) circle (.05);
	\draw[fill] (x4) circle (.05);
	\node at ($(C) + 1.2*(ls)$) {$y$};
	\node at ($(x1) + 1.2*(ls)$) {};
	\node at ($(x2) + 1.2*(ls)$) {};
	\node at ($(x5) + 1.2*(ls)$) {};
	\node at ($(x6) + 1.2*(ls)$) {};
	\node at ($(x4) + 1.2*(ls)$) {};
	\node at ($(x3) + 1.2*(ls)$) {};
\end{tikzpicture}\,
-\kappa\,\Bigg[ \, 720\,\,
\begin{tikzpicture}[line width=1.0 pt,baseline=-1.6*\fontsizeshift]
		\coordinate (C) at (0.75,-0.5);
	\coordinate (x1) at (0,0);
	\coordinate (x2) at (1.5,0);
	\coordinate (x3) at (0,-1);
	\coordinate (x4) at (1.5,-1);
	\coordinate (x5) at (1.5,-0.5);
	\coordinate (x6) at (0,-0.5);
	\coordinate (ls) at (0,-.3);
	\draw[scd] (x1) -- (C);
	\draw[scd] (x2) -- (C);
	\draw[scd] (x3) -- (C);
	\draw[scd] (x4) -- (C);
	\draw[scd] (x5) -- (C);
	\draw[scd] (x6) -- (C);
	\draw[fill] (C) circle (.05);
	\draw[fill] (x1) circle (.05);
	\draw[fill] (x2) circle (.05);
	\draw[fill] (x3) circle (.05);
	\draw[fill] (x4) circle (.05);
	\draw[fill] (x5) circle (.05);
	\draw[fill] (x6) circle (.05);
		\node at ($(C) + 1.2*(ls)$) {$z$};
	\node at ($(x1) + 1.2*(ls)$) {};
	\node at ($(x2) + 1.2*(ls)$) {};
	\node at ($(x3) + 1.2*(ls)$) {};
	\node at ($(x4) + 1.2*(ls)$) {};
	\node at ($(x5) + 1.2*(ls)$) {};
	\node at ($(x6) + 1.2*(ls)$) {};
\end{tikzpicture}\, + 5400\,\,
\begin{tikzpicture}[line width=1.0 pt,baseline=-2*\fontsizeshift]
		\coordinate (C) at (0.75,-0.5);
	\coordinate (x1) at (0,0);
	\coordinate (x2) at (0,-1);
	\coordinate (x3) at (1.5,0);
	\coordinate (x4) at (1.5,-1);
	\coordinate (ls) at (0,-.3);
	\coordinate (x5) at (0,-1.5);
	\coordinate (x6) at (1.5,-1.5);
	\draw[scd] (x1) -- (C);
	\draw[scd] (x2) -- (C);
	\draw[scd] (x3) -- (C);
	\draw[scd] (x4) -- (C);
	\draw[scd] (x5) -- (x6);
	\draw[fill] (C) circle (.05);
	\draw[fill] (x1) circle (.05);
	\draw[fill] (x2) circle (.05);
	\draw[fill] (x3) circle (.05);
	\draw[fill] (x4) circle (.05);
	\draw[fill] (x5) circle (.05);
	\draw[fill] (x6) circle (.05);
	\node at ($(C) + 1.2*(ls)$) {$z$};
	\node at ($(x1) + 1.2*(ls)$) {};
	\node at ($(x2) + 1.2*(ls)$) {};
	\node at ($(x3) + 1.2*(ls)$) {};
	\node at ($(x4) + 1.2*(ls)$) {};
	\node at ($(x5) + 1.2*(ls)$) {};
	\node at ($(x6) + 1.2*(ls)$) {};
	\begin{scope}[shift={(0,.34)}]\draw[scd] (0.75,-0.5) circle (.3);\end{scope}
\end{tikzpicture}\,\Bigg] \nonumber 
\end{eqnarray}
\begin{eqnarray}
&=&  \bigg[ K_{12} K_{34} K_{56} +  K_{12} K_{35} K_{46} + K_{12} K_{36} K_{45} + K_{13} K_{24} K_{56} + K_{13} K_{25} K_{46} + K_{13} K_{26} K_{45} + K_{14} K_{23} K_{56} \nonumber \\ 
&+& K_{14} K_{25} K_{36} + K_{14} K_{26} K_{35} + K_{15} K_{23} K_{46} + K_{15} K_{24} K_{36} + K_{15} K_{26} K_{34}  + K_{16} K_{23} K_{45} + K_{16} K_{24} K_{35} \nonumber \\ &+& K_{16} K_{25} K_{34} \bigg] - 24 \, \int d^{d_\text{in}}y\,\lambda\, \bigg[  K_{1y}K_{2y}K_{3y}K_{4y} K_{56} + K_{1y}K_{2y}K_{3y}K_{5y} K_{46} + K_{1y}K_{2y}K_{4y}K_{5y} K_{36} 
\nonumber \\ &+& K_{1y}K_{3y}K_{4y}K_{5y} K_{26} + K_{2y}K_{3y}K_{4y}K_{5y} K_{16} + K_{1y}K_{2y}K_{3y}K_{6y} K_{45} + K_{1y}K_{2y}K_{4y}K_{6y} K_{35} \nonumber \\ &+& K_{1y}K_{3y}K_{4y}K_{6y} K_{25} 
+ K_{2y}K_{3y}K_{4y}K_{6y} K_{15} + K_{1y}K_{2y}K_{5y}K_{6y} K_{34} + K_{1y}K_{3y}K_{5y}K_{6y} K_{24} \nonumber \\ &+& K_{2y}K_{3y}K_{5y}K_{6y} K_{14} + K_{1y}K_{4y}K_{5y}K_{6y} K_{23} 
+ K_{2y}K_{4y}K_{5y}K_{6y} K_{13} + K_{3y}K_{4y}K_{5y}K_{6y} K_{12}\bigg] \nonumber
\\ &-& 720 \, \int d^{\din} z \, \kappa \, K_{1z} K_{2z} K_{3z} K_{4z} K_{5z} K_{6z} \nonumber
- 360 \, \int d^{\din} z \, \kappa \, \bigg[ K_{zz} K_{1z}K_{2z}K_{3z}K_{4z} K_{56} \\ \nonumber &+& K_{zz} K_{1z}K_{2z}K_{3z}K_{5z} K_{46} + K_{zz}K_{1z}K_{2z}K_{4z}K_{5z} K_{36} 
+ K_{zz}K_{1z}K_{3z}K_{4z}K_{5z} K_{26} \\ \nonumber &+& K_{zz}K_{2z}K_{3z}K_{4z}K_{5z} K_{16} + K_{zz}K_{1z}K_{2z}K_{3z}K_{6z} K_{45} + K_{zz}K_{1z}K_{2z}K_{4z}K_{6z} K_{35}  \\ \nonumber &+& K_{zz}K_{1z}K_{3z}K_{4z}K_{6z} K_{25} 
+ K_{zz}K_{2z}K_{3z}K_{4z}K_{6z} K_{15} + K_{zz}K_{1z}K_{2z}K_{5z}K_{6z} K_{34} \\ \nonumber &+& K_{zz}K_{1z}K_{3z}K_{5z}K_{6z} K_{24} + K_{zz}K_{2z}K_{3z}K_{5z}K_{6z} K_{14} + K_{zz}K_{1z}K_{4z}K_{5z}K_{6z} K_{23} 
\\ &+&  K_{zz}K_{2z}K_{4z}K_{5z}K_{6z} K_{13} + K_{zz}K_{3z}K_{4z}K_{5z}K_{6z} K_{12}\bigg],
\end{eqnarray}
where again the analytic expression corresponds to the second set of Feynman
diagrams according to the Feynman rules. Note the simplicity of the second diagrammatic representation relative to its
corresponding analytic expression.

\subsection{Neural Network Coupling Constants and GP Symmetries} \label{sec-technical-naturalness}

We have argued that non-Gaussian corrections are crucial in neural networks and they correspond
to turning on interactions from the QFT point of view. Such corrections arise as
terms in $\Delta S$, each with a coefficient encoding the interaction strength, 
such as $\lambda$ or $\kappa$ in \eqref{eq:eft-action}.

For convenience we have thus far ignored important questions about any interaction
coefficient appearing in $\Delta S$: are they constants? In QFT 
language such quantities are called coupling constants, but this is simply an artifact of the translation invariant QFTs that physicists most often study. For instance, in the Standard Model of particle 
physics a parameter analogous to $\lambda$ quantifies the self-interaction of the
Higgs boson, and all such coupling constants have historically proven to be invariant 
under spatial translations on Earth; they are independent of whether the scattering
experiment is done at CERN in Switzerland or Fermilab in Illinois. This follows from the
translation invariance of the entire Standard Model action, which therefore also predicts
that the same couplings would be measured in proton scattering experiments near Alpha Centauri.
That is, couplings are constants in spacetime.

But in NGPs associated with neural network architectures, do we expect\footnote{In our Feynman rules, we have left the coefficients $\lambda$
and $\kappa$ inside the integrals in anticipation of this question.} the coefficients
such as $\lambda$ and $\kappa$ to truly be constants? What role should translation invariance
play in such considerations? To this end we employ a powerful principle due to 't Hooft:
\vspace{.5cm}
\begin{center}
\begin{minipage}{30em}
	\textbf{Technical Naturalness:} a coupling $g$ appearing in $\Delta S$ may be
	small relative to $\Lambda$ if a symmetry is restored when $g$ is set to zero.
\end{minipage}
\end{center}
\vspace{.5cm}
A concept underlying technical naturalness is that $g$ itself may receive large corrections in varying the cutoff\footnote{That is, due to Wilsonian renormalization group flow, as we will exemplify in Section \ref{sec:rg}.} $\Lambda$, i.e. in
varying $\Lambda \to \Lambda'$,
\begin{equation}
	g \to g + \Delta g \,\,\,\, \text{with} \,\,\,\, \frac{\Delta g}{g} \gg 1.
\end{equation}
However,
in some cases it is possible to show that $\Delta g = 0$ if $S$ has a symmetry. If $g$ 
itself breaks the symmetry, but the symmetry is restored when $g\to 0$, then 
as $g\to 0$ we must have $\Delta g \to 0$ as well. In some concrete
cases, this is enforced by $\Delta g = g\, \alpha$ where $\alpha$ is some function
mild enough that $\Delta g \to 0$ as $g \to 0$. If $g$ is small, this ensures that 
corrections to
it are also small, so that $g$ is small for all values of the cutoff. In such a 
case, $g$ is said to be \emph{technically natural}. A simple example is the electron
mass $m_e$ in quantum electrodynamics: $m_e$ is small relative to the electroweak
scale and corrections
to the mass are $\Delta m_e\propto m_e$, ensuring that it remains small as the cutoff is
varied. As $m_e\to 0$, a symmetry of electrons called the ``chiral symmetry'' is restored. The small electron mass $m_e$
is technically natural.

Technical naturalness is directly applicable to our discussion of coupling
constants versus coupling functions in NGPs associated with neural networks. For
instance, is $\lambda$ a constant or a function of the input space? To examine this question
in light of technical naturalness, consider a simple case where $\lambda$ is the only
non-Gaussian coefficient and is decomposed as
\begin{equation}
\lambda(x) = \bar \lambda + \delta \lambda(x),
\end{equation}
with a constant piece $\bar \lambda$ plus some non-constant piece $\delta \lambda(x)$. The variance of $\lambda$ is determined by $\delta \lambda$, which in
general is not invariant under translations $T:x \to x + c$. When $\delta\lambda = 0$, $\lambda$
is a constant, and when it is small, $\lambda$ is effectively a constant. But should $\delta\lambda$ be small? This is where technical naturalness is useful. It states that 
\begin{equation}
\frac{\delta \lambda}{\lambda} \ll 1
\end{equation}
is reasonable to expect if there is a symmetry in the $\delta \lambda \to 0$ limit. Since
$\delta \lambda$ breaks $T$, a relevant question is whether $T$ is restored when $\delta \lambda \to 0$. This occurs when $\sgp$, and specifically its kernel $K(x,y)$, is $T$-invariant. In examples with multiple couplings, the relevant question is whether $T$ is restored
when all couplings go to zero, i.e. again whether or not $K(x,y)$ is $T$-invariant.

Technical naturalness therefore leads to the following concrete conjecture:
\begin{center}
	\vspace{.5cm}
	\begin{minipage}{30em}
		\textbf{Conjecture:} couplings in NGPs associated to neural network architectures
		are constants (or nearly constants) if the kernel $K(x,y)$ associated with
		their GP limit is translationally invariant.
	\end{minipage}
	\vspace{.5cm}
\end{center}
This is a conjecture rather than a theorem because technical naturalness is not proven
in general, but is instead a guiding principle in physics. In our experiments, however,
the conjecture can be tested: while the kernels associated with Erf-net and ReLU-net
are not $T$-invariant, the \gnet $\,$ kernel is $T$-invariant. We will verify the conjecture
in these examples by demonstrating that couplings are effectively constants in  \gnet.

\amnew{
\subsection{Independence in EFT of Single-Layer Networks}
\label{sec:exeft}
Recall that a neural network with linear output layer, such as our examples, is a process $\mathcal{P}$ from which $f = f_b + f_W$ is drawn; $f_b$ and $f_W$ are drawn from independent processes $f_b \sim \mathcal{P}_b$ and $f_W \sim \mathcal{P}_W$. Processes $\mathcal{P}_b$ are independent of width $N$ and always Gaussian, provided the biases are Gaussian; whereas $\mathcal{P}_W$ admits non-Gaussianities and $N$-dependence. Independence means that the distribution is a product, and therefore  the log-likelihood of the Gaussian process can be expressed as
\begin{eqnarray}
S_{GP} = S^b_{GP} + S^w_{GP} = \int d^{\din}x d^{\din}y \bigg[ f_b(x) \Xi_b(x,y) f_b(y) + f_W(x) \Xi_W(x,y) f_W(y) \bigg],
\end{eqnarray}
where 
\begin{eqnarray}
\int d^{d_\text{in}} x' \, K(x,x')_{W/b} \, \Xi(x',x'')_{W/b} = \delta^{(d_\text{in})}(x-x''),
\end{eqnarray}
$W/b$ denotes that the equation applies to both the weight and bias pieces,
and the neural network kernel is given by $K(x,y) = K_b(x,y) + K_W(x,y)$. Classical scaling dimensions of independent outputs $f_W$ and $f_b$ become
\begin{eqnarray}
[f_{W/b}] = - \frac{2\din + [\Xi_{W/b}] }{2},
\end{eqnarray}
which may also be expressed in terms of $[K_{W/b}]$.

Due to the independence of $f_b$ and $f_W$ in our examples,  the non-Gaussian correction to the log-likelihood of non-Gaussian process is $\Delta S = \Delta S_b + \Delta S_W$, where the first (second) term only depends on $f_b$ ($f_W$). Since in our examples the bias entries are Gaussian, $\Delta S_b = 0$. Therefore, $S = S_{GP} + \Delta S_W$, and a local $\Delta S_W$ is given by
\begin{equation}
\Delta S_W = \int  \, d^{\din}x \, \left[ g \, f_W(x)^3 + \lambda \, f_W(x)^4 + \alpha \, f_W(x)^5 + \kappa \, f_W(x)^6 + \dots \right].
\end{equation}
We will consider a simple non-locality in Section \ref{sec:eft-couplings}, and see that for the inputs we study the local approximation works well.
Classical scaling dimensions of general operators $\mathcal{O}_k := g_k f_W(x)^k$ is given by
\begin{eqnarray}
[g_k] = - \din + \frac{k(2\din + [\Xi_W])}{2} = - \din - \frac{k[K_W]}{2} .
\end{eqnarray}
If $[K_W]\geq 0$, operators $\mathcal{O}_k$ can be sufficiently ignored for large $k$, as all $g_k$ have scaling dimensions with same sign. Considering corrections to the $n$-pt function from $\Delta S_W = \int \din x g_k f_W(x)^k$ with $k > 2$, we have
\begin{eqnarray}
 G^{(n)}(x_{1},\dots, x_{n}) = \frac{\int df \,\, f(x_{1})\dots f(x_{n}) \left[1 - \int d^{d_\text{in}}x \, g_k f_W(x)^k + O(g_k^2)\right]\, e^{-\sgp }/\zgpzero}{\int df \,\, \left[1 - \int d^{d_\text{in}}x \, g_k f_W(x)^k + O(g_k^2)\right]e^{-\sgp }/\zgpzero},
\end{eqnarray}
which may be computed to any given order using Feynman diagrams.

The effective action utilized for our examples is 
\begin{eqnarray}
S = \sgp + \int \, d^{d_{\text{in}}} x \,\, \left[\lambda \, f_W(x)^4 + \kappa \, f_W(x)^6\right].
\end{eqnarray}
The Feynman rules remain unchanged, except now any propagator attached to internal vertices are those associated to $f_W$, e.g. 
$\begin{tikzpicture}[line width=1.0 pt,baseline=-2.8*\fontsizeshift]
            \coordinate (C) at (0.75,-1);
      \coordinate (x1) at (0,-1);
      \coordinate (ls) at (0,-.3);
      \draw[scd] (x1) -- (C);
      \draw[fill] (x1) circle (.05);
      \draw[fill] (C) circle (.05);
      \node at ($(x1) + 1.2*(ls)$) {$u$};
      \node at ($(C) + 1.2*(ls)$) {$v$};
  \end{tikzpicture}$ corresponding to $K_W(u,v)$.
With this, the $2$-pt remains unaltered; whereas $4$-pt and $6$-pt functions are respectively given by
\begin{eqnarray}  \label{exfull4-ptdiagrams}
G^{(4)}(x_1,x_2, x_3, x_4) &=&  3\, 
\begin{tikzpicture}[line width=1.0 pt,baseline=-2*\fontsizeshift]
		\coordinate (C) at (0,0);
	\coordinate (x3) at (0,-0.5);
	\coordinate (x4) at (1,-0.5);
	\coordinate (x2) at (1,-1);
	\coordinate (x1) at (0,-1);
	\coordinate (ls) at (0,-.3);
	\draw[scs] (x1) -- (x2);
	\draw[scs] (x3) -- (x4);
		\draw[fill] (x1) circle (.05);
	\draw[fill] (x2) circle (.05);
	\draw[fill] (x3) circle (.05);
	\draw[fill] (x4) circle (.05);
			\node at ($(x1) + 1.2*(ls)$) {};
	\node at ($(x2) + 1.2*(ls)$) {};
	\node at ($(x3) + 1.2*(ls)$) {};
	\node at ($(x4) + 1.2*(ls)$) {};
\end{tikzpicture}\, - \lambda \, \Bigg[\,
72\,\, \begin{tikzpicture}[line width=1.0 pt,baseline=-3*\fontsizeshift]
		\coordinate (C) at (0.75,-1);
	\coordinate (x1) at (0,-1);
	\coordinate (x2) at (1.5,-1);
	\coordinate (x5) at (0, -1.5);
	\coordinate (x6) at (1.5, -1.5);
	\coordinate (ls) at (0,-.3);
	\draw[scs] (x1) -- (C);
	\draw[scs] (x2) -- (C);
	\draw[scs] (x5) -- (x6);
	\draw[fill] (C) circle (.05);
	\draw[fill] (x1) circle (.05);
	\draw[fill] (x2) circle (.05);
	\draw[fill] (x5) circle (.05);
	\draw[fill] (x6) circle (.05);
	\node at ($(C) + 1*(ls)$) {$y$};
	\node at ($(x1) + 1.2*(ls)$) {};
	\node at ($(x2) + 1.2*(ls)$) {};
	\node at ($(x5) + 1.2*(ls)$) {};
	\node at ($(x6) + 1.2*(ls)$) {};
	\begin{scope}[shift={(0,.27)}]\draw[scs] (0.75,-1) circle (.26);\end{scope}
\end{tikzpicture}\,
+ 24\,\,\, \begin{tikzpicture}[line width=1.0 pt,baseline=-3*\fontsizeshift]
		\coordinate (C) at (0.75,-1);
	\coordinate (x1) at (0,-0.5);
	\coordinate (x2) at (1.5,-0.5);
	\coordinate (x5) at (0, -1.5);
	\coordinate (x6) at (1.5, -1.5);
	\coordinate (ls) at (0,-.3);
	\draw[scs] (x1) -- (C);
	\draw[scs] (x2) -- (C);
	\draw[scs] (x5) -- (C);
	\draw[scs] (x6) -- (C);
	\draw[fill] (C) circle (.05);
	\draw[fill] (x1) circle (.05);
	\draw[fill] (x2) circle (.05);
	\draw[fill] (x5) circle (.05);
	\draw[fill] (x6) circle (.05);
					\end{tikzpicture} \Bigg] \, \nonumber \\
&-& \kappa \, \Bigg[ \,
540\,\, \begin{tikzpicture}[line width=1.0 pt,baseline=-1.5*\fontsizeshift]
		\coordinate (C) at (0.75,0);
	\coordinate (x1) at (0,0);
	\coordinate (x2) at (1.5,0);
	\coordinate (x3) at (0,-1);
	\coordinate (x4) at (1.5,-1);
	\coordinate (ls) at (0,-.3);
	\draw[scs] (x1) -- (C);
	\draw[scs] (C) -- (x2);
	\draw[scs] (x3) -- (x4);
	\draw[fill] (x1) circle (.05);
	\draw[fill] (x2) circle (.05);
	\draw[fill] (x3) circle (.05);
	\draw[fill] (x4) circle (.05);
	\draw[fill] (C) circle (.05);
	\begin{scope}[shift={(0,.26)}]\draw[scs] (0.75,0) circle (.26);\end{scope}
	\begin{scope}[shift={(0,-.26)}]\draw[scs] (0.75,0) circle (.26);\end{scope}
	\node at ($(C) + 1*(ls)$) {$z$};
				\end{tikzpicture} \, \,+ 360\,\, 
\begin{tikzpicture}[line width=1.0 pt,baseline=-1.5*\fontsizeshift]
		\coordinate (C) at (0.75,-0.5);
	\coordinate (x1) at (0,0);
	\coordinate (x2) at (0,-1);
	\coordinate (x3) at (1.5,0);
	\coordinate (x4) at (1.5,-1);
	\coordinate (ls) at (0,-.3);
	\draw[scs] (x1) -- (C); 
	\draw[scs] (x2) -- (C);
	\draw[scs] (x3) -- (C);
	\draw[scs] (x4) -- (C);
	\draw[fill] (C) circle (.05);
	\draw[fill] (x1) circle (.05);
	\draw[fill] (x2) circle (.05);
	\draw[fill] (x3) circle (.05);
	\draw[fill] (x4) circle (.05);
	\node at ($(C) + 1.2*(ls)$) {$z$};
					\begin{scope}[shift={(0,.34)}]\draw[scs] (0.75,-0.5) circle (.3);\end{scope}
\end{tikzpicture} \Bigg]  \nonumber \\[10pt]
&=& 3\, 
\begin{tikzpicture}[line width=1.0 pt,baseline=-2*\fontsizeshift]
		\coordinate (C) at (0,0);
	\coordinate (x3) at (0,-0.5); 
	\coordinate (x4) at (1,-0.5);
	\coordinate (x2) at (1,-1);
	\coordinate (x1) at (0,-1);
	\coordinate (ls) at (0,-.3);
	\draw[scd] (x1) -- (x2);
	\draw[scd] (x3) -- (x4);
		\draw[fill] (x1) circle (.05);
	\draw[fill] (x2) circle (.05);
	\draw[fill] (x3) circle (.05);
	\draw[fill] (x4) circle (.05);
			\node at ($(x1) + 1.2*(ls)$) {};
	\node at ($(x2) + 1.2*(ls)$) {};
	\node at ($(x3) + 1.2*(ls)$) {};
	\node at ($(x4) + 1.2*(ls)$) {};
\end{tikzpicture}
- 24\,\,\lambda \,\, \begin{tikzpicture}[line width=1.0 pt,baseline=-2.8*\fontsizeshift]
		\coordinate (C) at (0.75,-1);
	\coordinate (x1) at (0,-0.5);
	\coordinate (x2) at (1.5,-0.5);
	\coordinate (x5) at (0, -1.5);
	\coordinate (x6) at (1.5, -1.5);
	\coordinate (ls) at (0,-.3);
	\draw[scd] (x1) -- (C);
	\draw[scd] (x2) -- (C);
	\draw[scd] (x5) -- (C);
	\draw[scd] (x6) -- (C);
	\draw[fill] (C) circle (.05);
	\draw[fill] (x1) circle (.05);
	\draw[fill] (x2) circle (.05);
	\draw[fill] (x5) circle (.05);
	\draw[fill] (x6) circle (.05);
	\node at ($(C) + 1.2*(ls)$) {$y$};
	\node at ($(x1) + 1.2*(ls)$) {};
	\node at ($(x2) + 1.2*(ls)$) {};
	\node at ($(x5) + 1.2*(ls)$) {};
	\node at ($(x6) + 1.2*(ls)$) {};
\end{tikzpicture}\,\,
-  360\,\, \kappa \,\,
\begin{tikzpicture}[line width=1.0 pt,baseline=-1.7*\fontsizeshift]
		\coordinate (C) at (0.75,-0.5);
	\coordinate (x1) at (0,0);
	\coordinate (x2) at (0,-1);
	\coordinate (x3) at (1.5,0);
	\coordinate (x4) at (1.5,-1);
	\coordinate (ls) at (0,-.3);
	\draw[scd] (x1) -- (C);
	\draw[scd] (x2) -- (C);
	\draw[scd] (x3) -- (C);
	\draw[scd] (x4) -- (C);
	\draw[fill] (C) circle (.05);
	\draw[fill] (x1) circle (.05);
	\draw[fill] (x2) circle (.05);
	\draw[fill] (x3) circle (.05);
	\draw[fill] (x4) circle (.05);
	\node at ($(C) + 1.2*(ls)$) {$z$};
					\begin{scope}[shift={(0,.34)}]\draw[scd] (0.75,-0.5) circle (.3);\end{scope}
\end{tikzpicture} \nonumber \\[10pt]
&=& K(x_1,x_2)K(x_3,x_4) + K(x_1,x_3)K(x_2,x_4) + K(x_1,x_4)K(x_2,x_3) \nonumber \\[10pt]
&-& 24 \,  \int d^{d_\text{in}}y\, \lambda \, K_W(x_1,y)K_W(x_2,y)K_W(x_3,y)K_W(x_4,y) \nonumber \\[10pt]
&-& 360 \, \int d^{d_\text{in}}z\, \kappa \, K_W(x_1,z)K_W(x_2,z)K_W(x_3,z)K_W(x_4,z)K_W(z,z) 
\end{eqnarray}
\begin{eqnarray} \label{ex6-ptlambda}
G^{(6)}(x_1,x_2, x_3, x_4, x_5, x_6) &=& 15\,\, 
\begin{tikzpicture}[line width=1.0 pt,baseline=-1.6*\fontsizeshift]
		\coordinate (C) at (0,0);
	\coordinate (x3) at (0,-0.5);
	\coordinate (x5) at (0,-1);
	\coordinate (x4) at (1,-0.5);
	\coordinate (x6) at (1,-1);
	\coordinate (x2) at (1,0);
	\coordinate (x1) at (0,0);
	\coordinate (ls) at (0,-.3);
	\draw[scs] (x1) -- (x2);
	\draw[scs] (x3) -- (x4);
	\draw[scs] (x5) -- (x6);
		\draw[fill] (x1) circle (.05);
	\draw[fill] (x2) circle (.05);
	\draw[fill] (x3) circle (.05);
	\draw[fill] (x4) circle (.05);
	\draw[fill] (x5) circle (.05);
	\draw[fill] (x6) circle (.05);
			\node at ($(x1) + 1.2*(ls)$) {};
	\node at ($(x2) + 1.2*(ls)$) {};
	\node at ($(x3) + 1.2*(ls)$) {};
	\node at ($(x4) + 1.2*(ls)$) {};
	\node at ($(x5) + 1.2*(ls)$) {};
	\node at ($(x6) + 1.2*(ls)$) {};
\end{tikzpicture}\, - \lambda \, \Bigg[\,
540\, \, \begin{tikzpicture}[line width=1.0 pt,baseline=-3.75*\fontsizeshift]
		\coordinate (C) at (0.75,-1);
	\coordinate (x1) at (0,-1);
	\coordinate (x2) at (1.5,-1);
	\coordinate (x5) at (0, -1.5);
	\coordinate (x6) at (1.5, -1.5);
	\coordinate (x3) at (0, -2);
	\coordinate (x4) at (1.5, -2);
	\coordinate (ls) at (0,-.3);
	\draw[scs] (x1) -- (C);
	\draw[scs] (x2) -- (C);
	\draw[scs] (x5) -- (x6);
	\draw[scs] (x3) -- (x4);
	\draw[fill] (C) circle (.05);
	\draw[fill] (x1) circle (.05);
	\draw[fill] (x2) circle (.05);
	\draw[fill] (x5) circle (.05);
	\draw[fill] (x6) circle (.05);
	\draw[fill] (x3) circle (.05);
	\draw[fill] (x4) circle (.05);
	\node at ($(C) + 1*(ls)$) {$y$};
	\node at ($(x1) + 1.2*(ls)$) {};
	\node at ($(x2) + 1.2*(ls)$) {};
	\node at ($(x5) + 1.2*(ls)$) {};
	\node at ($(x6) + 1.2*(ls)$) {};
	\node at ($(x4) + 1.2*(ls)$) {};
	\node at ($(x3) + 1.2*(ls)$) {};
	\begin{scope}[shift={(0,.27)}]\draw[scs] (0.75,-1) circle (.26);\end{scope}
\end{tikzpicture}\,
+ 360\,\,\, \begin{tikzpicture}[line width=1.0 pt,baseline=-3.75*\fontsizeshift]
		\coordinate (C) at (0.75,-1);
	\coordinate (x1) at (0,-0.5);
	\coordinate (x2) at (1.5,-0.5);
	\coordinate (x5) at (0, -1.5);
	\coordinate (x6) at (1.5, -1.5);
	\coordinate (x3) at (0, -2);
	\coordinate (x4) at (1.5, -2);
	\coordinate (ls) at (0,-.3);
	\draw[scs] (x1) -- (C);
	\draw[scs] (x2) -- (C);
	\draw[scs] (x5) -- (C);
	\draw[scs] (x6) -- (C);
	\draw[scs] (x3) -- (x4);
	\draw[fill] (C) circle (.05);
	\draw[fill] (x1) circle (.05);
	\draw[fill] (x2) circle (.05);
	\draw[fill] (x5) circle (.05);
	\draw[fill] (x6) circle (.05);
	\draw[fill] (x3) circle (.05);
	\draw[fill] (x4) circle (.05);
	\node at ($(C) + 1.2*(ls)$) {$y$};
	\node at ($(x1) + 1.2*(ls)$) {};
	\node at ($(x2) + 1.2*(ls)$) {};
	\node at ($(x5) + 1.2*(ls)$) {};
	\node at ($(x6) + 1.2*(ls)$) {};
	\node at ($(x4) + 1.2*(ls)$) {};
	\node at ($(x3) + 1.2*(ls)$) {};
\end{tikzpicture}\,\, \Bigg] \nonumber \\
&-&\kappa\,\Bigg[ \, 720\,\,
\begin{tikzpicture}[line width=1.0 pt,baseline=-1.6*\fontsizeshift]
		\coordinate (C) at (0.75,-0.5);
	\coordinate (x1) at (0,0);
	\coordinate (x2) at (1.5,0);
	\coordinate (x3) at (0,-1);
	\coordinate (x4) at (1.5,-1);
	\coordinate (x5) at (1.5,-0.5);
	\coordinate (x6) at (0,-0.5);
	\coordinate (ls) at (0,-.3);
	\draw[scs] (x1) -- (C);
	\draw[scs] (x2) -- (C);
	\draw[scs] (x3) -- (C);
	\draw[scs] (x4) -- (C);
	\draw[scs] (x5) -- (C);
	\draw[scs] (x6) -- (C);
	\draw[fill] (C) circle (.05);
	\draw[fill] (x1) circle (.05);
	\draw[fill] (x2) circle (.05);
	\draw[fill] (x3) circle (.05);
	\draw[fill] (x4) circle (.05);
	\draw[fill] (x5) circle (.05);
	\draw[fill] (x6) circle (.05);
		\node at ($(C) + 1.2*(ls)$) {$z$};
	\node at ($(x1) + 1.2*(ls)$) {};
	\node at ($(x2) + 1.2*(ls)$) {};
	\node at ($(x3) + 1.2*(ls)$) {};
	\node at ($(x4) + 1.2*(ls)$) {};
	\node at ($(x5) + 1.2*(ls)$) {};
	\node at ($(x6) + 1.2*(ls)$) {};
\end{tikzpicture}\, + 5400\,\,
\begin{tikzpicture}[line width=1.0 pt,baseline=-2.6*\fontsizeshift]
		\coordinate (C) at (0.75,-0.5);
	\coordinate (x1) at (0,0);
	\coordinate (x2) at (0,-1);
	\coordinate (x3) at (1.5,0);
	\coordinate (x4) at (1.5,-1);
	\coordinate (ls) at (0,-.3);
	\coordinate (x5) at (0,-1.5);
	\coordinate (x6) at (1.5,-1.5);
	\draw[scs] (x1) -- (C);
	\draw[scs] (x2) -- (C);
	\draw[scs] (x3) -- (C);
	\draw[scs] (x4) -- (C);
	\draw[scs] (x5) -- (x6);
	\draw[fill] (C) circle (.05);
	\draw[fill] (x1) circle (.05);
	\draw[fill] (x2) circle (.05);
	\draw[fill] (x3) circle (.05);
	\draw[fill] (x4) circle (.05);
	\draw[fill] (x5) circle (.05);
	\draw[fill] (x6) circle (.05);
	\node at ($(C) + 1.2*(ls)$) {$z$};
	\node at ($(x1) + 1.2*(ls)$) {};
	\node at ($(x2) + 1.2*(ls)$) {};
	\node at ($(x3) + 1.2*(ls)$) {};
	\node at ($(x4) + 1.2*(ls)$) {};
	\node at ($(x5) + 1.2*(ls)$) {};
	\node at ($(x6) + 1.2*(ls)$) {};
	\begin{scope}[shift={(0,.34)}]\draw[scs] (0.75,-0.5) circle (.3);\end{scope}
\end{tikzpicture}\,
+ 4050\,\,\, \begin{tikzpicture}[line width=1.0 pt,baseline=-2.3*\fontsizeshift]
		\coordinate (C) at (0.75,0);
	\coordinate (x1) at (0,0);
	\coordinate (x2) at (1.5,0);
	\coordinate (x3) at (0,-1);
	\coordinate (x4) at (1.5,-1);
	\coordinate (x5) at (0,-1.5);
	\coordinate (x6) at (1.5,-1.5);
	\coordinate (ls) at (0,-.3);
	\draw[scs] (x1) -- (C);
	\draw[scs] (C) -- (x2);
	\draw[scs] (x3) -- (x4);
	\draw[scs] (x5) -- (x6);
	\draw[fill] (x1) circle (.05);
	\draw[fill] (x2) circle (.05);
	\draw[fill] (x3) circle (.05);
	\draw[fill] (x4) circle (.05);
	\draw[fill] (x5) circle (.05);
	\draw[fill] (x6) circle (.05);
	\draw[fill] (C) circle (.05);
	\begin{scope}[shift={(0,.26)}]\draw[scs] (0.75,0) circle (.26);\end{scope}
	\begin{scope}[shift={(0,-.26)}]\draw[scs] (0.75,0) circle (.26);\end{scope}
	\node at ($(C) + 1*(ls)$) {$z$};
	\node at ($(x1) + 1.2*(ls)$) {};
	\node at ($(x2) + 1.2*(ls)$) {};
	\node at ($(x3) + 1.2*(ls)$) {};
	\node at ($(x4) + 1.2*(ls)$) {};
	\node at ($(x5) + 1.2*(ls)$) {};
	\node at ($(x6) + 1.2*(ls)$) {};
\end{tikzpicture}\,\, \Bigg] \nonumber \\
= 15\,\, 
\begin{tikzpicture}[line width=1.0 pt,baseline=-1.6*\fontsizeshift]
		\coordinate (C) at (0,0);
	\coordinate (x3) at (0,-0.5);
	\coordinate (x5) at (0,-1);
	\coordinate (x4) at (1,-0.5);
	\coordinate (x6) at (1,-1);
	\coordinate (x2) at (1,0);
	\coordinate (x1) at (0,0);
	\coordinate (ls) at (0,-.3);
	\draw[scd] (x1) -- (x2);
	\draw[scd] (x3) -- (x4);
	\draw[scd] (x5) -- (x6);
		\draw[fill] (x1) circle (.05);
	\draw[fill] (x2) circle (.05);
	\draw[fill] (x3) circle (.05);
	\draw[fill] (x4) circle (.05);
	\draw[fill] (x5) circle (.05);
	\draw[fill] (x6) circle (.05);
			\node at ($(x1) + 1.2*(ls)$) {};
	\node at ($(x2) + 1.2*(ls)$) {};
	\node at ($(x3) + 1.2*(ls)$) {};
	\node at ($(x4) + 1.2*(ls)$) {};
	\node at ($(x5) + 1.2*(ls)$) {};
	\node at ($(x6) + 1.2*(ls)$) {};
\end{tikzpicture}\,
&-& 360\, \lambda \,\, \begin{tikzpicture}[line width=1.0 pt,baseline=-3.5*\fontsizeshift]
		\coordinate (C) at (0.75,-1);
	\coordinate (x1) at (0,-0.5);
	\coordinate (x2) at (1.5,-0.5);
	\coordinate (x5) at (0, -1.5);
	\coordinate (x6) at (1.5, -1.5);
	\coordinate (x3) at (0, -2);
	\coordinate (x4) at (1.5, -2);
	\coordinate (ls) at (0,-.3);
	\draw[scd] (x1) -- (C);
	\draw[scd] (x2) -- (C);
	\draw[scd] (x5) -- (C);
	\draw[scd] (x6) -- (C);
	\draw[scd] (x3) -- (x4);
	\draw[fill] (C) circle (.05);
	\draw[fill] (x1) circle (.05);
	\draw[fill] (x2) circle (.05);
	\draw[fill] (x5) circle (.05);
	\draw[fill] (x6) circle (.05);
	\draw[fill] (x3) circle (.05);
	\draw[fill] (x4) circle (.05);
	\node at ($(C) + 1.2*(ls)$) {$y$};
	\node at ($(x1) + 1.2*(ls)$) {};
	\node at ($(x2) + 1.2*(ls)$) {};
	\node at ($(x5) + 1.2*(ls)$) {};
	\node at ($(x6) + 1.2*(ls)$) {};
	\node at ($(x4) + 1.2*(ls)$) {};
	\node at ($(x3) + 1.2*(ls)$) {};
\end{tikzpicture}\,
-\kappa\,\Bigg[ \, 720\,\,
\begin{tikzpicture}[line width=1.0 pt,baseline=-1.6*\fontsizeshift]
		\coordinate (C) at (0.75,-0.5);
	\coordinate (x1) at (0,0);
	\coordinate (x2) at (1.5,0);
	\coordinate (x3) at (0,-1);
	\coordinate (x4) at (1.5,-1);
	\coordinate (x5) at (1.5,-0.5);
	\coordinate (x6) at (0,-0.5);
	\coordinate (ls) at (0,-.3);
	\draw[scd] (x1) -- (C);
	\draw[scd] (x2) -- (C);
	\draw[scd] (x3) -- (C);
	\draw[scd] (x4) -- (C);
	\draw[scd] (x5) -- (C);
	\draw[scd] (x6) -- (C);
	\draw[fill] (C) circle (.05);
	\draw[fill] (x1) circle (.05);
	\draw[fill] (x2) circle (.05);
	\draw[fill] (x3) circle (.05);
	\draw[fill] (x4) circle (.05);
	\draw[fill] (x5) circle (.05);
	\draw[fill] (x6) circle (.05);
		\node at ($(C) + 1.2*(ls)$) {$z$};
	\node at ($(x1) + 1.2*(ls)$) {};
	\node at ($(x2) + 1.2*(ls)$) {};
	\node at ($(x3) + 1.2*(ls)$) {};
	\node at ($(x4) + 1.2*(ls)$) {};
	\node at ($(x5) + 1.2*(ls)$) {};
	\node at ($(x6) + 1.2*(ls)$) {};
\end{tikzpicture}\, + 5400\,\,
\begin{tikzpicture}[line width=1.0 pt,baseline=-2*\fontsizeshift]
		\coordinate (C) at (0.75,-0.5);
	\coordinate (x1) at (0,0);
	\coordinate (x2) at (0,-1);
	\coordinate (x3) at (1.5,0);
	\coordinate (x4) at (1.5,-1);
	\coordinate (ls) at (0,-.3);
	\coordinate (x5) at (0,-1.5);
	\coordinate (x6) at (1.5,-1.5);
	\draw[scd] (x1) -- (C);
	\draw[scd] (x2) -- (C);
	\draw[scd] (x3) -- (C);
	\draw[scd] (x4) -- (C);
	\draw[scd] (x5) -- (x6);
	\draw[fill] (C) circle (.05);
	\draw[fill] (x1) circle (.05);
	\draw[fill] (x2) circle (.05);
	\draw[fill] (x3) circle (.05);
	\draw[fill] (x4) circle (.05);
	\draw[fill] (x5) circle (.05);
	\draw[fill] (x6) circle (.05);
	\node at ($(C) + 1.2*(ls)$) {$z$};
	\node at ($(x1) + 1.2*(ls)$) {};
	\node at ($(x2) + 1.2*(ls)$) {};
	\node at ($(x3) + 1.2*(ls)$) {};
	\node at ($(x4) + 1.2*(ls)$) {};
	\node at ($(x5) + 1.2*(ls)$) {};
	\node at ($(x6) + 1.2*(ls)$) {};
	\begin{scope}[shift={(0,.34)}]\draw[scd] (0.75,-0.5) circle (.3);\end{scope}
\end{tikzpicture}\,\Bigg] \nonumber 
\end{eqnarray}
\begin{eqnarray}
&=&  \bigg[ K_{12} K_{34} K_{56} +  K_{12} K_{35} K_{46} + K_{12} K_{36} K_{45} + K_{13} K_{24} K_{56} + K_{13} K_{25} K_{46} + K_{13} K_{26} K_{45} + K_{14} K_{23} K_{56} \nonumber \\ 
&+& K_{14} K_{25} K_{36} + K_{14} K_{26} K_{35} + K_{15} K_{23} K_{46} + K_{15} K_{24} K_{36} + K_{15} K_{26} K_{34}  + K_{16} K_{23} K_{45} + K_{16} K_{24} K_{35} \nonumber \\ &+& K_{16} K_{25} K_{34} \bigg] - 24 \, \int d^{d_\text{in}}y\,\lambda\, \bigg[  K_{W, 1y}K_{W, 2y}K_{W, 3y}K_{W, 4y} K_{56} + K_{W, 1y}K_{W, 2y}K_{W, 3y}K_{W, 5y} K_{46} \nonumber \\ &+& K_{W, 1y}K_{W, 2y}K_{W, 4y}K_{W, 5y} K_{36} 
 + K_{W, 1y}K_{W, 3y}K_{W, 4y}K_{W, 5y} K_{26} + K_{W, 2y}K_{W, 3y}K_{W, 4y}K_{W, 5y} K_{16} \nonumber \\ &+& K_{W, 1y}K_{W, 2y}K_{W, 3y}K_{W,6y} K_{45} + K_{W,1y}K_{W,2y}K_{W,4y}K_{W,6y} K_{35} + K_{W,1y}K_{W, 3y}K_{W, 4y}K_{W, 6y} K_{25} 
\nonumber \\ &+& K_{W, 2y}K_{W, 3y}K_{W, 4y}K_{W, 6y} K_{15} + K_{W, 1y}K_{W, 2y}K_{W, 5y}K_{W, 6y} K_{34} + K_{W, 1y}K_{W, 3y}K_{W, 5y}K_{W, 6y} K_{24} \nonumber \\ &+& K_{W, 2y}K_{W, 3y}K_{W, 5y}K_{W, 6y} K_{14} + K_{W, 1y}K_{W, 4y}K_{W, 5y}K_{W, 6y} K_{23} 
+ K_{W, 2y}K_{W, 4y}K_{W, 5y}K_{W, 6y} K_{13}  \nonumber \\ &+& K_{W, 3y}K_{W, 4y}K_{W, 5y}K_{W, 6y} K_{12}\bigg] -  720 \, \int d^{\din} z \, \kappa \, K_{W, 1z} K_{W, 2z} K_{W, 3z} K_{W, 4z} K_{W, 5z} K_{W, 6z} \nonumber \\
&-& 360 \, \int d^{\din} z \, \kappa \, \bigg[ K_{W, zz} K_{W, 1z}K_{W, 2z}K_W, {3z}K_{W, 4z} K_{56} + K_{W, zz} K_{W, 1z}K_{W, 2z}K_{W,3z}K_{W, 5z} K_{46} \nonumber \\ &+& K_{W, zz}K_{W, 1z}K_{W, 2z}K_{W, 4z}K_{W, 5z} K_{36} 
+ K_{W, zz}K_{W, 1z}K_{W, 3z}K_{W, 4z}K_{W, 5z} K_{26} + K_{W, zz}K_{W, 2z}K_{W, 3z}K_{W, 4z}K_{W, 5z} K_{16} \nonumber \\ &+& K_{W, zz}K_{W, 1z}K_{W, 2z}K_{W, 3z}K_{W, 6z} K_{45} + K_{W, zz}K_{W, 1z}K_{W, 2z}K_{W, 4z}K_{W, 6z} K_{35}  + K_{W, zz}K_{W, 1z}K_{W, 3z}K_{W, 4z}K_{W, 6z} K_{25} \nonumber \\
&+& K_{W, zz}K_{W, 2z}K_{W, 3z}K_{W, 4z}K_{W, 6z} K_{15} + K_{W, zz}K_{W, 1z}K_{W, 2z}K_{W, 5z}K_{W, 6z} K_{34} + K_{W, zz}K_{W, 1z}K_{W, 3z}K_{W, 5z}K_{W, 6z} K_{24} \nonumber \\ &+& K_{W, zz}K_{W, 2z}K_{W, 3z}K_{W, 5z}K_{W, 6z} K_{14} + K_{W, zz}K_{W, 1z}K_{W, 4z}K_{W, 5z}K_{W, 6z} K_{23} 
+  K_{W, zz}K_{W, 2z}K_{W, 4z}K_{W, 5z}K_{W, 6z} K_{13} \nonumber \\ &+& K_{W, zz}K_{W, 3z}K_{W, 4z}K_{W, 5z}K_{W, 6z} K_{12}\bigg],
\end{eqnarray}
We have expanded to leading order in $\lambda$ and $\kappa$. Since the connected $4$-pt and $6$-pt functions
scale as $1/N$ and $1/N^2$, respectively, (see Section \ref{sec:nscaling}), we infer the same scaling for $\lambda$
and $\kappa$, allowing us to ignore $\kappa$ at leading order in $1/N$ in our experiments.
The statement of technical naturalness gets trivially modified to an associated statement for $K_W$.

}
\subsection{Experiments: Correlations in Single-Layer Networks with EFT \label{sec:EFTforSingle}}

Having explained how to utilize effective field theory to analyze neural network
architectures and their associated NGPs, in this Section we verify the ideas experimentally. 
The logic is simple, as effective field theory techniques are supposed to accomplish three things:
\begin{itemize}
\item Give a candidate \amnew{$\Delta S_W$} for the NGP.
\item Fix coefficients of operators in \amnew{$\Delta S_W$} with experiments.
\item Once fixed, make predictions for other experiments and verify them. 
\end{itemize}
In this Section, we will use experimental measurements of $G^{(4)}$ at fixed width to determine 
$\lambda$, and then use the determined $\lambda$ to make predictions for $G^{(6)}$ at the same width and verify it against experiments. \amnew{We choose widths $5$, $20$, and $1000$ for Erf-net, ReLU-net, and Gauss-net experiments respectively, to ensure that the networks of our interest receive non-trivial corrections to the GP, as demonstrated experimentally in Section \ref{sec:FreeGP}.}

The coupling $\lambda$ may be extracted from experimental measurements of $G^{(4)}$. When
$\kappa$ is negligible\footnote{We will verify this in the large cutoff limit experimentally,
and theoretically
in Section \eqref{sec:rge}.}
and $\lambda$ is a constant, \eqref{full4-ptdiagrams} gives

\bea 
\label{eqn:lambdabar1}
\lambda = \frac{  K(x_1,x_2)K(x_3,x_4) + K(x_1,x_3)K(x_2,x_4) + K(x_1,x_4)K(x_2,x_3) -G^{(4)}(x_1,x_2, x_3, x_4)  }{ 24 \int\,d^{d_\text{in}}y \, K_\jimnew{W}(x_1,y)K_\jimnew{W}(x_2,y)K_\jimnew{W}(x_3,y)K_\jimnew{W}(x_4,y)}.
\eea
Therefore, by measuring $G^{(4)}(x_1,x_2,x_3,x_4)$ in experiments and performing the 
theoretical computations for the rest of the expression, $\lambda$ may be experimentally
measured. 

Slight complications arise when, more generally, the coupling is a function 
\be \lambda(y) = \bar \lambda + \delta \lambda(y) \ee with $\bar \lambda$ a constant.
In that case \eqref{full4-ptdiagrams}  gives
\bea
\label{eqn:lambdabar}
\bar \lambda = \frac{  K_{12}K_{34} + K_{13}K_{24} + K_{14}K_{23} -G^{(4)}(x_1,x_2, x_3, x_4)  }{ 24 \int\,d^{d_\text{in}}y \, \Delta_{1234y}} - \frac{\int\,d^{d_\text{in}}y \,\delta \lambda(y)\, \Delta_{1234y}}{\int\,d^{d_\text{in}}y \, \Delta_{1234y}} \\ \nonumber
\eea
where $K_{\jimnew{W},ij}$ abbreviates $K_\jimnew{W}(x_i,x_j)$ and $\Delta_{1234y}= K_{\jimnew{W},1y}K_{\jimnew{W},2y}K_{\jimnew{W},3y}K_{\jimnew{W},4y}$. If 
$\delta \lambda$ is small then the first term
\be
\label{eqn:lambdam}
\lambda_m(x_1, x_2, x_3, x_4):=\frac{  K_{12}K_{34} + K_{13}K_{24} + K_{14}K_{23} -G^{(4)}(x_1,x_2, x_3, x_4)  }{ 24 \int\,d^{d_\text{in}}y \, \Delta_{1234y}}
\ee
dominates, which may
be measured experimentally by measuring $G^{(4)}$ and then using the known theoretical expressions
of the GP kernels to compute $\lambda_m$, which becomes a rank four tensor 
on a fixed set of inputs. Since $\bar \lambda$ is a constant and \eqref{eqn:lambdabar} holds, variance in $\lambda_m$ correlates with variance in $\delta \lambda$, and when the variance
is small
\begin{equation}
\lambda \simeq \bar \lambda \simeq \text{mean}(\lambda_m(x_1, x_2, x_3, x_4)).
\end{equation}
The approximation is exact in the limit of constant $\lambda$, i.e. $\delta \lambda \to 0$. Therefore, by measuring $\lambda_m$ we approximately measure $\lambda$ as $\text{mean}(\lambda_m(x_1, x_2, x_3, x_4))$\footnote{The rank four tensor $\lambda_m(x_1, x_2, x_3, x_4)$ is symmetric and the mean is taken over its unique (upper triangular) elements.}, which we refer to as the measured value of $\lambda$. 

Given the measured
value of $\lambda$, we would like to use it to make theoretical predictions
for $G^{(6)}$ that may be experimentally verified. To this end, we define the $6$-pt deviation
\begin{eqnarray}
\label{6-ptdeviation}
\mathcal{\delta}^\prime(x_1,\dots,x_6)
&:=& G^{(6)}(x_1,\dots,x_6) - \sum_{\text{15 combinations}} \Bigg[ K(x_i, x_j)K(x_k,x_l)K(x_m,x_n) \nonumber \\  &-& 24\int \, d^{d_\text{in}}y\,\, \lambda \, \,K_\jimnew{W}(x_i,y)K_\jimnew{W}(x_j,y)K_\jimnew{W}(x_k,y)K_\jimnew{W}(x_l,y)K(x_m,x_n)  \Bigg] \nonumber \\
&=& -\kappa\,\Bigg[ \, 720\,\,
\begin{tikzpicture}[line width=1.0 pt,baseline=-1.6*\fontsizeshift]
		\coordinate (C) at (0.75,-0.5);
	\coordinate (x1) at (0,0);
	\coordinate (x2) at (1.5,0);
	\coordinate (x3) at (0,-1);
	\coordinate (x4) at (1.5,-1);
	\coordinate (x5) at (1.5,-0.5);
	\coordinate (x6) at (0,-0.5);
	\coordinate (ls) at (0,-.3);
	\draw[scd] (x1) -- (C);
	\draw[scd] (x2) -- (C);
	\draw[scd] (x3) -- (C);
	\draw[scd] (x4) -- (C);
	\draw[scd] (x5) -- (C);
	\draw[scd] (x6) -- (C);
	\draw[fill] (C) circle (.05);
	\draw[fill] (x1) circle (.05);
	\draw[fill] (x2) circle (.05);
	\draw[fill] (x3) circle (.05);
	\draw[fill] (x4) circle (.05);
	\draw[fill] (x5) circle (.05);
	\draw[fill] (x6) circle (.05);
		\node at ($(C) + 1.2*(ls)$) {$z$};
	\node at ($(x1) + 1.2*(ls)$) {};
	\node at ($(x2) + 1.2*(ls)$) {};
	\node at ($(x3) + 1.2*(ls)$) {};
	\node at ($(x4) + 1.2*(ls)$) {};
	\node at ($(x5) + 1.2*(ls)$) {};
	\node at ($(x6) + 1.2*(ls)$) {};
\end{tikzpicture}\, + 5400\,\,
\begin{tikzpicture}[line width=1.0 pt,baseline=-2.4*\fontsizeshift]
		\coordinate (C) at (0.75,-0.5);
	\coordinate (x1) at (0,0);
	\coordinate (x2) at (0,-1);
	\coordinate (x3) at (1.5,0);
	\coordinate (x4) at (1.5,-1);
	\coordinate (ls) at (0,-.3);
	\coordinate (x5) at (0,-1.5);
	\coordinate (x6) at (1.5,-1.5);
	\draw[scd] (x1) -- (C);
	\draw[scd] (x2) -- (C);
	\draw[scd] (x3) -- (C);
	\draw[scd] (x4) -- (C);
	\draw[scd] (x5) -- (x6);
	\draw[fill] (C) circle (.05);
	\draw[fill] (x1) circle (.05);
	\draw[fill] (x2) circle (.05);
	\draw[fill] (x3) circle (.05);
	\draw[fill] (x4) circle (.05);
	\draw[fill] (x5) circle (.05);
	\draw[fill] (x6) circle (.05);
	\node at ($(C) + 1.2*(ls)$) {$z$};
	\node at ($(x1) + 1.2*(ls)$) {};
	\node at ($(x2) + 1.2*(ls)$) {};
	\node at ($(x3) + 1.2*(ls)$) {};
	\node at ($(x4) + 1.2*(ls)$) {};
	\node at ($(x5) + 1.2*(ls)$) {};
	\node at ($(x6) + 1.2*(ls)$) {};
	\begin{scope}[shift={(0,.34)}]\draw[scd] (0.75,-0.5) circle (.3);\end{scope}
\end{tikzpicture} \,\, \Bigg]  
\end{eqnarray} 
where the last equality follows from \eqref{6-ptlambda}. To obtain a sense of the size of the deviation
we normalize it with respect to the experimental $6$-pt function of the NGP,
\be \delta(x_1,\dots,x_6) := \frac{\delta'(x_1,\dots,x_6)}{G^{(6)}(x_1,\dots,x_6)}\ee 
For perturbation theory to hold, $|\delta(x_1,\dots,x_6)| \ll 1$ is required. Since $\kappa$ is negligible in the limit of large cutoff $\Lambda$ \jimnew{and it is suppressed by $1/N$ relative to $\lambda$}, we expect $|\delta(x_1,\dots,x_6)|$ to converge to $0$ \jimnew{in the large-$N$} and large-$\Lambda$ limits. For each NN architecture, $|\delta(x_1,\dots,x_6)|$ is given by the difference between the orange line and $1.0$ in Figures \ref{fig2.1}, \ref{fig:reludelta} and \ref{fig:erfdelta}. 

\subsubsection*{\gnet}

\begin{figure}[t]
\centering
\includegraphics[width=0.49\textwidth]{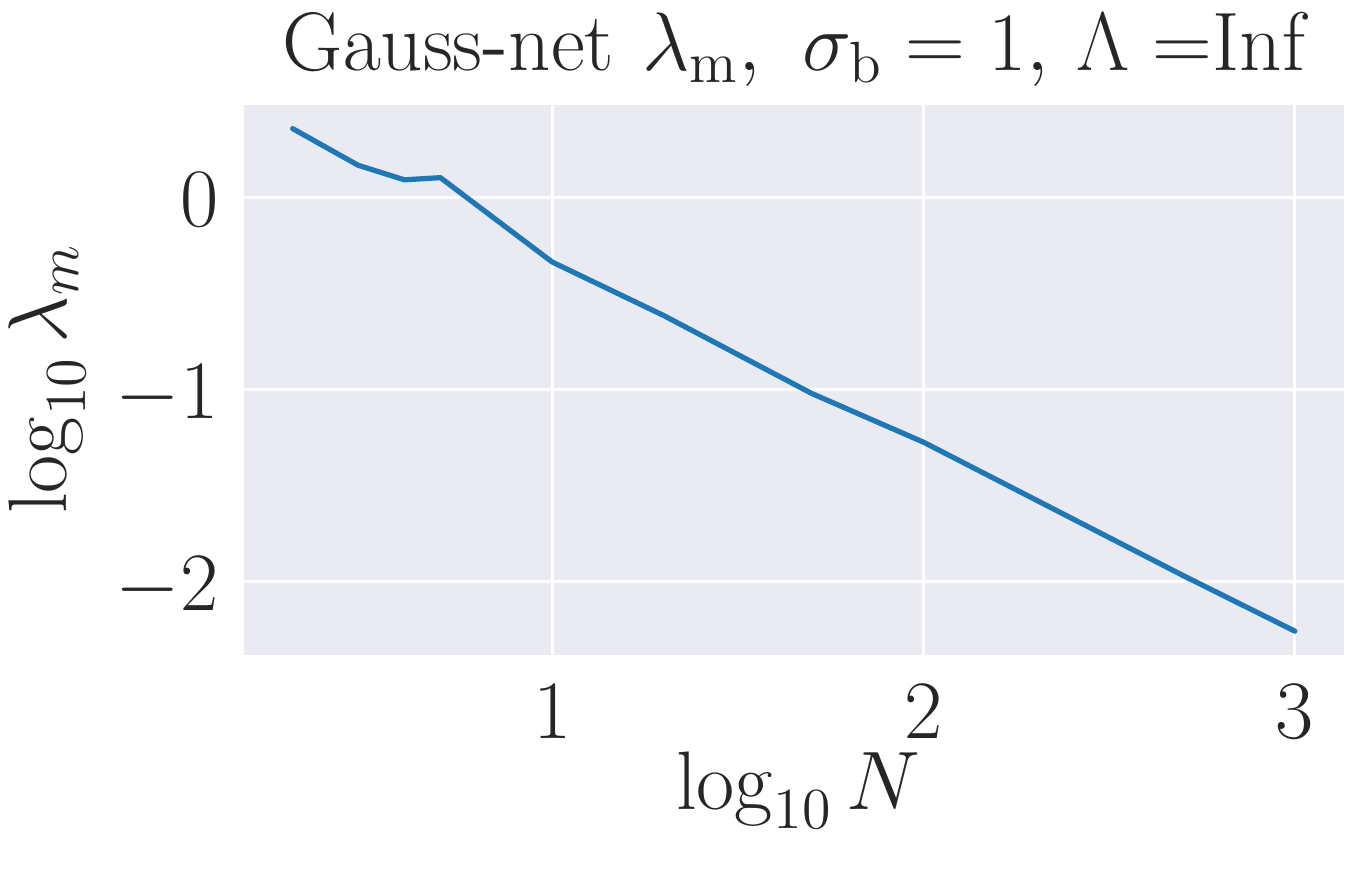}
\includegraphics[width=0.495\textwidth]{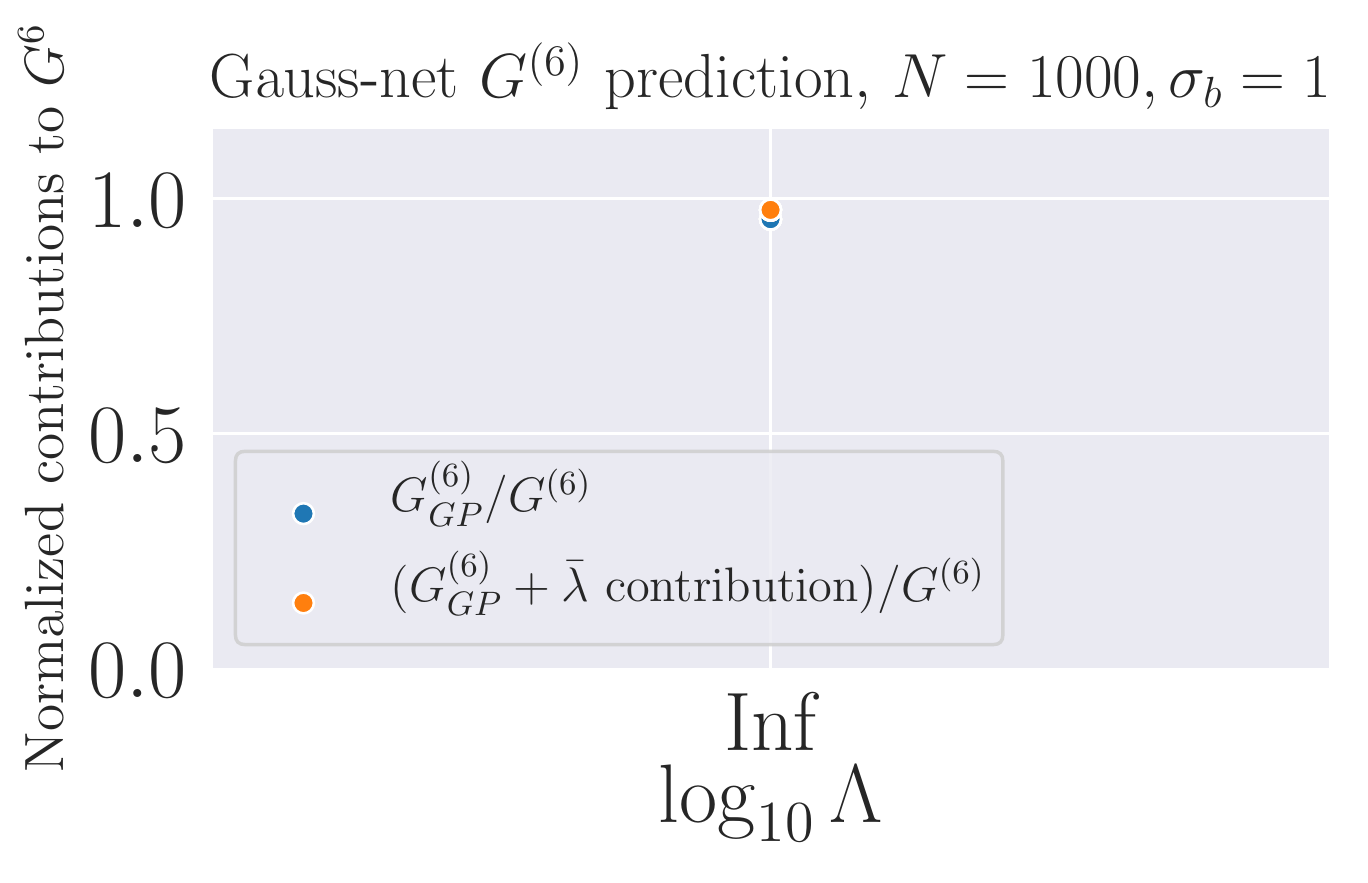}
\caption{(left): Measured $\lambda_{m}$ tensor elements at different widths for \gnet~with infinite cutoff. (right): GP prediction alone, and GP prediction + $\bar \lambda$ correction of $6$-pt function $G^{(6)}$ for \gnet~at width $N = 1000$, normalized with respect to $G^{(6)}$.}
\label{fig2.1}
\end{figure}

In Figure \ref{fig2.1} we show the results of \gnet $\,$ experiments,
which were carried out with the same set of parameters introduced in Section
\ref{sec:networksetup}. All experiments in this Section were performed with $n_{\text{nets}} = 10^{7}$. The fixed-cutoff experiments are done at 
\be
\Lambda \rightarrow \infty, \qquad N \in \{2, 3, 4, 5, 10, 20, 50, 100, 500, 1000\}.
\ee
\amnew{The $G^{(6)}$ prediction experiments using $\lambda_{m}$ are done at width $N = 1000$ and $\Lambda \rightarrow \infty$, since the \gnet~kernel goes to zero at large $x$ and therefore its EFT integrals converge.}

In the left plot we present measurements for the rank-$4$ tensor $\lambda_m$ at fixed cutoff. Despite being a rank-$4$ tensor, we see that there is very little variance across the elements,
indicating that $\lambda_m$ is effectively constant; we will discuss this
in a moment. In the right plot we present the GP contribution to $G^{(6)}$, normalized by $G^{(6)}$ so that a correct theoretical prediction
would be $1$; we see $\ggp^{(6)}$ falls short of this. However,
we also demonstrate excellent match to experiment when the perturbative correction in $\lambda$ is added, implying that $\delta$ is effectively
zero; the EFT has correctly predicted 
the experimental measurements of the $6$-pt function. 

As discussed in Section \ref{sec-technical-naturalness}, by technical naturalness a parameter is allowed to be small when setting it to zero restores a symmetry. We observe from the (lack of) variance in the left plot in Figure \ref{fig2.1} that $\delta \lambda / \bar \lambda$ is indeed small, i.e. $\lambda$ is effectively a constant. This is expected because $\lambda$ is technically natural for \gnet: translation invariance is recovered when $\delta \lambda$ vanishes, as the \gnet $\,$ GP
kernel is translation invariant. In fact, this was by design. We started with a translation-invariant kernel since we desire an example with coupling constants rather than coupling functions, which we expected would be enforced by technical naturalness.

\subsubsection*{ReLU-net}
\begin{figure}[t]
\centering
\includegraphics[width=0.49\textwidth]{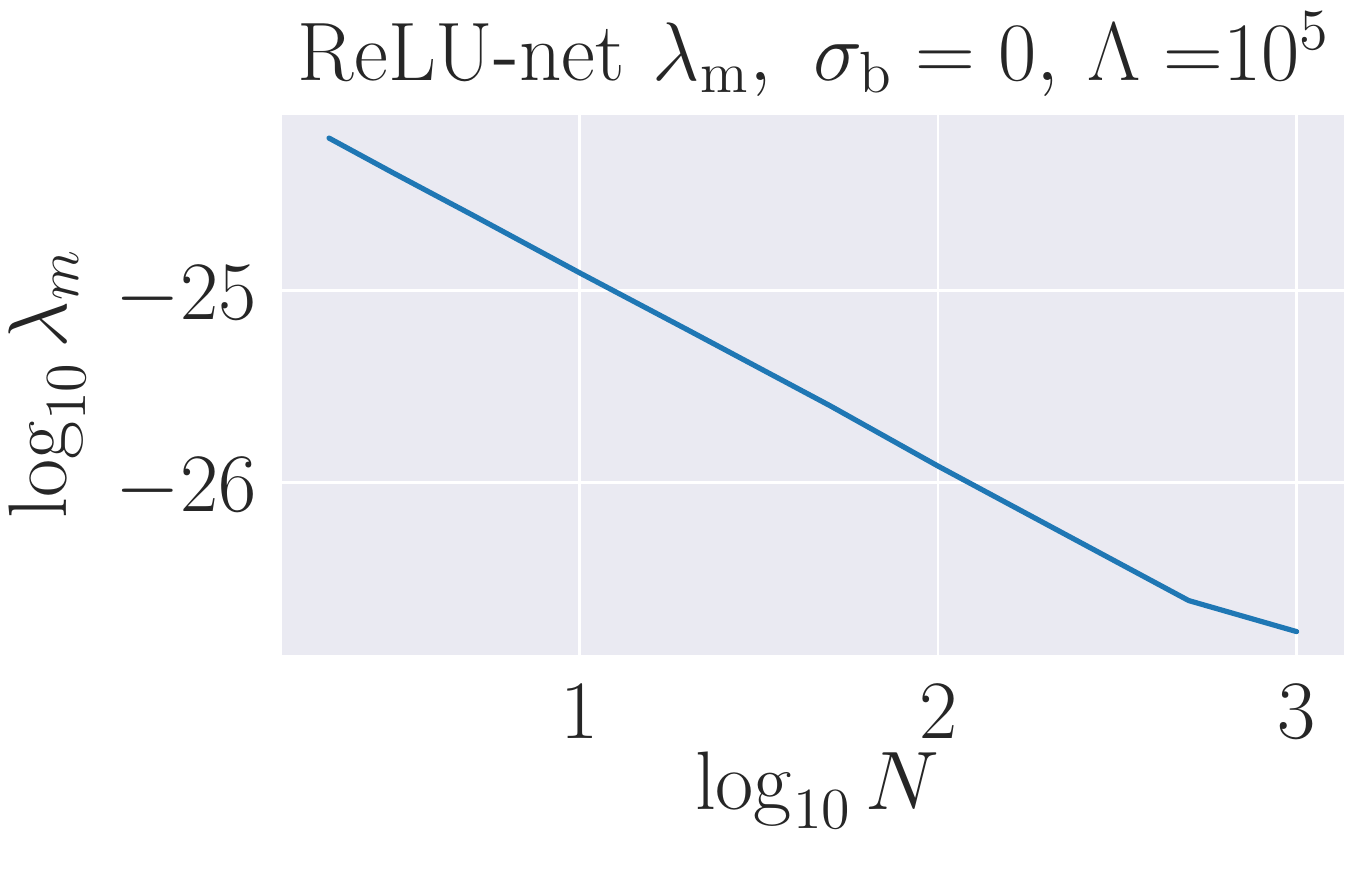}
\includegraphics[width=0.49\textwidth]{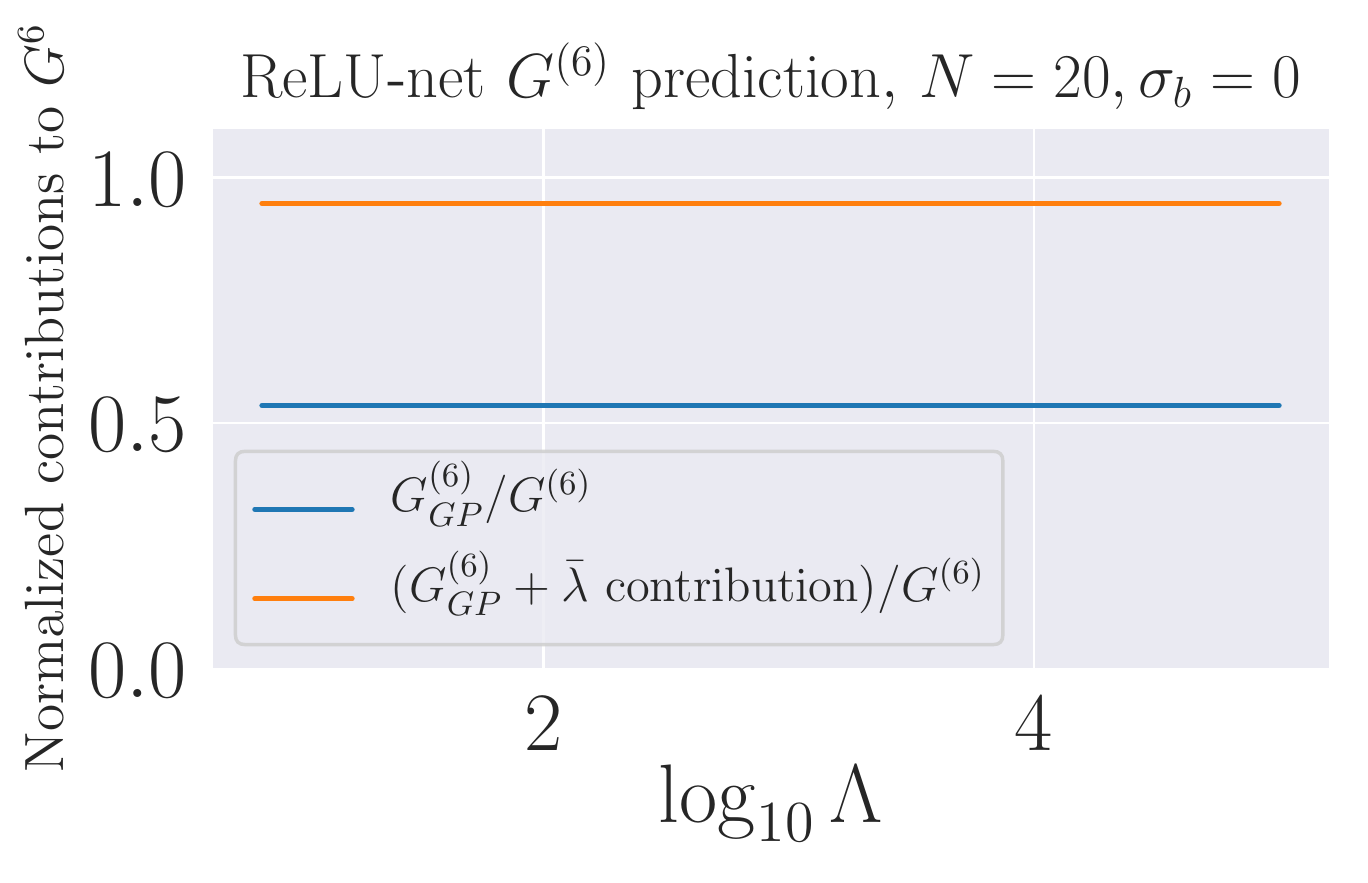}
\caption{(left): Measured $\lambda_{m}$ tensor elements at different widths for ReLU-net with fixed cutoff. (right): GP prediction alone, and GP prediction + $\bar \lambda$ correction of 6pt function $G^{(6)}$ for ReLU-net at width \amnew{$N = 20$}, normalized with respect to $G^{(6)}$.}  
\label{fig:reludelta}
\end{figure}

Further, we read off $\lambda_m$ tensor and its components from ReLU-net in Figure \ref{fig:reludelta}. The fixed-cutoff experiments were done with $\Lambda = 10^{5}$, and the fixed-width $\lambda_{m}$ experiments and $G^{(6)}$ predictions are done at width and cutoffs
\begin{align} \label{cutoffs}
N=20, \qquad  \Lambda \in \{7, 10, 15, 20, 30, 40, 50, 70, 100, 200, 500, 1000, 2000, 5000, 7000, 10000, \\\nonumber 20000, 40000, 60000, 80000, 100000\}.
\end{align}
Similar results can be shown at other widths.

The widths were chosen so the 4-pt and 6-pt functions were NGPs and above the background level, as seen in Figure \ref{fig1}. 
As with \gnet, we show in the left plot all elements of the rank-4 tensor $\lambda_{m}$ at various widths and fixed cutoff $\Lambda = 10^{5}$.
Using the measured $\lambda$, we show the results of the corresponding $G^{(6)}$ prediction in the right plot. We see similarly that the GP prediction $G_{\text{GP}}^{(6)}$ does not account for the 6-pt function $G^{(6)}$, and the $\lambda$ contribution gives a significant additive correction.

\subsubsection*{Erf-net}
\begin{figure}[t]
		\centering
	\includegraphics[width=0.49\textwidth]{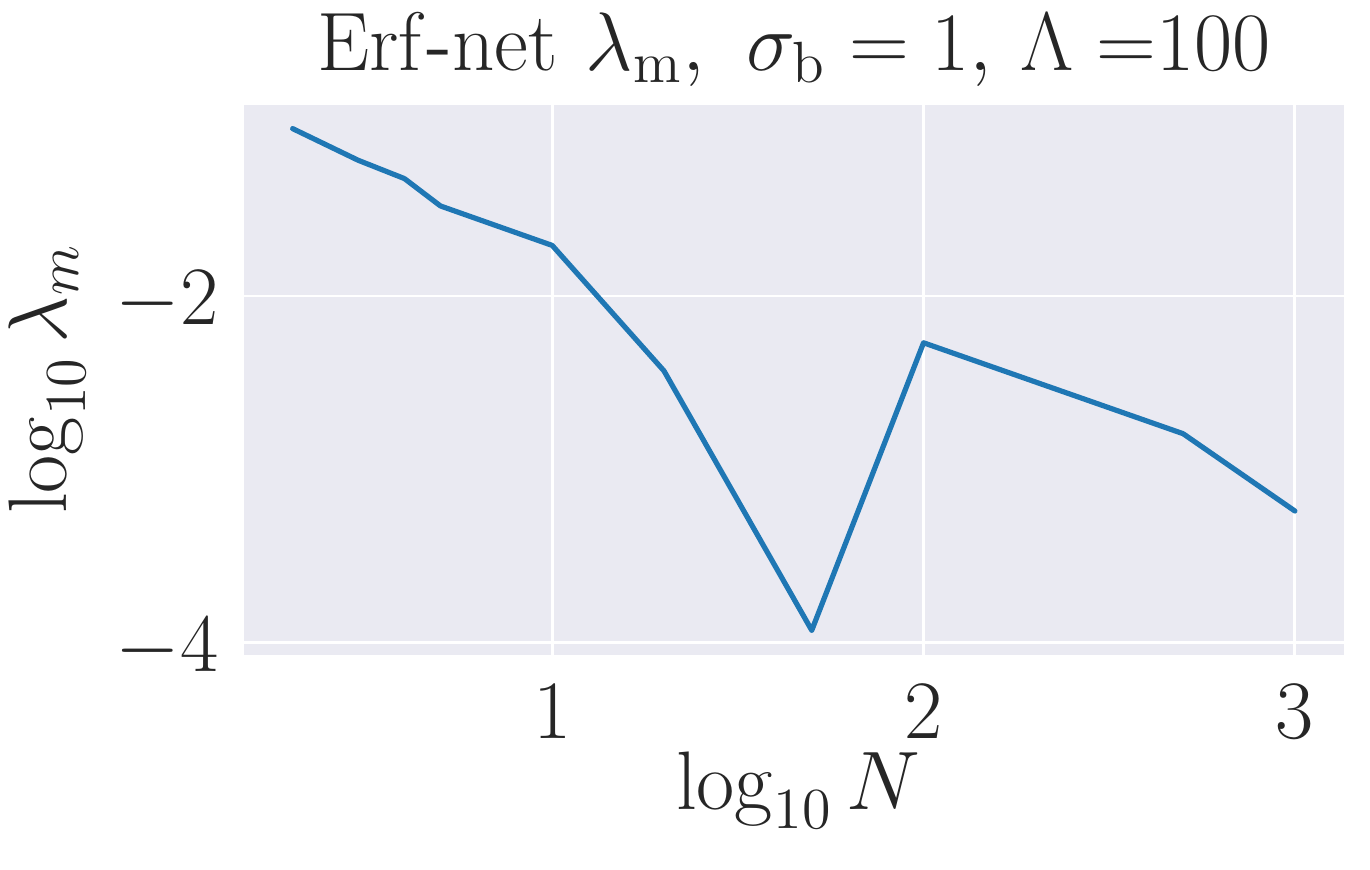}
	\includegraphics[width=0.49\textwidth]{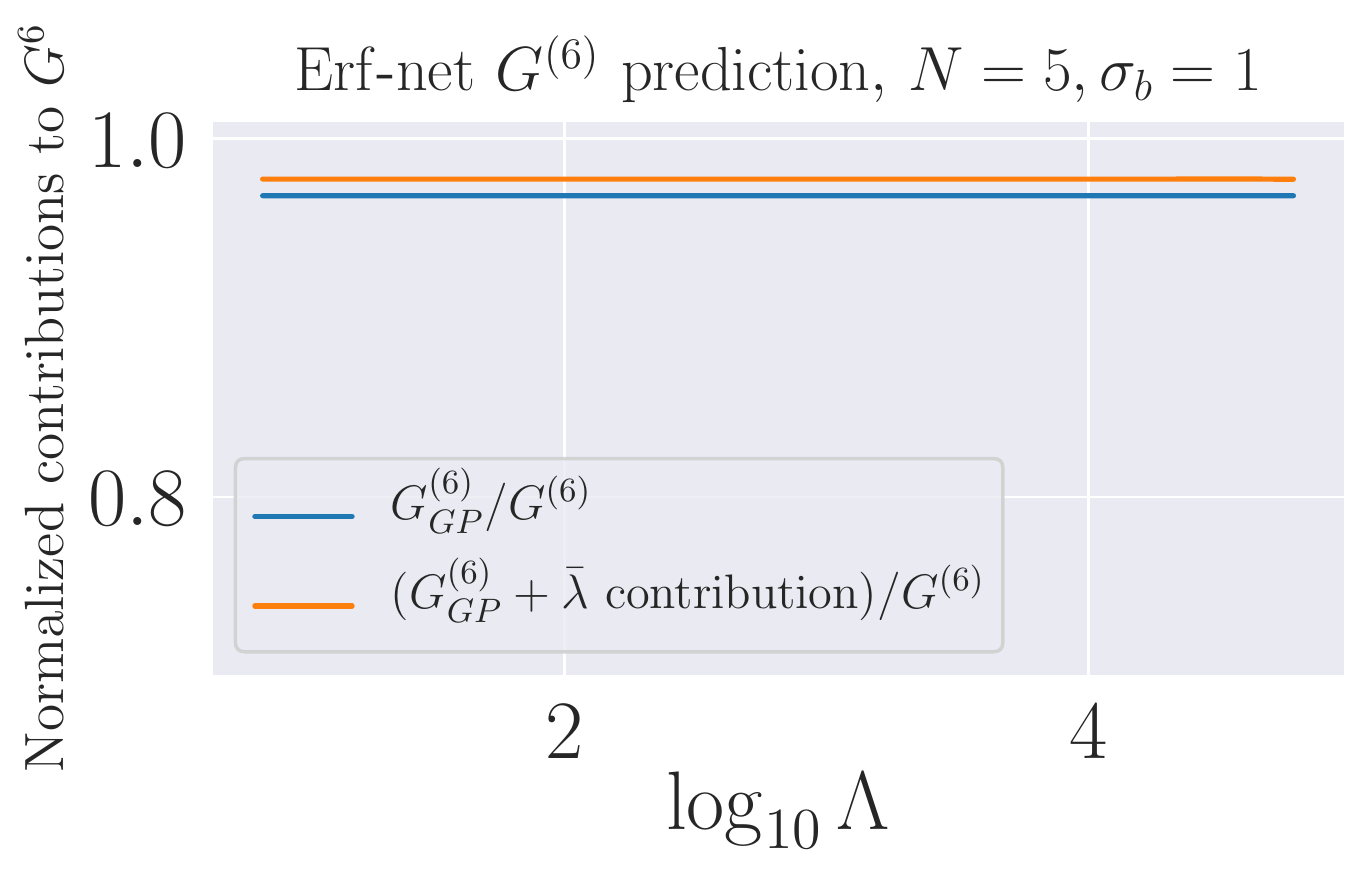}
	\caption{(left): Measured $\lambda_{m}$ tensor elements at different widths for Erf-net with fixed cutoff. (right): GP prediction alone, and GP prediction + $\bar \lambda$ correction of 6pt function $G^{(6)}$ for Erf-net at width $N = 5$, normalized with respect to $G^{(6)}$.}
	\label{fig:erfdelta}
	\end{figure}
We obtain $\lambda_m$ tensor and its independent components from experimental results for the third and last architecture Erf-net in Figure \ref{fig:erfdelta}. As with ReLU-net, the fixed-cutoff experiments were done at $\Lambda = 10^{5}$, and the fixed-width experiments were performed at width $N = 5$. 

As with the previous two networks, we show all elements of the tensor $\lambda_{m}$ in the left plot at the given widths, the quartic coupling appears effectively constant due to smallness of variations across its tensor elements. The right plot shows the GP prediction $G_{GP}^{(6)}$ being corrected by the $\bar \lambda$ term, and together they give a good approximation of the 6-pt function $G^{(6)}$.

\amnew{
\subsection{Experiments: Precision Quartic Fitting and EFT Tests of $G^{(4)}$} \label{sec:eft-couplings}

We now provide another test of EFT techniques via a test-train split for $\Delta G^{(4)}$ predictions. Specifically, since $\Delta G^{(4)}$ is a continuous function of inputs, we may fit EFT parameters for a finite set of inputs and then test whether they make accurate predictions for unseen inputs. We may do this using only the quartic couplings because they are leading in $1/N$, as demonstrated in Section \ref{sec:nscaling}. In the process, it introduces another method for measuring couplings.

In Section \ref{sec:EFTforSingle} we considered the $6$-pt functions receiving leading corrections due to the term
\begin{eqnarray}
\Delta S_W = \int d^{\din} x\, \lambda \, f_W(x)^4,
\end{eqnarray}
with $\lambda = \lambda_0$, a constant. In general however, one might consider $\lambda$ being a local and/or non-local function of $x$, and test out the best functional form of $\lambda$ for a given architecture. We design three such models
\begin{eqnarray}
M_1 : \Delta S^1_W &=& \int d^{\din} x\, \lambda_0\, f_W(x)^4 , \\
M_2 : \Delta S^2_W &=& \int d^{\din} x \,( \lambda_0 + \lambda_2 x^2 ) \, f_W(x)^4 , \\
M_3 : \Delta S^3_W &=& \int d^{\din} x \,( \lambda_0 + \lambda_2 x^2 )\,  f_W(x)^4 + \int d^{\din}xd^{\din}y \, \lambda_{\text{NL}}\, f_W(x)^2f_W(y)^2 .
\end{eqnarray}

Since this fit was done using the 4-pt function, an input consists of a 4-tuple of external points $(x_{1}, x_{2}, x_{3}, x_{4})$. The $10^{7}$ experiments were run twice, using two disjoint sets of external points called the test set and train set, described in Table \ref{tab:testtrain}. 
First, we train the model.
For each of the three model types, we compute $\Delta G^{(4)}$ averaged over $10^{7}$ experiments, and also piecewise contributions from the EFT (proportional to the couplings), for all combinations of the external points in the train set. \jimnew{This gives a rank four tensor $\Delta G^{(4)}_\text{exp}$ and EFT corrections $\Delta G^{(4)}_\text{EFT}$ . At leading order in the $\lambda$-couplings the EFT corrections are a sum of three terms, each proportional to a coupling, giving 
\begin{equation}
\Delta G^{(4)}_\text{EFT} = \lambda_0 T_0(x_1,\dots,x_4) + \lambda_2 T_2(x_1,\dots,x_4) + \lambda_{\text{NL}} T_{\text{NL}}(x_1,\dots,x_4).
\end{equation}
Via perturbation theory, we have
\begin{align}
T_0(x_1,\dots,x_4) &= 24\, \int \, d^{\din} x\, K_W(x_1, x)K_W(x_2, x)K_W(x_3, x)K_W(x_4, x) \nonumber \\
T_2(x_1,\dots,x_4) &= 24\, \int \, d^{\din} x\,\,\, x^2 \,\,\, K_W(x_1, x)K_W(x_2, x)K_W(x_3, x)K_W(x_4, x) \nonumber \\
T_\text{NL}(x_1,\dots,x_4) &= 8\, \int \, d^{\din} x  d^{\din} y\, \Bigg[ K_W(x_1, x)K_W(x_2, x)K_W(x_3, y)K_W(x_4, y) +  \nonumber \\
&K_W(x_1, x)K_W(x_2, y)K_W(x_3, x)K_W(x_4, y) + \nonumber \\ & K_W(x_1, x)K_W(x_2, y)K_W(x_3, y)K_W(x_4, x) \Bigg].
\end{align}
These tensors can then be compared elementwise and the values of $\lambda$ can be found such that they minimize a cost function over all tensor elements. This will yield an EFT whose functional form best approximates the true functional form of the $\Delta G^{(4)}_\text{exp}$.} We find the optimal parameter fits by using PyTorch to minimize the mean squared error between experimental measures of $\Delta G^{(4)}$ and corresponding EFT predictions $\Delta G^{(4)}_\text{EFT}$ for each model $M_1$, $M_2$, and $M_3$ using stochastic gradient descent. We test the effectiveness of the measured $\lambda_0$, $\lambda_2$ and $\lambda_{\text{NL}}$ by making predictions for the test set, computing the MSE loss and MAPE between $\Delta G^{(4)}$ and the EFT model contribution. MAPE is the mean absolute percent error, defined as
\begin{eqnarray}
\text{MAPE} = \frac{1}{p}\sum_{i = 1}^p \frac{A_i - F_i}{A_i}.
\end{eqnarray} 
$A_i$, $F_i$ and $p$ are respectively the experimental $\Delta G^{(4)}(x_1, x_2, x_3, x_4)$, corresponding EFT prediction and total number of tensor elements in $4$-pt function. 

\begin{table}[htb]
\centering
\begin{tabular}{|l|}
\hline
$\{x_{i}^{\text{\gnet}}\}    = \{-0.01, -0.006, -0.002, +0.002, +0.006, +0.01\}$    \\ \hline
$\{x_{i}^{\text{ReLU-net}} \} = \{ +0.2, +0.4, +0.6, +0.8, +1.0, +1.2\}$             \\ \hline
$\{x_{i}^{\text{Erf-net}}\} = \{ +0.002, +0.004, +0.006, +0.008, +0.010, +0.012\}$ \\ \hline
\end{tabular} \\ \vspace{0.75cm}
	\begin{tabular}{|l|l|l|}
	\hline
			 & train set inputs ${x_{i}}$                           & test set inputs ${x_{i}}$ \\ \hline
	\gnet    & $ \frac{\sqrt{2}}{2} x_{i}^{\text{\gnet}}$ & $x_{i}^{\text{\gnet}}$                          \\ \hline
	ReLU-net & $\frac{\sqrt{2}}{2} x_{i}^{\text{ReLU-net}}$         & $x_{i}^{\text{ReLU-net}}$                          \\ \hline
	Erf-net & $\frac{\sqrt{2}}{2} x_{i}^{\text{Erf-net}}$         & $x_{i}^{\text{Erf-net}}$                          \\ \hline
	\end{tabular}
	\caption{Inputs for test-sets and train-sets}
	\label{tab:testtrain}
\end{table}

\begin{table}[htb]
  \centering
  \begin{tabular}{|ccc|}
  \hline
  & $(\lambda_0,\lambda_2,\lambda_\text{NL})$ & Test (MAPE, MSE) \\ \hline
  \hline
  Gauss $M_0$ & $(0.0,0.0,0.0)$ & $(100,0.019)$ \\ \hline
  Gauss $M_1$ & $(0.0046,0.0,0.0)$ & $(0.0145 ,6.8\times 10^{-10})$  \\ \hline
  Gauss $M_2$ & $(0.0043,0.0011,0.0)$ & $(0.0144,6.7\times 10^{-10})$  \\ \hline
  Gauss $M_3$ & $(0.00062,0.00016,0.0015)$ & $(0.0156,7.5\times 10^{-10})$  \\ \hline \hline
  ReLU $M_0$ & $(0.0,0.0,0.0)$ & $(100, 0.003)$  \\ \hline
  ReLU $M_1$ & $(6.2\times 10^{-11},0.0,0.0)$ & $(0.0035, 7.6 \times 10^{-12})$  \\ \hline
  ReLU $M_2$ & $(1.2 \times 10^{-18}, 8.7 \times 10^{-15},0.0)$ & $(0.0013, 1.5 \times 10^{-12})$  \\ \hline
  ReLU $M_3$ & $(1.2 \times 10^{-18}, 8.7 \times 10^{-15}, 6.8 \times 10^{-17})$ & $( 0.0012 , 1.2 \times 10^{-12})$  \\ \hline  \hline
  Erf $M_0$ & $(0.0,0.0,0.0)$ & $(100, 0.006)$  \\ \hline
  Erf $M_1$ & $(0.039,0.0,0.0)$ & $(0.030, 8.3 \times 10^{-10})$  \\ \hline
  Erf $M_2$ & $(0.040,-0.00043,0.0)$ & $(0.0042, 1.9 \times 10^{-11})$  \\ \hline
  Erf $M_3$ & $(0.0019, -0.0054, 0.0063)$ & $( 0.037 , 1.1 \times 10^{-9})$  \\ \hline \hline
  \end{tabular}
  \caption{Optimized values of $\lambda_0$, $\lambda_2$, $\lambda_{\text{NL}}$ and resulting MAPE and MSE loss upon application to test-sets. }
  \label{tab:lamfit}
  \end{table}

$M_0$ denotes a control model defined by $(\lambda_0,\lambda_2,\lambda_\text{NL})=(0,0,0)$, which we present to show the MSE loss before optimizing. The results show that adding a new local term to the model, i.e. going from $M_1$ to $M_2$, significantly improves the accuracy of the EFT at predicting $\Delta G^{(4)}$ in all three architectures. This can be seen by comparing the MAPE and MSE results, both measurements of error, which decrease when moving to model $M_2$ from model $M_1$. The nonlocal model $M_3$ only improves test set fitting in ReLU-net. Since the model $M_3$ consists of only one of many possible nonlocal quartic EFT terms, we claim this type of term only improves the local ReLU-net EFT away from the train set. This suggests that some of the functional dependence of $\Delta G^{(4)}$ for ReLU-net is captured by the nonlocal term in $M_3$. Despite still being reasonably good models, the other two architectures become less accurate when this term is added, which can be seen by the increased MAPE and MSE errors of model $M_3$ compared with $M_1$ and $M_2$ in \gnet~and Erf-net. We claim this is because $\Delta G^{(4)}$ for these networks have a nonlocality that is not well described by the final term in $M_3$; their functional dependence is more complicated than the models we have tested here. 
Deeper analysis of any given network architecture can be done to find a more predictive local or nonlocal EFT term. We will do this in future work.

}

\section{Minimal Non-Gaussian Process Likelihoods with Wilsonian Renormalization} \label{sec:rg}

In Section \ref{sec:EFT} we demonstrated that 
we can use a Wilsonian EFT approach to neural networks to make 
experimentally verifiable predictions for correlation functions of the associated
NGPs. This procedure can be carried out for any fixed cutoff $\Lambda$ that yields
a perturbative NGP.
In doing so we claimed that we could ignore higher-than-quartic terms
in the NGP effective action, and promised to justify the assumption. We do so now,
 and will also explain and experimentally demonstrate the relationship between the EFTs at different values of
 $\Lambda$.

 The correct physics framework for these ideas is the Wilsonian renormalization group. This seminal
 idea in quantum field theory has led to some of the deepest results in both condensed matter
 and high energy physics, including multiple Nobel prizes, and is the subject of extensive
 discussions and presentation in any modern QFT textbook. We will present a streamlined perspective
 on it that is relevant for understanding applications in neural networks.

 Our explanation of the renormalization group applied to neural network NGPs will be grounded in
 experiments\footnote{A wonderful
 physics-version of this discussion is in Part III.1 of \cite{Zee}.},
 which applies in particular  to the sort of neural network experiments that we have carried out. 
 Consider any experiment that draws an ensemble
 of neural networks $f_\alpha$ 
 and computes their outputs
 on a fixed set of inputs
 \begin{equation}
\mathcal{S}_{\text{in}} = \{x_1,\dots, x_{N_\text{in}} \}.
\end{equation}
Given these measured outputs, the experimental correlation functions are measured via
\begin{equation}
G^{(n)}(x_1,\dots, x_n) = \frac{1}{n_\text{nets}} \sum_{\alpha\in \text{nets}}^{n_\text{nets}}\, f_\alpha(x_1)\, \dots \, f_\alpha(x_n),
\end{equation}
which contain essential information about the distribution of neural networks
induced by the chosen architecture and parameter distribution, as well as data if they have been trained.
These results are concrete experimental facts, and the goal of theory is to explain them.

To that end, noting that wide varieties \cite{yang} of both trained and untrained networks admit a limit
in which they are drawn from Gaussian processes, we propose constructing theories of the neural
networks away from the GP limit by perturbing it to an NGP, encoded in
\begin{equation}
\label{eqn:DelSInfinity}
\Delta S = \int^\infty_{-\infty} d^{d_\text{in}} x\, \sum_{l\leq k} g_{\cO_l} \,\cO_l,
\end{equation}
where the sum is over all operators $\cO_l$ of scaling dimension $l \leq k$ for some fixed $k$ and
the operators are themselves functions of neural network outputs; they are monomials in our cases.
This yields an action for the NGP, $S_{\text{NGP}}$, with distribution
\begin{equation}
P[f] = e^{-S_\text{NGP}[f]}.
\end{equation}
Using standard techniques reviewed in Appendix \ref{app:gaussian} and carried out in Section \ref{sec:EFT}, correlation functions $G^{(n)}(x_1,\dots,x_n)$
may be computed perturbatively in the coefficients $g_{\cO_l}$ and matched to experiments, where
the GP kernel plays a crucial role, although GP predictions for $n$-pt functions are modified by 
various ``interaction'' terms of strength $g_{\cO_l}$.

\bigskip
However, in carrying out such computations one regularly encounters divergences arising 
from integrals over the space of inputs to $f$, which are neural network inputs in our framework, 
and usually
position or momentum space in QFT. Needless to say, these divergences prevent the 
theoretical correlation functions from matching the finite ones measured in experiments.
In QFT these are infinities that were ``swept under the rug'' in complaints from the 1960s,
but this viewpoint is archaic and incorrect from a modern perspective, since the issue
is solved by a proper understanding of renormalization, particularly the one developed by Wilson.

Since the divergences arise from the boundaries of the $\Delta S$ integral being $\pm \infty$,\footnote{\amnew{This is not an issue with \gnet~since its kernel tends toward 0 in a way where all EFT integrals are finite, as will be discussed later.}}
it is natural to attempt to get rid of them by cutting them off,
as was physically motivated in Section \ref{eftsec}. That is, replace \eqref{eqn:DelSInfinity}
by
\begin{equation}
	\label{eqn:DelSLambda}
	\Delta S_\Lambda = \int^\Lambda_{-\Lambda} d^{d_\text{in}} x\, \sum_{l\leq k} g_{\cO_l}(\Lambda) \,\cO_l.
\end{equation}
\amnew{In the QFT analogy, this refers to the finite width NNs receiving corrections to the GP in terms of interactions involving generation and annihilation of ``virtual particles / excitations'' that spans the input space between $\{-\Lambda, \Lambda\}$. In case of real-life neural networks, an example of cutoff scale and its insignificance to actual NN correlation functions can be understood through an MNIST with greyscale handwritten digits. Let the input space be $\mathbb{R}^{28 \times 28}$, with the real numbers corresponding to each of $28 \times 28$ pixels being $-\infty$, $0$ and $\infty$ respectively at pure black, pure grey and pure white colors; i.e. the real number represents brightness. Images in the dataset that are grey correspond to experimental inputs close to zero. Computing NGP correlation functions via perturbation theory, integrating from pure black to pure white colors, may introduce divergences; a way to regulate this is by limiting the integration variable within some threshold black and white colors, leading to $\Delta S \rightarrow \Delta S_\Lambda$. This threshold corresponds to a maximum brightness and maximum darkness scale, which in a Wilsonian sense is unimportant as a slight change in very high threshold amount of black and white colors cannot meaningfully affect correlation functions associated with grey images, i.e. inputs near zero. Thus, high scale brightness and darkness thresholds in MNIST are analogous to high scale cutoffs in EFT.} 

Doing this for some fixed value of $\Lambda$ such that $|x_i| \ll \Lambda$ for all $x_i \in \mathcal{S}_\text{in}$, one may use the experimental results for $G^{(n)}(x_1,\dots,x_m)$ to extract 
the values of $g_{\cO_l}(\Lambda)$ at that value of $\Lambda$, which may then be used to make 
other predictions that may be experimentally verified; this is what we did in Section \ref{sec:EFTforSingle}.
 In general, since the operators are truncated 
at dimension $l\leq k$ there are a finite number of $g_{\cO_l}(\Lambda)$, yet an infinite
number of $n$-pt functions receive corrections from them, i.e., we have a finite number
of quantities are making an infinite number of predictions.

We now arrive at the essence of the Wilsonian renormalization group. In passing from 
\eqref{eqn:DelSInfinity} to \eqref{eqn:DelSLambda}, we passed from one $\Delta S$
to a one-parameter family  $\Delta S_\Lambda$ with parameter $\Lambda$. Yet there is one set 
of experimental $n$-pt functions and a continuous infinity of theories associated with $\Delta S_\Lambda$, each of which
by the above process may make correct predictions for the experiments. It is clear,
therefore, that
\begin{equation}
	\label{eqn:dGdlamzero}
\frac{dG^{(n)}(x_1,\dots, x_n)}{d\Lambda} = 0.
\end{equation}
Applying this to the theoretically computed $n$-pt functions yields differential
equations that describe how the couplings $g_{\cO_l}(\Lambda)$ vary as $\Lambda$ is varied. 
This is to be expected: if we have an EFT making correct predictions at some
value of $\Lambda$ with $x_i \ll \Lambda$, we should be able to do the same thing
for infinitesimally shifted cutoff $\Lambda + \delta \Lambda$, which will lead
to slightly different couplings  $g_{\cO_l}(\Lambda+ \delta \Lambda)$. 
The differential equations are referred to as renormalization group equations
(RGEs), which include the $\beta$-function associated to the coupling, defined as
\begin{equation}
\beta(g_{\cO_l}) := \frac{d\,g_{\cO_l}}{d\,log\Lambda},
\end{equation}
which may be extracted from \eqref{eqn:dGdlamzero}. The RGEs give rise to a
flow in the couplings induced by varying $\Lambda$, known 
as renormalization group (RG) flow.

Now consider an RG flow in the space of couplings that begins at a point in coupling
space where 
they are all non-trivial. Given this initial condition, the $\beta$-functions
determine a trajectory through coupling space and the opposite directions along the
trajectory correspond to raising and lowering $\Lambda$. Couplings 
that decrease, increase, or stay the same along one direction of the flow are said to be \emph{irrelevant}, \emph{relevant}, and \emph{marginal}, respectively; their corresponding operators inherit the same names. The names are clear: go far enough along
an RG flow and the irrelevant couplings go to zero, they may be ignored. Upon switching the direction
of the flow, the names are reversed, though in physics one often considers 
unidirectional flows to low energies (long distances) since at higher energies the
flows may be altered by the existence of a yet-undiscovered particle.

\medskip
Consider the effective action \eqref{eq:eft-action} in light
of this discussion of irrelevant, relevant, and marginal operators.
It contains 4-pt and 6-pt couplings only, $\lambda$ and $\kappa$. From \eqref{eq:gdim}, the couplings associated have input-space dimensions 
\be
\label{opdimana}
[\lambda] = [g_{4}] = -d_\text{in} - 2[K], \qquad [\kappa] = [g_{6}] = - d_\text{in} - 3[K],   
\ee
respectively, where $[K]$ denotes the scaling dimension of the NN kernel in the GP limit. For $[K] \geq 0$, both $\lambda$ and $\kappa$ are irrelevant (in the large $\Lambda$ limit) due to having negative dimensions. If $[K] = 0$, they decrease at the same rate since they would both scale with the same dimension $-[d_\text{in}]$. \amnew{When $[K] > 0$}, $\kappa$ decreases more quickly
than $\lambda$ as the cutoff increases, since $[\kappa] < [\lambda]$; and in the limit of large cutoff, the $n$-pt functions receive negligible corrections from $\kappa$, in comparison to corrections due to $\lambda$,
and one may effectively ignore $\kappa$.

\jimnew{We emphasize that the dimension of the couplings depends explicitly on $\din$. This is analogous to a well-known effect from quantum field theory, where the dimension of spacetime affects the RG flow; similarly, $\din$ affects the flow in neural networks. We will see this explicitly in the example of ReLU-net, where $[\lambda]=-\din-4$, which will appear explicitly in slopes of appropriate RG experimental plots, as well as theoretical analysis. Throughout this work $\dout=1$, so we will leave associated experiments to future work, but in analogy to quantum field theory it is clear that changing the number of scalar fields --- changing $\dout$ --- can affect the RG flow. For instance, in QFT the location of the Wilson-Fisher fixed point depends on the number of scalar fields $N$.}

Given that $\lambda$ is irrelevant, one may wonder why it cannot
also be ignored. The proper prescription is that the leading operators necessary to account for a given phenomenon must be included. If we ignored \emph{all} couplings $g_k$ since they
are irrelevant, we would be back in the GP limit and unable to account for the finite width experiments. To explain the experiments, one must keep the most relevant coupling; $g_3=0$ since the $3$-pt function is zero, and therefore $\lambda = g_4$ is the leading coupling, despite being termed irrelevant. 

As an example of this phenomenon in physics,
consider an effective field theory of the blue sky, which describes
the scattering of light off of neutral spin-$0$ particles in the atmosphere. The lowest dimension operator that can describe this scattering process is
\begin{equation}
\frac{c}{\Lambda^2} \, \phi^* \phi F_{\mu\nu}F^{\mu\nu},
\end{equation}
where the $\phi$ field and its conjugate represent the spin-$0$ particles and $F$ is the gauge-invariant field strength tensor of electromagnetism associated with the photon. We have used the particle physics convention of writing 
the coupling as a dimensionless $c$ with the cutoff explicitly
introduced. This operator is irrelevant, but since it is the
lowest dimension operator that gives rise to gauge-invariant scattering of light off of neutral atoms, it must be included in the EFT. In fact, it reproduces the Rayleigh cross-section that explains why blue light scatters more strongly
than red light.

\subsection{Neural Network Non-Gaussian Process Flows with $\beta$-functions}

Having introduced the central ideas of the renormalization group, we now address more concretely how these $\beta$-functions
may be extracted from cutoff-dependent correlation functions, organizing the calculations according to model (architecture) 
independent and dependent pieces. This split arises naturally for some architectures. For instance, some terms in the kernels are shared amongst all single layer
fully-connected networks, while others depend on the specific architecture, as determined by the activation function.

The kernels associated to a class of neural network architectures can be expressed in terms of a model independent (within the architecture class) term $\alpha$ and a model dependent term $\varsigma$.
\begin{eqnarray}
K(x,x') = \alpha + \varsigma(x,x'), \label{generalkernelrearrange}
\end{eqnarray}
where we assumed that the first and second terms are input independent and dependent, respectively, since it is true in the networks we study and more generally to deep fully-connected networks; it is straightforward to study the case where the first-term is input-dependent, as well.
For instance, the kernels in Section \ref{sec:networksetup} allow for explicit comparison.

Substituting this form of the kernel into the $4$-pt function in \eqref{full4-ptdiagrams} lets us rewrite the $G^{(4)}(x_1,x_2,x_3,x_4)$ in terms of model independent terms $\gamma_{4,i}$ and model dependent terms $\varrho_{4,i}$, for simplicity analyzing the case where $\lambda$ and $\kappa$ are approximately constants. The subscripts $4$ and $i$ refer to the order of the correlation function and $O(i)$ corrections to the GP expression respectively. 
\bea
G^{(4)}(x_1,x_2,x_3,x_4) = \gamma_{4,0} + \varrho_{4,0} - \lambda \int^{\Lambda}_{-\Lambda} \,d^{d_\text{in}}x\, (\gamma_{4, \lambda} + \varrho_{4,\lambda}) -  \kappa \int_{-\Lambda}^{\Lambda} \,d^{d_\text{in}}x\, (\gamma_{4,\kappa} + \varrho_{4,\kappa}) \label{full4-pt} 
\eea
Similarly \eqref{generalkernelrearrange} can be substituted into the $6$-pt function given in \eqref{6-ptlambda} to express it in terms of model independent terms $\gamma_{6,i}$ and model dependent terms $\varrho_{6,i}$, as the following
\bea
G^{(6)}(x_1,x_2,x_3,x_4, x_5, x_6) = \gamma_{6,0} + \varrho_{6,0} - \lambda \int^{\Lambda}_{-\Lambda} \,d^{d_\text{in}}x\, (\gamma_{6, \lambda} + \varrho_{6,\lambda}) -  \kappa \int^{\Lambda}_{-\Lambda} \,d^{d_\text{in}}x\, (\gamma_{6,\kappa} + \varrho_{6,\kappa}). \label{full6-pt}
\eea
Irrespective of the NN structure, the terms $\gamma_{n,0}$ and $\varrho_{n,0}$ are independent of the integration variable $x$, as they are tree level corrections to the $n$-pt functions, and independent of any interaction vertices in respective Feynman diagrams.

The RG equations can be obtained by taking derivatives of the $4$-pt and $6$-pt function with respect to $\log$ of the cutoff scale $\Lambda$,
\bea 
\frac{ \partial G^{(4)}(x_1,x_2,x_3,x_4)}{\partial \log \Lambda} &=& 0 \,=\, \frac{ \partial \lambda}{\partial \log \Lambda} \, \int^{\Lambda}_{-\Lambda} d^{d_\text{in}}x\, (\gamma_{4, \lambda} + \varrho_{4,\lambda})\,  +\lambda \, \frac{ \partial  (\int^{\Lambda}_{-\Lambda} d^{d_\text{in}}x\, (\gamma_{4, \lambda} + \varrho_{4,\lambda}) )}{\partial \log \Lambda}  \nonumber \\
&+& \frac{ \partial \kappa}{\partial \log \Lambda} \, \int^{\Lambda}_{-\Lambda} d^{d_\text{in}}x\, (\gamma_{4, \kappa} + \varrho_{4,\kappa})\,  +\kappa \, \frac{ \partial  (\int^{\Lambda}_{-\Lambda} d^{d_\text{in}}x\, (\gamma_{4, \kappa} + \varrho_{4,\kappa} )) }{\partial \log \Lambda} \,\, , \label{RG4-pt}
\eea
\bea 
\frac{ \partial G^{(6)}(x_1,x_2,x_3,x_4,x_5,x_6)}{\partial \log \Lambda} &=& 0 \,=\, \frac{ \partial \lambda}{\partial \log \Lambda} \, \int^{\Lambda}_{-\Lambda} d^{d_\text{in}}x\, (\gamma_{6, \lambda} + \varrho_{6,\lambda})\,  +\lambda \, \frac{ \partial  (\int^{\Lambda}_{-\Lambda} d^{d_\text{in}}x\, (\gamma_{6, \lambda} + \varrho_{6,\lambda}) )}{\partial \log \Lambda}  \nonumber \\
&+& \frac{ \partial \kappa}{\partial \log \Lambda} \, \int^{\Lambda}_{-\Lambda} d^{d_\text{in}}x\, (\gamma_{6, \kappa} + \varrho_{6,\kappa})\,  +\kappa \, \frac{ \partial  (\int^{\Lambda}_{-\Lambda} d^{d_\text{in}}x\, (\gamma_{6, \kappa} + \varrho_{6,\kappa} )) }{\partial \log \Lambda} \,\, . \label{RG6-pt}
\eea
In the limit of large $\Lambda$, $\kappa$ is negligible, and the last terms in \eqref{RG4-pt} and \eqref{RG6-pt} vanish. In that case, \eqref{RG4-pt} and \eqref{RG6-pt} can be simplified to the following 
\bea
\frac{ \partial \lambda}{\partial \log \Lambda} \, \int^{\Lambda}_{-\Lambda} d^{d_\text{in}}x\, \Bigg[\frac{\int^{\Lambda}_{-\Lambda} d^{d_\text{in}}x\, (\gamma_{6, \lambda} + \varrho_{6,\lambda})}{\int^{\Lambda}_{-\Lambda} d^{d_\text{in}}x\, (\gamma_{6, \kappa} + \varrho_{6,\kappa})}    -  \frac{\int^{\Lambda}_{-\Lambda} d^{d_\text{in}}x\, (\gamma_{4, \lambda} + \varrho_{4,\lambda})}{\int^{\Lambda}_{-\Lambda} d^{d_\text{in}}x\, (\gamma_{4, \kappa} + \varrho_{4,\kappa})}   \Bigg] \,  + \nonumber  \\
\lambda \, \Bigg[\frac{1}{\int^{\Lambda}_{-\Lambda} d^{d_\text{in}}x\, (\gamma_{6, \kappa} + \varrho_{6,\kappa})}  \frac{ \partial  (\int^{\Lambda}_{-\Lambda} d^{d_\text{in}}x\, (\gamma_{6, \lambda} + \varrho_{6,\lambda}) )}{\partial \log \Lambda} 
-  \frac{1}{\int^{\Lambda}_{-\Lambda} d^{d_\text{in}}x\, (\gamma_{4, \kappa} + \varrho_{4,\kappa})}  \frac{ \partial  (\int^{\Lambda}_{-\Lambda} d^{d_\text{in}}x\, (\gamma_{4, \lambda} + \varrho_{4,\lambda}) )}{\partial \log \Lambda}   \Bigg] = 0    \nonumber 
\label{RGflow} 
\eea
Solving the above gives us the RG equation of $\lambda$, $\beta(\lambda)$.

\amnew{
\subsection{Renormalization Analysis for Fully-Connected Networks}
\label{sec:expRGrules}

In this Section we specify the renormalization analysis of the previous Section to the case of a network with linear output layer and Gaussian biases, which include fully-connected networks and specifically the single-layer networks of our experiments. 

Doing so is important for clarity because, as emphasized in Section \ref{sec:exeft}, the architectures we study are drawn from processes $\mathcal{P}$ of functions of the form $f = f_b + f_W$, where $f_b$ and $f_W$ are drawn from independent processes $f_b \sim \mathcal{P}_b$ and $f_W \sim \mathcal{P}_W$, with $\mathcal{P}_b$ Gaussian. The associated log-likelihood correction is therefore $\Delta S = \Delta S_W = \int^\infty_{-\infty} d^{\din} x \sum_{l \leq k} g_{\mathcal{O}_l} \mathcal{O}_l$, which contributes to the distribution $P_W = e^{-S_{\text{NGP}} [f_W] }$; that is, all the non-Gaussianities involve only $f_W$, not $f_b$. Divergences may arise in perturbation theory from the boundaries of $\Delta S_W$ being $\pm \infty$, but can be removed using a cutoff scale $\Lambda$, yielding
\begin{eqnarray}
\Delta S_{W, \Lambda} = \int^\Lambda_{-\Lambda} d^{\din} x \sum_{l \leq k} g_{\mathcal{O}_l} \mathcal{O}_l,
\end{eqnarray}
which serves as the basis for perturbation theory and renormalization for any fixed $\Lambda$.

 Input-space dimensions of $4$-pt and $6$-pt couplings $\lambda$ and $\kappa$ in \eqref{eq:eft-action} are now
\be
\label{exopdim}
[\lambda] = [g_{4}] = -d_\text{in} - 2[K_W], \qquad [\kappa] = [g_{6}] = - d_\text{in} - 3[K_W],   
\ee
respectively, where $[K_W]$ denotes scaling dimension of the kernel associated with $\mathcal{P}_W$ in GP limit.} 
\amnew{
Since $K_W$ is the only part of the kernel that may give rise to divergences when computing correlation functions perturbatively, as only $f_W$ appears in $\Delta S_W$, we focus on the non-Gaussian process flows in terms of $K_W$. For a general class of neural network architectures
\begin{eqnarray}
K_W(x,x') = \varsigma(x,x'), \label{exkernelrearrange}
\end{eqnarray}
which will be utilized in concrete analyses.

A crucial difference from equations \eqref{full4-pt} and \eqref{full6-pt} is that these examples do not have model-independent terms from output biases in the interaction contributions to the correlation functions. Specifically,
we can rewrite $G^{(4)}(x_1,x_2,x_3,x_4)$ in terms of a model-independent term $\gamma_{4,0}$ (that does not arise from interactions) and model dependent terms $\varrho_{4,i}$. For simplicity we analyze the case where $\lambda$ and $\kappa$ are approximately constants. The subscripts $4$ and $i$ refer to the order of the correlation function and $O(i)$ corrections to the GP expression respectively. 
\bea
G^{(4)}(x_1,x_2,x_3,x_4) = \gamma_{4,0} + \varrho_{4,0} - \lambda \int^{\Lambda}_{-\Lambda} \,d^{d_\text{in}}x\, \varrho_{4,\lambda} -  \kappa \int_{-\Lambda}^{\Lambda} \,d^{d_\text{in}}x\, \varrho_{4,\kappa} \label{ex-full4-pt} 
\eea
Similarly the $6$-pt function can be expressed in terms of a model independent term $\gamma_{6,0}$ (that does not arise from interactions) and model dependent terms $\varrho_{6,i}$, as the following
\bea
G^{(6)}(x_1,x_2,x_3,x_4, x_5, x_6) = \gamma_{6,0} + \varrho_{6,0} - \lambda \int^{\Lambda}_{-\Lambda} \,d^{d_\text{in}}x\, \varrho_{6,\lambda} -  \kappa \int^{\Lambda}_{-\Lambda} \,d^{d_\text{in}}x\,  \varrho_{6,\kappa}. \label{ex-full6-pt}
\eea
Irrespective of the NN structure, the terms $\gamma_{n,0}$ and $\varrho_{n,0}$ are independent of the integration variable $x$, as they are tree level corrections to the $n$-pt functions, and independent of any interaction vertices in respective Feynman diagrams.

The RG equations can be obtained by taking derivatives of above $4$-pt and $6$-pt function with respect to $\log$ of the cutoff scale $\Lambda$,
\bea 
\frac{ \partial G^{(4)}(x_1,x_2,x_3,x_4)}{\partial \log \Lambda} = 0 &=& \frac{ \partial \lambda}{\partial \log \Lambda} \, \int^{\Lambda}_{-\Lambda} d^{d_\text{in}}x\,  \varrho_{4,\lambda}\,  +\lambda \, \frac{ \partial  (\int^{\Lambda}_{-\Lambda} d^{d_\text{in}}x\, \varrho_{4,\lambda} )}{\partial \log \Lambda}  + \frac{ \partial \kappa}{\partial \log \Lambda} \, \int^{\Lambda}_{-\Lambda} d^{d_\text{in}}x\, \varrho_{4,\kappa}\,  \nonumber \\ &+& \kappa \, \frac{ \partial  (\int^{\Lambda}_{-\Lambda} d^{d_\text{in}}x\, \varrho_{4,\kappa} ) }{\partial \log \Lambda} \,\, , \label{exRG4-pt}
\eea
\bea 
\frac{ \partial G^{(6)}(x_1,x_2,x_3,x_4,x_5,x_6)}{\partial \log \Lambda} = 0 &=& \frac{ \partial \lambda}{\partial \log \Lambda} \, \int^{\Lambda}_{-\Lambda} d^{d_\text{in}}x\,  \varrho_{6,\lambda}\,  +\lambda \, \frac{ \partial  (\int^{\Lambda}_{-\Lambda} d^{d_\text{in}}x\, \varrho_{6,\lambda} )}{\partial \log \Lambda}  + \frac{ \partial \kappa}{\partial \log \Lambda} \, \int^{\Lambda}_{-\Lambda} d^{d_\text{in}}x\, \varrho_{6,\kappa}\,  \nonumber \\ &+& \kappa \, \frac{ \partial  (\int^{\Lambda}_{-\Lambda} d^{d_\text{in}}x\, \varrho_{6,\kappa} ) }{\partial \log \Lambda} \,\, , \label{exRG6-pt}
\eea
In the limit of large $N$, and potentially of large $\Lambda$, we may ignore $\kappa$ and equations \eqref{exRG4-pt} and \eqref{exRG6-pt} can be simplified to 
\bea
\frac{ \partial \lambda}{\partial \log \Lambda} \, \int^{\Lambda}_{-\Lambda} d^{d_\text{in}}x\, \Bigg[\frac{\int^{\Lambda}_{-\Lambda} d^{d_\text{in}}x\,  \varrho_{6,\lambda}}{\int^{\Lambda}_{-\Lambda} d^{d_\text{in}}x\,  \varrho_{6,\kappa}}    -  \frac{\int^{\Lambda}_{-\Lambda} d^{d_\text{in}}x\,  \varrho_{4,\lambda}}{\int^{\Lambda}_{-\Lambda} d^{d_\text{in}}x\, \varrho_{4,\kappa}}   \Bigg] \,  + 
\lambda \, \Bigg[\frac{1}{\int^{\Lambda}_{-\Lambda} d^{d_\text{in}}x\, \varrho_{6,\kappa}}  \frac{ \partial  (\int^{\Lambda}_{-\Lambda} d^{d_\text{in}}x\,  \varrho_{6,\lambda}) }{\partial \log \Lambda} 
- \nonumber  \\  \frac{1}{\int^{\Lambda}_{-\Lambda} d^{d_\text{in}}x\, \varrho_{4,\kappa} }  \frac{ \partial  (\int^{\Lambda}_{-\Lambda} d^{d_\text{in}}x\, \varrho_{4,\lambda}) }{\partial \log \Lambda}   \Bigg] = 0   \,. \nonumber \label{exRGflow} 
\eea
Solving the above gives us the RG equation of $\lambda$, $\beta(\lambda)$ in our examples.

}

\subsection{Experiments: Flows in Single-Layer Networks} \label{sec:rge}
In this subsection we analyze the flow in the coupling $\lambda$ according to the techniques introduced in Section \ref{sec:expRGrules}. 

\subsubsection*{\gnet}

Recall that the \gnet~kernel is given by
\begin{eqnarray}
   K_{\text{Gauss}}(x, x') = \sigma_{b}^2 + \sigma_{W}^2\,\,\exp\left[- \frac{\sigma_{W}^{2}|x - x^\prime |^2}{2 \din} \right].
 \end{eqnarray}  
 Note that $|K_\jimnew{W}(x, x')| < 1 ~\forall~x\neq x';~\text{and}~ |K_\jimnew{W}(x,x') | = 1 ~\text{when}~ x= x'$.  \amred{At large $x$ and fixed $x'$ (which arises, for instance, when $x$ is an integration variable and $x'$ an external input) $K_W$ exponentially dies off to zero, leading to convergent interaction integrals.

At small input $x'$, a careful analysis using Taylor expansion shows that in the regimes we are studying, i.e. small $x$, $[K_\jimnew{W}]=\epsilon <0$. }
The scaling of the couplings can be obtained from \eqref{exopdim}, giving $[\lambda] = - d_\text{in} - 2\epsilon$ and $[\kappa] = - d_\text{in} - 3\epsilon$. This suggests that $\lambda$ scales to zero slightly faster than $\kappa$, but we see in our $6$-pt experiments that the $\lambda$ contributions nevertheless dominate over the $\kappa$ contributions.  Therefore at sufficiently large cutoff, $\kappa$ becomes negligible, giving another reason to ignore it beyond the parametric control in $1/N$.  Thus, in the limit of large cutoff scales, where $\kappa$ becomes negligible, the variations in $4$-pt functions of \gnet~with respect to $\log(\Lambda)$ can be used to obtain the RG equation of $\lambda$ for \gnet. 

\amnew{
Due to the fast convergence of $K_W$ to $0$, corrections to $n$-pt functions  of \gnet~from $\Delta S_W$ are all finite, and do not require renormalization. Put differently, the integrals do not diverge in the limit $\Lambda\to \infty$. Though not necessary, it is still possible to introduce a cutoff and study the flow of couplings, e.g. $\beta(\lambda)$ by various choice of cutoffs $\Lambda$, it is just not particularly interesting because for sufficiently large $\Lambda$, the couplings no longer change. 

}

\begin{table}[t]
    \centering
    \begin{tabular}{|l|l|l|}
    \hline
        &
        $\din = 2$ &
        $\din = 3$ \\ \hline
    ReLU-net &
        \begin{tabular}[c]{@{}l@{}}
    \{$(0.5000, 0.5000),$ \\
    $(0.5000, 1.0000),$ \\
     $(1.0000, 0.5000),$  \\
     $(1.0000, 1.0000)$\}        
        \end{tabular} &
        \begin{tabular}[c]{@{}l@{}}$\{(0.2000, 0.2000, 0.2000),$ \\ $(1.0000, 1.0000, 0.2000), $\\ $(0.2000, 1.0000, 1.0000), $\\ $(1.0000, 0.2000, 1.0000)\}$\end{tabular} \\ \hline
    \end{tabular}
    \caption{Inputs $\{x_{i}\}$ for ReLU-net RG experiments in Section \ref{sec:rge} when $\din = 2, 3$.}
    \label{tab:din23parameters}
    \end{table}

\subsubsection*{ReLU-net}
Next we obtain the RG equation for ReLU-net architecture in the limit where $\kappa$ becomes negligible.

Recall that the ReLU kernel is given by
\begin{eqnarray}
   K_{\text{ReLU}}(x, x') &=& \sigma_{b}^2 + \sigma_{W}^2\,. \frac{1}{2\pi} \sqrt{(\sigma_{b}^2 + \frac{\sigma_{W}^{2}}{\din}\, x\cdot x)(\sigma_{b}^2 + \frac{\sigma_{W}^{2}}{\din}\, x^\prime\cdot x^\prime)}(\sin \theta + (\pi - \theta)\cos\theta ) , \nonumber \\
  \theta &=& \arccos \Bigg[\frac{\sigma_{b}^2 + \frac{\sigma_{W}^{2}}{\din}\, x\cdot x^\prime}{\sqrt{(\sigma_{b}^2 + \frac{\sigma_{W}^{2}}{\din}\, x\cdot x)(\sigma_{b}^2 + \frac{\sigma_{W}^{2}}{\din}\, x^\prime\cdot x^\prime)}} \Bigg] . \nonumber
\end{eqnarray} 
Comparing this with \eqref{exkernelrearrange},
\begin{eqnarray}
\jimnew{K_W(x,x')} = \varsigma(x,x') &=&  \sigma_{W}^2\,. \frac{1}{2\pi} \sqrt{(\sigma_{b}^2 + \frac{\sigma_{W}^{2}}{\din}\, x\cdot x)(\sigma_{b}^2 + \frac{\sigma_{W}^{2}}{\din}\, x^\prime\cdot x^\prime)}(\sin \theta + (\pi - \theta)\cos\theta ) \nonumber \\
&=& h_1(x,x')\, h_2(\theta) , \\
\text{where}~~~h_1(x,x') &=&  \sqrt{(\sigma_{b}^2 + \frac{\sigma_{W}^{2}}{\din}\, x\cdot x)(\sigma_{b}^2 + \frac{\sigma_{W}^{2}}{\din}\, x^\prime\cdot x^\prime)} , \nonumber \\
h_2(\theta) &=& (\sin \theta + (\pi - \theta)\cos\theta )  . \nonumber
\end{eqnarray}
$K_\jimnew{W}$ has been further separated into a bounded $\theta$-dependent component $h_{2}$ and an unbounded component $h_{1}$.

The $4$-pt function receives a large contribution from the $\lambda$-corrections when $x >{\frac{\sigma_b \, \sqrt{d_\text{in}}}{\sigma_W}}$ and $x^\prime$ is one of the inputs in our experiments; in this limit, which is relevant for our experiments since $\sigma_b=0$, $h_1(x,x')$ can be approximated as the following
\begin{eqnarray}
\label{reluscaling}
h_1(x,x') = \frac{\sigma_W^2}{d_\text{in}}|x|\, |x'| \,,
\end{eqnarray}
then the $4$-pt function becomes 
\amnew{
\bea
G^{(4)}(x_1,x_2,x_3,x_4)&=& \Lambda\text{-independent term} - \lambda \, \int^{\Lambda}_{-\Lambda}\,d^{d_\text{in}}x\, 24\,\frac{\sigma_W^8}{d^4_\text{in}}|x|^4|x_1||x_2||x_3||x_4|\, \prod_{i=1}^4\,h_2(\theta_i), \nonumber
\eea}\amred{where each $\theta_i$ denotes the angle between external vertex $x_i$ and internal vertex $x$.} The integral is carried out by rotating the input-space to the spherical coordinates, where the $|x|$ and angular integrals may be carried out separately, since the $\theta_i$ do not depend on the magnitude of $x$ or $x_i$. \amred{Each of $\theta_i$ remains bounded in $[0,2\pi]$ as $x$ is integrated across $(-\infty, \infty)$, and therefore, is not affected by the $x$-scaling. Thus each of $h_2(\theta_i)$ are bounded functions, and unaffected by the $x$-scaling; leading to the scaling dimension $\bigg[\prod_{i=1}^4 h_2(\theta_i)\bigg] = 0$.} 

Combining this with \eqref{reluscaling}, \amred{and the choice of $\sigma_b^2 =0$ in both first and final linear layers,} the scaling dimension of ReLU kernel is given by $[K(x, x')] = 2$ \amnew{and $[K_w(x, x')] = 2$}, when $x > {\frac{\sigma_b \, \sqrt{d_\text{in}}}{\sigma_W}}$ and $x'$ is one of the inputs in our experiments.
Following \eqref{exopdim}, the $4$-pt and $6$-pt couplings scale as $[\lambda] = -d_\text{in} - 4$ and $[\kappa] = -d_\text{in} - 6$ respectively. This implies that $\kappa$ becomes increasingly more negligible
with respect to $\lambda$ as $\Lambda$ becomes larger; $\kappa$ is also negligible at large $N$ due to being $1/N$ suppressed relative to $\lambda$. In particular, this implies that the RG equation for $\lambda$ can be entirely obtained in terms of $\mathcal{O}(\lambda)$ corrections to the $4$-pt function; i.e., the $\kappa$ corrections may be ignored. 

Evaluating the full $4$-pt function at $\mathcal{O}(\lambda)$ in large cutoff limit, we obtain 
\amnew{
\bea
G^{(4)}(x_1,x_2,x_3,x_4)&=&  \Lambda\text{-independent term} - \lambda \, j_1\Lambda^{d_\text{in}+4}  ,
\eea}where $j_1$ is obtained by redefining the products of external vertices with other constants; trivial check shows that it is independent of $\Lambda$, and therefore we need not compute it, because it will drop out of $\beta(\lambda)$. Taking derivatives, we can obtain the RG equation as the following
\amnew{ \bea
\frac{\partial \lambda}{\partial \log \Lambda} \Lambda^{\din + 4} + (\din +4)\lambda \Lambda^{\din + 4} &=& 0 \,,
\eea}which yields
\bea
\beta(\lambda) := \frac{\partial \lambda}{\partial \log(\Lambda)} = - (d_\text{in}+4)\lambda \,.
\eea
After integration we get
\bea \label{rgpredictrelu}
\log \lambda \, = \, -(d_\text{in}+4)\log \Lambda + p_1,
\eea 
where $p_1$ is the constant of integration.

We verify above RG equation against the log-log variations of experimental value(s) of $\lambda_{m}$ at different cutoff scales $\Lambda$. \eqref{eqn:lambdam} is used to obtain the rank-$4$ tensor $\lambda_{m}$ at different choice of cutoff scales, given in \eqref{cutoffs}. All independent tensor elements of $\lambda_{m}$ are plotted below against respective $\Lambda$, in a log-log format, at different values of $d_\text{in}$.
\begin{figure}[th]
    \centering
  \makebox[\textwidth][c]{
  \includegraphics[width=0.38\textwidth]{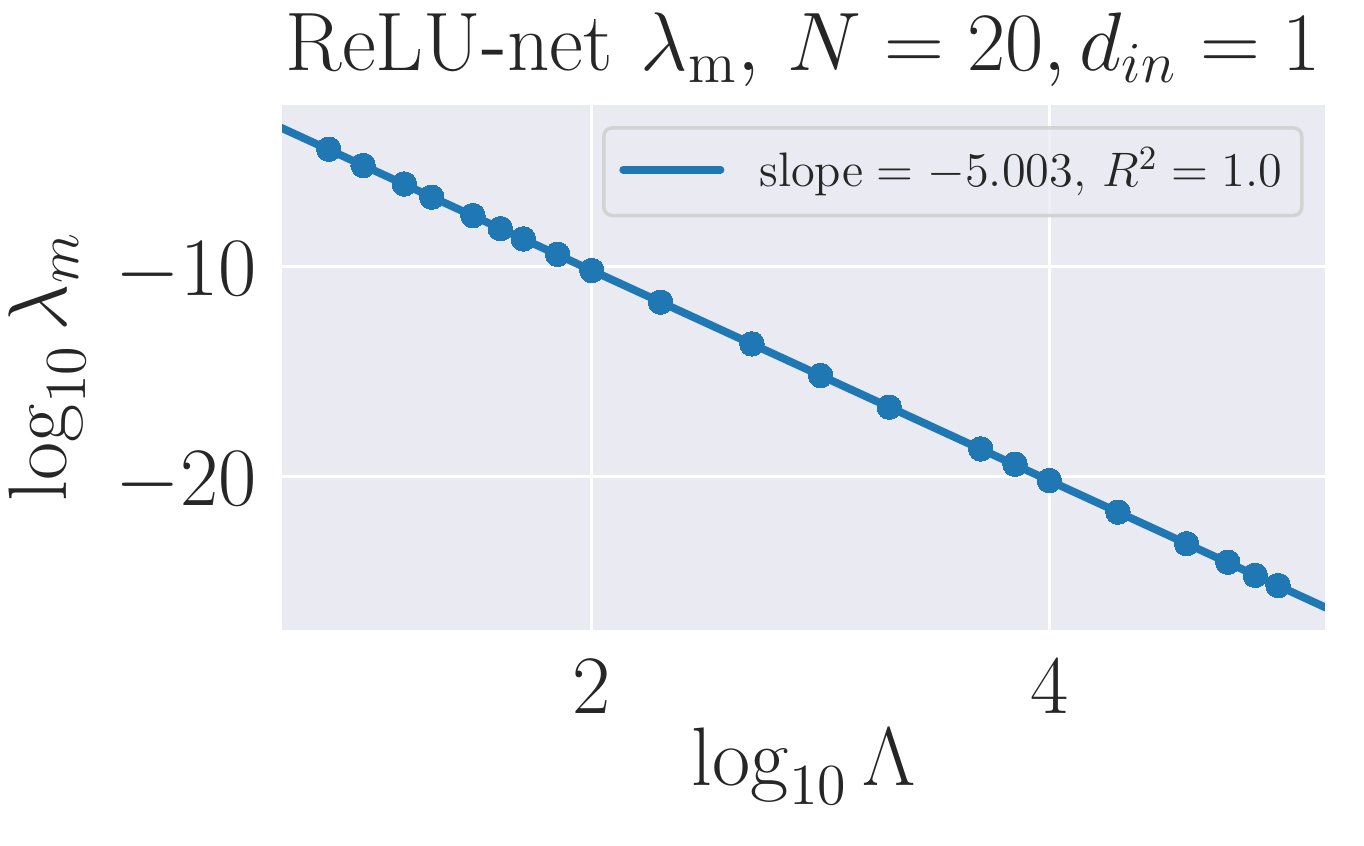}
  \includegraphics[width=0.38\textwidth]{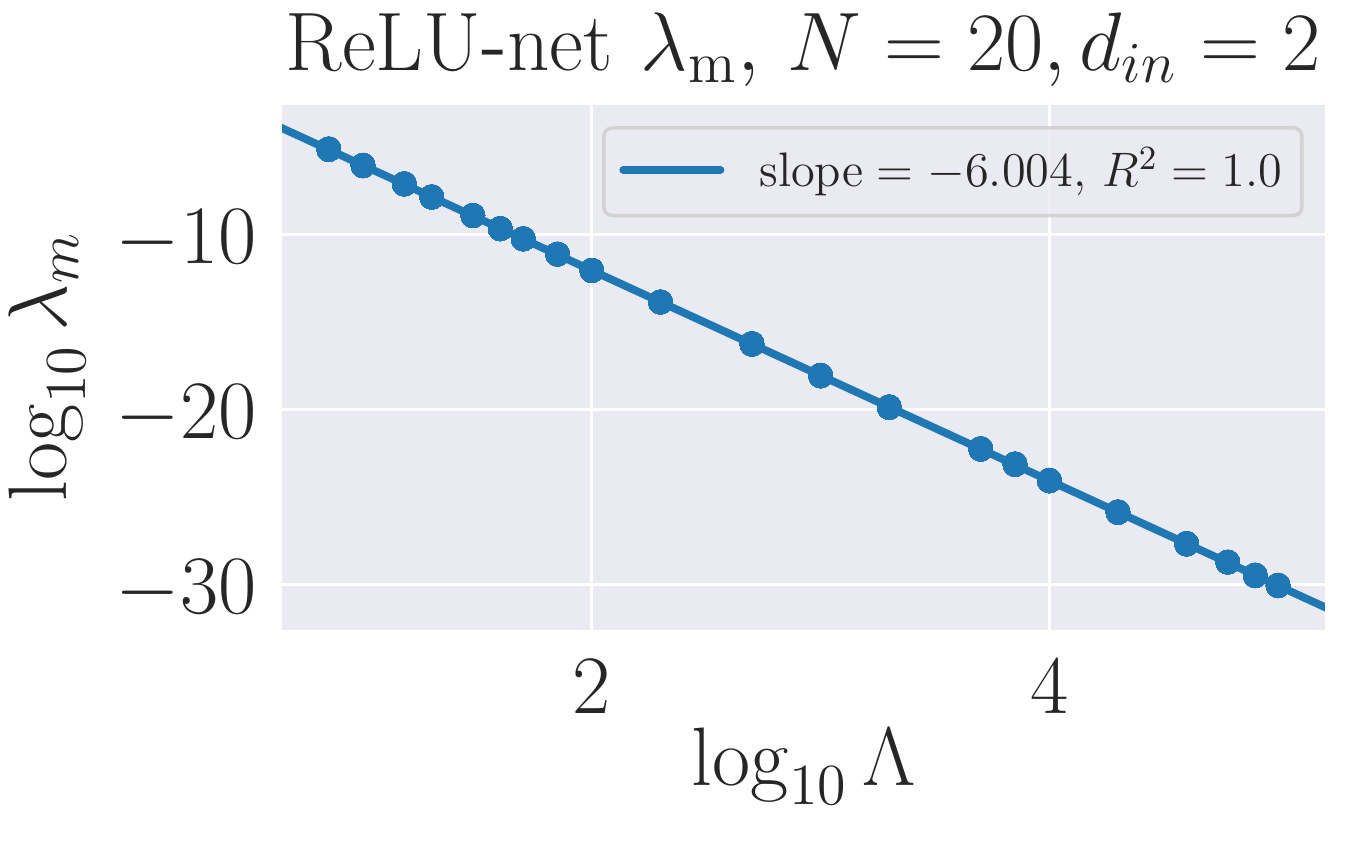}
  \includegraphics[width=0.38\textwidth]{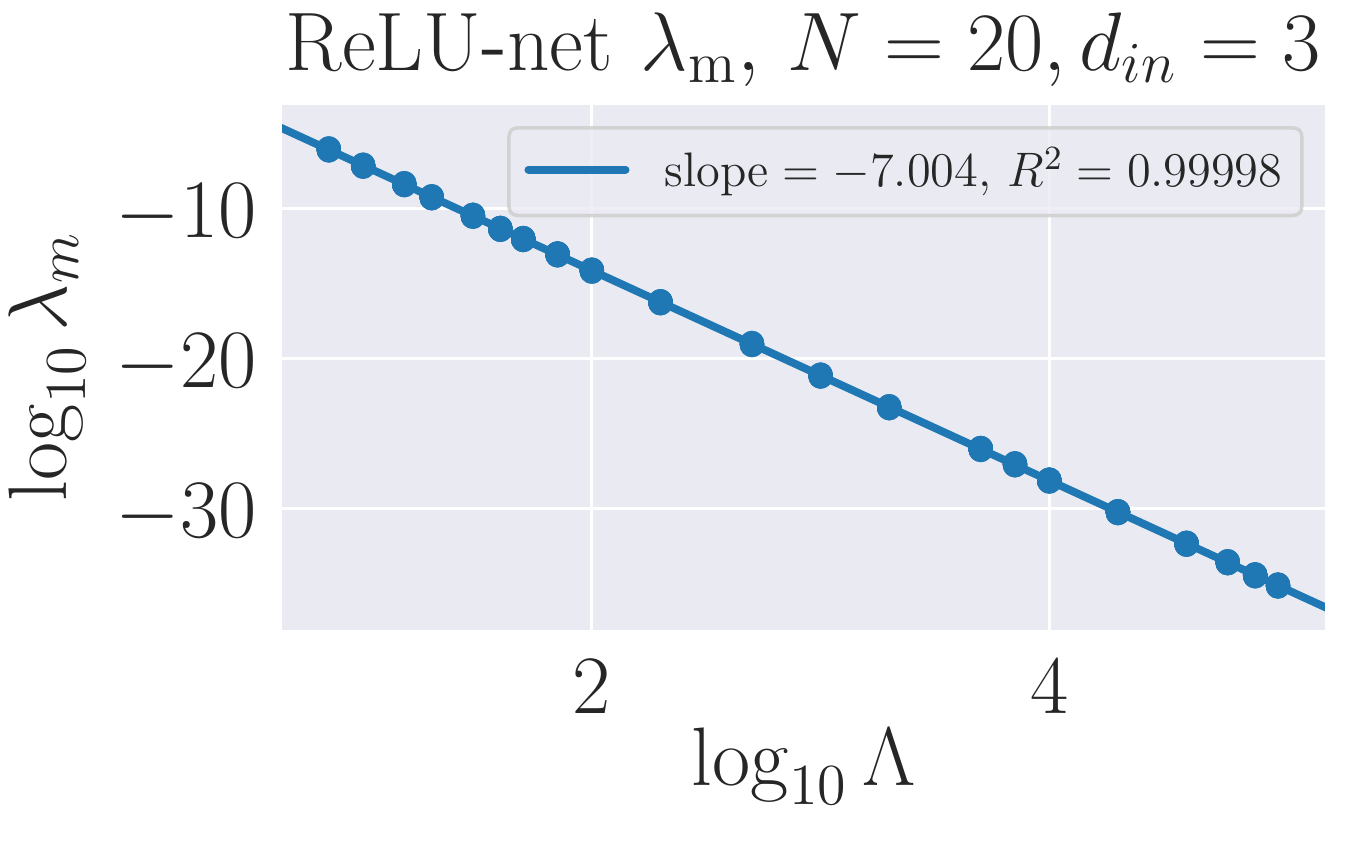}}
  \caption{RG equation for ReLU-net: $\log(\lambda_{m})$ as a function of $\log(\Lambda)$ for $d_\text{in} = 1, 2, 3$.}  
  \label{fig3.2}
  \end{figure}

The best fit lines in Fig. \ref{fig3.2} are in excellent agreement with the theoretical prediction at \eqref{rgpredictrelu}; indeed, the slopes are $\sim - (d_\text{in}+4)$ to excellent approximation. \amnew{We emphasize the explicit dependence on $\din$.} Here $\lambda$ appears to be a constant, but there is a small amount of variance on the same order displayed in Figure \ref{fig:reludelta} that is not visible due to the scale of the $y$-axis.

\section{Conclusions}

In this paper we have developed a correspondence between neural networks and quantum field theory, providing a theoretical understanding for a function-space approach to neural networks.

The central idea is to treat neural network architectures
$f_{\theta,N}$ that become Gaussian processes (GPs) in the $N\to \infty$ limit using techniques from 
Wilsonian effective field theory (EFT). Specifically, at finite $N$ the distribution on function space from which the
neural networks are drawn is no longer Gaussian; the GP distribution is corrected to an 
associated non-Gaussian process (NGP) distribution that may be treated using EFT. The corrections have coefficients known as couplings, in physics, that encode interaction strengths.

Overall, we focused our developments on the two central ideas in EFT, applied in the context of neural networks:
\begin{itemize}
\item \textbf{EFT is effective.} A finite number of measured couplings yield many verifiable predictions.
\item \textbf{EFT yields minimal NGP likelihoods.} Couplings vary with cutoff according to differential equations that govern flow on coupling space. Along the flow, some couplings become negligible, yielding a likelihood where all but the most important pieces can be safely ignored.
\end{itemize}
These were verified in concrete experiments in the simplest class of models admitting a GP limit: single layer fully-connected networks. We strongly emphasize, however, that our techniques are applicable to any architecture admitting a GP limit. Since we provided a concrete summary of our results in the introduction, we omit such a summary here.

\medskip
Instead, we would like to conclude \amnew{mentioning important areas for progress that we have not addressed}, and then making some general comments and providing outlook.

\amnew{Our work has developed a treatment of NN distributions in terms of QFT, focusing on the near-Gaussianity of the distributions at large-but-finite $N$. By focusing on the most essential elements of the correspondence, we have left a number of interesting and important directions for future work. One would be to understand when a local operator approach to NGP distributions, as we took in some of our models, suffices. Another is to understand the flow of the NGP distribution during training, perhaps from a neural tangent kernel \cite{Jacot2018NeuralTK} point of view. Finally, it would be interesting to understand how other types of fields --- e.g., fermionic fields or the vector fields of electromagnetism --- might arise in neural networks, where scalar fields are most natural.}

Physicist readers may not be familiar with thinking of neural networks from a function space point of view, despite the fact that they think this way in QFT. However, many have
trained a randomly initialized neural network, and then repeated it a number (say, $M$) of times to verify stability of the results. In doing so one is drawing $M$ untrained neural networks from the prior induced by the parameter distributions and network architecture, and training the networks turns them into draws from some other distribution, the trained distribution. From this perspective, the process of supervised learning is simply learning the one-point function of the trained distribution; given an ensemble of trained nets, one should generate predictions from the ensemble expectation value of the outputs, the one-point function. Due to the prevalence of GP limits for many different architectures, in both randomly initialized and appropriately trained networks, the desired distributions on function space are nearly Gaussian, and therefore amenable to QFT techniques.

In the ML literature, it was (borrowing the inspirational language) ``dreamed'' in \cite{Yaida2019NonGaussianPA} that flows in distributions of 
preactivations are akin to renormalization group flows; we would like to comment on this in light of our work. We
did not emphasize it so as to not introduce too many ideas,
but our techniques apply not only to GPs associated with network outputs, but also those that may arise at intermediate layers. For instance, in deep fully-connected networks the central limit theorem may be applied \cite{lee} not just to the network outputs, but also the preactivations of the hidden layer, leading to an EFT description of the NGP associated with each layer's preactivation. However, the details of the NGP effective action associated to each layer depends on both the previous layers' activation functions and the width of that layer; i.e., the number of preactivations, which is the dimension of the GP distribution or number of quantum fields in the correspondence. This gives rise to structural differences in the effective actions that go beyond a simple flow in the couplings, so in general the relationship between preactivations flows and RG flows is (again borrowing language) a ``spiritual resemblance,'' not an exact correspondence. Choosing all hidden layers to have the same non-linear activation function and width may make the analogy more direct.

Continuing down this direction, at a given layer of width $N$ one could write the NGP effective action (log-likelihood) such that the $N$ preactivations depend on the preactivations of a previous layer of width $M$, rather than the $d_\text{in}$ inputs to the overall network. In this case it would naively appear that the EFT for the $N$ preactivations would be in $M$-dimensional space, rather than $d_\text{in}$-dimensional space. These would be two EFT descriptions of the same preactivation NGP, but in different dimensions. What seems like a puzzle here (the arbitrariness of the dimensions) is likely resolved by a careful consideration of dimensionality, namely the first preactivation distribution having $d_\text{in}$-dimensional support in the $M$-dimensional space.

Equipped with a sharp NN-QFT correspondence, it is natural to try to import other ideas from QFT into the study of 
neural networks from a function space perspective. Perhaps those, together with the ideas implemented in this paper, can lead to a richer theoretical understanding of the 
many empirical successes in deep learning over the last decade.

\bigskip
\noindent \textbf{Acknowledgments.} We thank Michael Douglas and Greg Yang for suggestions on a draft of this
manuscript and Microsoft Research for supporting the ``Physics $\cap$ ML'' meeting in $2019$ that partially inspired this work. We thank David Berman, Liam Fitzpatrick, Sergei Gukov, Kiel Howe, Neil Lawrence, S\'{e}bastien Racani\`{e}re, Magnus Rattray, Danilo Jimenez Rezende, Fabian R\"{u}hle, Matt Schwartz, and Jamie Sully for helpful conversations. J.H. is supported by 
NSF CAREER grant PHY-1848089.

\clearpage 
\appendix

\appendix

\section{Review of Gaussian Integrals \label{app:gaussian}}

\subsection*{Single and Multivariable Cases}

All Gaussian integrals can be evaluated exactly as
\begin{eqnarray}
\int^{+\infty}_{-\infty} dx\,\exp{\left(-\frac12 ax^2\right)} = \left( \frac{2\pi}{a}\right)^{\frac12} .   \label{eq:norm}
\end{eqnarray} 
The associated $2n$-pt function is the expectation value of the single variable operator $x^{2n}$; it can be obtained by repeatedly differentiating the above by $-2(d/da)$ to give
\begin{eqnarray}
\langle x^{2n} \rangle = \frac{ \int^{+\infty}_{-\infty} dx\,\exp{(-\frac12 ax^2)}\,x^{2n} }{ \int^{+\infty}_{-\infty} dx\,\exp{(-\frac12 ax^2)}} = \frac{1}{a^n}(2n-1)!! \,\, . \label{eq:npoint}
\end{eqnarray} 
The $(2n-1)!!$ factor in \eqref{eq:npoint} can be thought
of as the number of ways to connect $2n$ points in pairs, which
is a version of what is known as Wick contraction in the physics
literature.  Using this, $\langle x^{2n} \rangle$ can be computed graphically in terms of Feynman diagrams. 

Let us compute the $4$-pt function: Four vertices can be Wick contracted into two pairs in three distinct ways. Each connected pair contributes a factor of $1/a$ to the Gaussian integral in \eqref{eq:npoint}. The corresponding Feynman diagram, given below, is a sum of three distinct diagrams, each corresponding to a unique Wick contraction, and contributing a factor of $1/a^2$ from the two pairs. 
\begin{center}
  \begin{tikzpicture}[line width=1pt]
    \node at (0, 0) {\, \, \,};
    \node at (1, 0) {\, \, \,};
    \node at (1, 1) {\, \, \,};
    \node at (0, 1) {\, \, \,};
    \draw[scs] (0, 0) -- (0, 1);
	\draw[scs] (1, 0) -- (1, 1);
	\draw[fill] (0,0) circle (.05);
	\draw[fill] (1,0) circle (.05);
	\draw[fill] (1,1) circle (.05);
	\draw[fill] (0,1) circle (.05);
\draw (0,-1) node{$\bigg(\frac{1}{a}$};
\draw (0.5,-1) node{$.$};
\draw (1,-1) node{$\frac{1}{a}\bigg)$};
\draw (2,0.5) node{$+$};
\draw (2,-1) node{$+$};

    \begin{scope}[shift={(3, 0)}]
      \node at (0, 0) {\, \, \,};
      \node at (1, 0) {\, \, \,};
      \node at (1, 1) {\, \, \,};
      \node at (0, 1) {\, \, \,};
      \draw (0, 0) -- (1, 0);
	  \draw (0, 1) -- (1, 1);
	  \draw[fill] (0,0) circle (.05);
	  \draw[fill] (1,0) circle (.05);
	  \draw[fill] (1,1) circle (.05);
	  \draw[fill] (0,1) circle (.05);
    \end{scope}
    \draw (3,-1) node{$\bigg(\frac{1}{a}$};
\draw (3.5,-1) node{$.$};
\draw (4,-1) node{$\frac{1}{a}\bigg)$};
    \draw (5,0.5) node{$+$};
    \draw (5,-1) node{$+$};

   \begin{scope}[shift={(6, 0)}]
      \node at (0, 0) {\, \, \,};
      \node at (1, 0) {\, \, \,};
      \node at (1, 1) {\, \, \,};
      \node at (0, 1) {\, \, \,};
      \draw (0, 0) -- (0.4, 0.4);
      \draw (0.6, 0.6) -- (1, 1);
	  \draw (1, 0) -- (0, 1);
	  \draw[fill] (0,0) circle (.05);
	  \draw[fill] (1,0) circle (.05);
	  \draw[fill] (1,1) circle (.05);
	  \draw[fill] (0,1) circle (.05);
    \end{scope}
    \draw (6,-1) node{$\bigg(\frac{1}{a}$};
\draw (6.5,-1) node{$.$};
\draw (7,-1) node{$\frac{1}{a}\bigg)$};
\draw (8,-1) node{$=$};
\draw (9,-1) node{$\frac{3}{a^2}$};
  \end{tikzpicture}
\end{center} 
The total of three diagrams, $3/a^2$, matches with \eqref{eq:npoint} for $n=2$.

More generally, a Gaussian integral including a source term, described as the one below,
\begin{eqnarray}
\int^{+\infty}_{-\infty} dx\,\exp{\left(-\frac12 ax^2 + Jx\right)} \label{eq:source},
\end{eqnarray} 
can be evaluated by completing the square in the exponent: $-ax^2/2 + Jx = -(a/2)(x^2 - 2Jx/a) = -(a/2)(x-J/a)^2 + J^2/2a$, and shifting the variable to $x \rightarrow y =x + J/a $, giving
\begin{eqnarray}
\int^{+\infty}_{-\infty} dx\,\exp{\left(-\frac12 ax^2 + Jx\right)} &=& \int^{+\infty}_{-\infty} dy\,\exp{\left(-\frac{a}{2} y^2 + J^2/2a\right)} \nonumber \\
&=& \left( \frac{2\pi}{a} \right)^{\frac12} \exp{(J^2/2a)} \label{eq:sGauss} .
\end{eqnarray}
The $2n$-pt function $\langle x^{2n} \rangle$ can be alternatively calculated by acting with the derivative ${\delta} /{\delta J}$ on \eqref{eq:source} $2n$ times, then setting the source $J=0$ to retrieve the Gaussian integral. 

We can generalize \eqref{eq:sGauss} to a multivariate Gaussian integral by promoting $a$ to a real symmetric $N \times N$ matrix $A_{ij}$ and variable $x$ to a $(N \times 1)$ dimensional vector $x_i$, given below
\begin{eqnarray}
\int^{+\infty}_{-\infty}\ldots \int^{+\infty}_{-\infty}\,\int^{+\infty}_{-\infty} dx_1\,dx_2\dots dx_N \,\exp{\left(-\frac12x\cdot A \cdot x + J\cdot x\right)}= \left(\frac{(2\pi)^N}{\det{|A|}} \right)^{\frac{1}{2}} \exp{\left(\frac12 J A^{-1}J\right)} . \label{eq:mGauss}
\end{eqnarray}
Here $x\cdot A\cdot x = x_iA_{ij}x_j$ and $J\cdot x = J_ix_i$. This can be exactly computed by an orthogonal basis transformation $O$ and a diagonal matrix $D$, to give $A = O^{-1} D  O$ and $y_i = O_{ij}x_j$ in the $N$-dimensional Euclidean space, to result in
\begin{eqnarray}
\label{eq:multivariateGauss}
\int^{+\infty}_{-\infty}\ldots \int^{+\infty}_{-\infty}\,\int^{+\infty}_{-\infty} dy_1\,dy_2\dots dy_N\, \exp{\left(-\frac12y D y + (OJ) y\right)}
&=& \prod_{i = 1}^{N} \, dy_i \, \exp{\left(-\frac{1}{2}D_{ii}y^2_i + (OJ)_i y_i \right)} \nonumber \\
&=& \left(\frac{(2\pi)^N}{\det{[A]}} \right)^{\frac{1}{2}} \exp{\left(\frac12(OJ) D^{-1} (OJ)\right)} \nonumber \\ &=& \left(\frac{(2\pi)^N}{\det{[A]}} \right)^{\frac{1}{2}} \exp{\left(J (O^{-1}D^{-1}O) J \right)} \nonumber \\
&=& \left(\frac{(2\pi)^N}{\det{[A]}} \right)^{\frac{1}{2}} \exp{\left(\frac12 J A^{-1} J\right)} .
\end{eqnarray}

The general expression of the $n$-pt function can be obtained by applying Wick contraction to \eqref{eq:multivariateGauss}, to give 
\begin{eqnarray}
\langle x_1\,x_2\dots x_{n-1}\, x_n \rangle &=& \frac{\int^{+\infty}_{-\infty}\ldots \int^{+\infty}_{-\infty}\,\int^{+\infty}_{-\infty} \,dx_1\,dx_2\dots dx_N \,\exp{(-\frac12x A x)}\,x_1x_2\dots x_{n-1} x_n }{\int^{+\infty}_{-\infty}\ldots \int^{+\infty}_{-\infty}\,\int^{+\infty}_{-\infty} \,dx_1\,dx_2\dots dx_N \,\exp{(-\frac12x A x)} } \\
&=& \sum_{\text{Wick pairs}} \langle x_a\,x_b\rangle \dots \langle x_c\, x_d \rangle = \sum_{\text{Wick pairs}} (A^{-1})_{ab}\dots (A^{-1})_{cd}\,\, .
\end{eqnarray} 
For $n=1$ and $n=2$ these simplify into
\begin{eqnarray}
\langle x_i \rangle &=& \frac{\int^{+\infty}_{-\infty}\ldots \int^{+\infty}_{-\infty}\,\int^{+\infty}_{-\infty} \,dx_1\,dx_2\dots dx_N \,\exp{(-\frac12x A x)}\,x_i }{\int^{+\infty}_{-\infty}\ldots \int^{+\infty}_{-\infty}\,\int^{+\infty}_{-\infty} \,dx_1\,dx_2\dots dx_N \,\exp{(-\frac12x A x)} } = 0 \,\,, \\
\langle x_i\,x_j \rangle &=& \frac{\int^{+\infty}_{-\infty}\ldots \int^{+\infty}_{-\infty}\,\int^{+\infty}_{-\infty} \,dx_1\,dx_2\dots dx_N \,\exp{(-\frac12x A x)}\,x_ix_j }{\int^{+\infty}_{-\infty}\ldots \int^{+\infty}_{-\infty}\,\int^{+\infty}_{-\infty} \,dx_1\,dx_2\dots dx_N \,\exp{(-\frac12x A x)} } = A^{-1}_{ij} \,\,.
\end{eqnarray}

\subsection*{Continuous Number of Variables}
When the variables describing a Gaussian process are promoted to be continuous, the entire process can be defined in terms of a partition function, given by  
\begin{eqnarray}
\label{partition}
{Z}_{GP} &=& \frac{1}{{Z}_{GP,0}} \int df e^{-\frac12 \int d^d x \, d^d y \,\,  f(x) \Xi(x,y) f(y) -\frac12 \int d^d x \, J(x) f(x) -\frac12 \int d^d y \, J(y) f(y)} \nonumber  \\
&=& \frac{1}{{Z}_{GP,0}}{\int df e^{-({S}_{GP} + \Delta {S})} } \,\,,  
\eea
where
\bea
{S}_{GP} &=& \frac12 \int d^d x \, d^d y \,\,  f(x) \Xi(x,y) f(y)  \nonumber \\
\Delta {S} &=& \frac12 \int d^d x \, J(x) f(x) + \frac12\int d^d y J(y) f(y) \nonumber 
\eea
and ${Z}_{GP, 0} := Z_{GP}[J=0] = \int df e^{-{S}_{GP}}$ is the normalization constant.

$\Xi(x,y)$, a real symmetric $2$-tensor, is the continuous version of the inverse of the covariance matrix. We can evaluate \eqref{partition} by a basis transformation such that $\Xi(x,y)$ is diagonal. However, here we choose a more direct evaluation of $Z_{GP}$ by noting that it can be expressed as the following 
\begin{eqnarray}
\label{eqn:continuous_eval}  
\int df e^{-(S_{GP}+\Delta S)} &=& 
  e^{\frac12 \int d^dx \, d^d y \,  J(x) K(x,y) J(y)} \times \nonumber \\
  \int&df& e^{-\frac12 \int d^d x \, d^d y \,\, \left(f(x) + \int d^dw\, J(w) K(w,x)\right)\,\, \Xi(x,y)\,\,\left(f(y)+\int d^dz \, K(y,z) J(z)\right)} .
\end{eqnarray}
The terms in the last exponential of \eqref{eqn:continuous_eval} can be expanded to obtain 
\begin{align}
-\frac12 \int d^d x \, d^d y \, d^dw \, J(w) K(w,x) \Xi(x,y) f(y) & = -\frac12 \int d^dy \, d^dw \, J(w)\delta(w-y) f(y) \\ \nonumber &= -\frac12 \int d^dy \, J(y) f(y),
\end{align}
and a similar identity for $J(x)f(x)$. The remaining term in the last exponential of \eqref{eqn:continuous_eval} is simplified to be
\begin{align}
-\frac12 \int d^dx \, d^dy \, d^dw \, d^dz \,\, J(w)K(w,x)\Xi(x,y)K(y,z)J(z) \\ = -\frac12 \int \, d^dx \, d^dy\, J(x) K(x,y) J(y) .
\end{align}
The last identity arises from the $\delta^{(d)}(w-y)$ resulting from $x$-integral evaluation, followed by renaming integration variables to $x$ and $y$. Using these, the integral over output space on in \eqref{eqn:continuous_eval} can be simplified. Further, by diagonalization ${Z}_{GP,0} = \left[2\pi/\det{({\Xi})} \right]^{1/2}$. Putting it all together, we have the integrated form of the partition function
\begin{equation}
  {Z_{GP}}= \exp{\left(\frac12\int d^d x \, d^d y\, J(x) K(x,y) J(y) \right) } \label{eqn:GPint}, 
\end{equation}
where $K(x,y)$ is the functional / operator inverse of $\Xi(x,y)$, i.e. $\int dy\, K(x,y)\Xi(y,z) = \delta(x-z)$.

The $n$-pt function of the Gaussian process over continuous variables is given by
\begin{eqnarray}
G^{(n)}(x_1, \dots, x_n) &=& \frac{\int df\, f(x_1)\dots f(x_n)\, e^{-(S_{GP} + \Delta S)} }{Z_{GP,0}}.
\end{eqnarray}
It may be computed in a canonical way by taking functional $J$-derivatives,
\begin{eqnarray}
	G^{(n)}(x_1, \dots, x_n)   &=& \left[ \left(-\frac{\delta}{\delta J(x_1)}\right)\dots \left(-\frac{\delta}{\delta J(x_n)}\right) \frac{ \int df\, e^{-(S_{GP} + \Delta S)} }{Z_0}\right] \bigg|_{J=0} \nonumber \\
  &=& \left[ \left(-\frac{\delta}{\delta J(x_1)}\right)\dots \left(-\frac{\delta}{\delta J(x_n)}\right) Z_{GP}\right] \bigg|_{J=0} \, ,
\end{eqnarray}
which can be directly evaluated from \eqref{eqn:GPint}.

 For the sake of thoroughness we demonstrate the calculations of some of the $n$-pt 
 functions.  
The $1$-pt function is
 \begin{eqnarray}
G^{(1)}(x_1) &=&  \mathbb{E}[f(x_1)] =-\frac{\delta}{\delta J(x_1)}{Z_{GP}} \Bigg|_{J=0} = 0 ,
\end{eqnarray}
where the derivative results in a factor of $J$, which sets the $1$-pt function to $0$.
The $2$-pt function is given by 
\begin{eqnarray} \label{two_pt_kernel}
G^{(2)}(x_1,x_2) &=& \mathbb{E}[f(x_1)f(x_2)] = \Bigg[ (-1)^2 \frac{\delta}{\delta J(x_1)}\frac{\delta}{\delta J(x_2)} \exp{\left(\frac12 \int d^dx \, d^dy\, J(x)K(x,y) J(y)\right)} \Bigg] \Bigg|_{J=0} \nonumber \\
&=& \frac{\delta}{\delta J(x_1)} \Bigg[  e^{ \int d^dx \,  d^dy\, J(x)\,K J(y)} \int d^d x \, d^d y \,\bigg(\frac12 \delta(x_2-x) K(x,y) J(y) \nonumber \\
&+& \frac12 \delta(x_2-y) J(x)K(x,y) \bigg) \Bigg] \Bigg|_{J=0} \nonumber \\
&=& \frac{1}{2} \int d^dy\, \delta(x_1-y)K(x_2, y) + \frac{1}{2} \int d^dx\, \delta(x_1-x)K(x,x_2)  \nonumber \\
&=& \frac12 K(x_2,x_1) + \frac12 K(x_1,x_2) \nonumber \\
&=& K(x_1,x_2)  .
\end{eqnarray} 
The last line follows from the symmetry of the covariance $K(x_1,x_2)$.

The general expression of the $n$-pt function can be obtained by a similar calculation; it is given by
\begin{eqnarray} 
\label{n_odd}
G^{(n)}(x_1,\dots , x_n) &=& \mathbb{E}[f(x_1) \dots f(x_n)] = \Bigg[ (-)^n \frac{\delta}{\delta J(x_1)}\ldots \frac{\delta}{\delta J(x_n)}{Z_{GP}} \Bigg] \Bigg|_{J=0} \nonumber \\
&=& \sum_{\text{Wick pairs}}\,\mathbb{E}[f(x_a)f(x_b)] \dots \mathbb{E}[f(x_c)f(x_d)] \nonumber \\
&=& \sum_{\text{Wick pairs}}\, K(x_a,x_b) \dots K(x_c,x_d)~,~~\text{if $n$ even} \\
&=& 0~,~~\text{if $n$ odd}.
\end{eqnarray}

From \eqref{n_odd} we deduce that the $3$-pt function of a Gaussian process is identically zero. And the $4$-pt
function is given by
\begin{equation}
G^{(4)}(x_1,\dots,x_4) = K(x_1,x_2)K(x_3,x_4) + K(x_1,x_3)K(x_2,x_4) +K(x_1,x_4)K(x_2,x_3),
\end{equation}
which may be expressed in terms of Feynman diagrams, as done in \eqref{4-ptwick}.

\subsection*{Non-Gaussian Integrals via Perturbation Theory} 
Small perturbations away from the Gaussian process can still be understood in terms of order-by-order perturbative corrections to the Gaussian integral. Let us assume that the small perturbations can be described in terms of correction terms $g_nf(x)^n$ to $S_{GP}$. The associated new partition function is 
\begin{eqnarray}
&~&{Z}[g_n, J] = \frac{ \int df\,e^{-\frac12\int d^dx_1 d^dx_2\,f(x_1)\Xi(x_1,x_2)f(x_2) -\frac12\int d^dx_1\,J(x_1)f(x_1) - \frac12\int d^dx_2\,J(x_2)f(x_2)  - \int d^dx\, g_n f(x)^n } }{\int df\,e^{\frac12\int d^dx_1 d^dx_2\,f(x_1)\Xi(x_1,x_2)f(x_2) - \int d^dx \,g_n f(x)^n  }} \nonumber \\
&=& \frac{ \int df \, \sum_{m=0}^{\infty} \frac{(-1)^m}{m!} (g_n)^m [\int d^dx \,f(x)^n]^m e^{-\frac12\int d^dx_1 d^dx_2\,f(x_1)\Xi(x_1,x_2)f(x_2) -\frac12\int d^dx_1\,J(x_1)f(x_1) - \frac12\int d^dx_2\,J(x_2)f(x_2)} }{{Z}[g_n,J=0]} \nonumber \\
&=& \frac{ \int df \, \sum_{m=0}^{\infty} \frac{(-1)^m}{m!} (g_n)^m \left[\int d^dx \left(-\frac{\delta}{\delta J(x)}\right)^n\right]^m e^{-\frac12\int d^dx_1 d^dx_2\,f(x_1)\Xi(x_1,x_2)f(x_2) -\frac12\int d^dx_1\,J(x_1)f(x_1) - \frac12\int d^dx_2\,J(x_2)f(x_2)} }{{Z}[g_n,J=0]} \nonumber \\
&=& \frac{ \int df \, \sum_{m=0}^{\infty} \frac{(-1)^m}{m!} (g_n)^m \left[\int d^dx \left(-\frac{\delta}{\delta J(x)}\right)^n\right]^m e^{\frac12\int d^dx_1 d^dx_2\,J(x_1)K(x_1,x_2)J(x_2)}}{{Z}[g_n,J=0]} \nonumber \\
&=&\frac{\int df\, \sum_{s=0}^{\infty}\frac{1}{s!}\int d^dx_1\dots \int d^dx_s J(x_1)\dots J(x_s)e^{-{S}_{GP} - \int d^dx \,g_n \, f(x)^n} \, f(x_1)\dots f(x_s)}{ {Z}[g_n,J=0]} \nonumber \\
&=&{ \sum_{s=0}^{\infty}\frac{1}{s!}\int d^dx_1\dots \int d^dx_s J(x_1)\dots J(x_s) G^{(s)}(x_1,\dots x_s) }
\end{eqnarray}
where 
\begin{eqnarray}
\label{Gs}
G^{(s)}(x_1,\dots , x_s) &=& \frac{\int df\, e^{-{S}_{GP} - \int d^dx \,g_n f(x)^n} \, f(x_1)\dots f(x_s)}{{Z}[J=0, g_n]} \nonumber \\
&=&\Bigg( {\frac{\int df\, e^{-{S}_{GP} - \int d^dx \,g_n \,f(x)^n} \, f(x_1)\dots f(x_s)}{{Z}_0[J=0, g_n = 0]} }\Bigg) \Bigg/ \Bigg({ \frac{{Z}[J=0, g_n]}{{Z}_0[J=0, g_n = 0]} }\Bigg) 
\end{eqnarray}
The numerator and the denominator in \eqref{Gs} can be separately obtained using Wick contractions and Feynman diagrams; they both contain some diagrams where none of the internal vertices are  connected to any of the external vertices. Such diagrams are called ``vacuum bubbles'', and any diagram containing a vacuum bubble exactly cancels from the numerator and denominator upon series expansions, thus not contributing to the actual $n$-pt functions. 

We take an example where the non-Gaussian process is described by small perturbations away from the GP by $\Delta {S} = \int d^dx \, [g f(x)^3 + \lambda f(x)^4]$. The denominator of \eqref{Gs} is expanded below, in terms of Feynman diagrams, up to quadratic order terms in coupling constants $g$ and $\lambda$. 
\begin{eqnarray}
\mathbb{1} &-& 
\lambda  \Bigg[3\,
 \,\Bigg] \,\,.
\end{eqnarray}
The three point vertices correspond to coupling constant $g$ and occur at internal points labeled by $w$, and the four point vertices correspond to $\lambda$ and occur at points labeled by $y$. 

The numerator of the $1$-pt function, up to same orders of correction terms, is given below:
\begin{eqnarray} 
&-g& \Bigg[\,3\, \,
%
  
\Bigg] .
\end{eqnarray} 
After vacuum bubbles cancel from both the numerator and denominator, the $1$-pt function to quadratic orders in $g$ and $\lambda$ is
\begin{eqnarray} G^{(1)}(x_1) &=&
	-g  \Bigg[\,3\, \,
	%
 \,\,
	\Bigg] .
	\end{eqnarray}

Similarly, the numerator of the 2-pt function is obtained to be
\begin{eqnarray}
	%
\,\Bigg]\,\, .
\end{eqnarray}
After canceling all vacuum bubbles, the exact 2-pt function is
\begin{eqnarray}
	G^{(2)}(x_1,x_2) &=& 
	%
 
\,\Bigg ] \, . \nonumber
\end{eqnarray}

Each of the higher order $n$-pt functions contains different topologically distinct Feynman diagrams, with various permutations of external vertices; each of these diagrams should be treated separately when external vertices are fixed. However, for illustration, we present such topologically distinct Feynman diagrams ignoring all external labels, and take care of the combinatorics of external vertices by multiplying with the total number of copies. 

The numerator of the $3$-pt function is given by the following diagrams:
\begin{eqnarray}
&-g&\Bigg[\,9\, \,
 \, \, \, \Bigg] .
\end{eqnarray}

The final expression for $3$-pt function is 
\begin{eqnarray}
G^{(3)}(x_1,x_2,x_3) &=& -g\Bigg[\,9\, \,
%
\,\,\,
  \Bigg]
\end{eqnarray}

Following similar analysis the $4$-pt function is given by
\begin{eqnarray}
&G^{(4)}(x_1,x_2, x_3, x_4) = \, 3\,\, \,%
 
\,\, \, \Bigg]\,\, .
\end{eqnarray}

\section{$2$-pt Functions / Kernels of Example Networks} \label{kernel_derivation_appendix}
A neural network with activation function $\phi(x)$ and one hidden layer has the following output function
\begin{equation}
    f(x) = W_{1} \,  \phi \big( W_0 \, x + b_{0} \big) + b_{1}  .
\end{equation}
When the weights and biases $W_0$, $W_1$, $b_0$, $b_1$ are i.i.d. and drawn from a Gaussian distribution with mean $0$ and standard deviations $\sigma_{W_0}$, $\sigma_{W_1}$, $\sigma_{b_0}$, $\sigma_{b_1}$ respectively, the kernel or $2$-pt function is given by
\begin{eqnarray}
\mathbb{E}[f(x) f(x^{\prime})] = \sigma_{b_1}^2 + \sigma_{W_1}^2\, V_{\phi}[\phi(W_0 \, x + b_0),\phi(W_0 \, x^\prime + b_0) ] \, .
\end{eqnarray}

The $2$-pt function of the post-activation $V_{\phi}[\phi(W_0 \, x + b_0),\phi(W_0 \, x^\prime + b_0) ]$
can be evaluated by two methods: the first method is exact at any width, prescribed in \cite{williams}; and the second method is true in the GP limit, described in \cite{yang}. We will refer to the two methods by superscripts ``Williams'' and ``Yang'' respectively. They are the following
\begin{eqnarray}
V^{\text{Williams}}_{\phi} (x, x^\prime)= \frac{ \int\, \phi(W_0 \, x + b_0) \, \phi(W_0 \, x^\prime + b_0) e^{-\frac{1}{2} W_0^T\sigma^{-2}_{W_0} W_0 -\frac{1}{2} b_0^T\sigma^{-2}_{b_0} b_0 }\,dW_0\,db_0 }{ \int\,  \exp{\Big(-\frac{1}{2} W_0^T\sigma^{-2}_{W_0} W_0 -\frac{1}{2} b_0^T\sigma^{-2}_{b_0} b_0 \Big)}\,dW_0 \,db_0} \label{william} ,
\end{eqnarray}
and
\begin{equation}
V^{\text{Yang}}_{\phi}(x, x^\prime) = \mathbb{E}_{\substack{f \sim \mathcal{N}(0,K)} } [\phi(W_0 \, x + b_0),\phi(W_0 \, x^\prime + b_0) ]  
 \label{yang} ,
\end{equation}
where $K$ is the kernel or covariance function of the Gaussian distribution of $f$ in the GP limit. For the three NN architectures discussed in this paper, kernels evaluated by both prescriptions are shown to agree in the limit of infinite width, i.e. GP.

\subsection*{Erf-net}
We now turn to the case of Erf activation, which is given by 
\begin{equation}
\phi (x) = \frac{2}{\sqrt{\pi}} \int^x_0 \exp(-t^2)\,dt \label{erfact} .
\end{equation}
At any width $N$, the associated kernel can be obtained by the method in \cite{williams}. This involves substituting \eqref{erfact} in \eqref{william}, followed by computing the Gaussian integral and a transformation of variables. The kernel of Erf-net by this method is obtained to be
\begin{eqnarray}
\label{erfkernelappendix}
V^{\text{Williams}}_{\text{Erf-net}}[f(x)f(x^\prime)] = \sigma_{b_1}^2 + \sigma_{W_1}^2N\, \frac{2}{\pi} \arcsin \Bigg[ \frac{  2K(x,x^\prime) }{\sqrt{\left( 1 +  2K(x,x)\right) \left( 1 + 2K(x^\prime, x^\prime)\right)  }} \Bigg] .
\end{eqnarray}
\amnew{Normalization $\sigma_{W_1} = \frac{\sigma_{W_0}}{\sqrt{N}}$ is chosen to ensure width invariance of the kernel.}
The intermediate expressions, such as $K(x, x^\prime)$, are the kernels of the linear layer $W_0 \, x + b_0$, given below
\begin{eqnarray}
K(x,x) = \sigma_{b_0}^2 + \sigma_{W_0}^2\, x\cdot x~, ~~\,\, K(x^\prime, x^\prime) = \sigma_{b_0}^2 + \sigma_{W_0}^2\,x^\prime \cdot x^\prime~,~~\,\, K(x, x^\prime) = \sigma_{b_0}^2 + \sigma_{W_0}^2\, x\cdot x^\prime   \label{innerkernel} .
\end{eqnarray}

In the limit of infinite width, the method in \cite{yang} can be used. This involves substituting \eqref{erfact} in \eqref{yang}, defined in terms of PDF of output $f(x)$. Evaluating the Gaussian integral, followed by a change of variables, results in the same expressions as \eqref{erfkernelappendix} for $V^{\text{Yang}}_{\text{Erf-net}}(x,x^\prime)$. This shows that the Erf-net kernel is exact at any width \amnew{when $\sigma_{W_1} = \frac{\sigma_{W_0}}{\sqrt{N}}$}, given by
\begin{eqnarray}
\mathbb{E}_{\text{Erf-net}}[f(x)f(x^\prime)]= \sigma_{b_1}^2 + \sigma_{W_0}^2\, \frac{2}{\pi} \arcsin \Bigg[ \frac{  2(\sigma_{b_0}^2 + \sigma_{W_0}^2\, x\cdot x^\prime ) }{\sqrt{\left( 1 +  2(\sigma_{b_0}^2 + \sigma_{W_0}^2\, x\cdot x )\right) \left( 1 + 2(\sigma_{b_0}^2 + \sigma_{W_0}^2\,x^\prime \cdot x^\prime ) \right)  }} \Bigg] .
\end{eqnarray}

\subsection*{ReLU-net}
Next we study the case of ReLU activation, which is given by
\begin{equation}
\phi (x) = \max({0,x}) \label{reluact} .
\end{equation}
At finite width $N$, \eqref{reluact} can be substituted into \eqref{william}, followed by a rearrangement in terms of Heaviside function $\Theta(x) = \frac{1}{2}(1+ \text{sgn}{(x)})$, to obtain
 \begin{eqnarray} 
V_{\text{ReLU-net}}^{\text{William}}(x, x^\prime) = \frac{ \int\, \Theta(W_0 \, x + b_0)\,(W_0 \, x + b_0) \, \Theta(W_0 \, x^\prime + b_0)\,(W_0 \, x^\prime + b_0) e^{-\frac{1}{2} (W_0^T\sigma^{-2}_{W_0} W_0 + b_0^T\sigma^{-2}_{b_0} b_0 )}\,dW_0\,db_0 }{ \int\,  \exp{\Big(-\frac{1}{2} W_0^T\sigma^{-2}_{W_0} W_0 -\frac{1}{2} b_0^T\sigma^{-2}_{b_0} b_0 \Big)}\,dW_0 \,db_0} \label{chosaul} \nonumber .
\end{eqnarray}
The arguments of the Heaviside function are chosen to lie in the first quadrant. After a basis transformation it results in the following 
 \begin{eqnarray} 
V_{\text{ReLU-net}}^{\text{William}}(x, x^\prime) = \frac{N \,(\sin \theta)^3}{2\pi} \sqrt{K(x,x)K(x^\prime, x^\prime)} \int\,da\,db\, a\,b\, \exp \left(-\frac{1}{2}|a|^2 -\frac{1}{2}|b|^2 - a\cdot b\,\cos \theta \right)  .
\end{eqnarray}
Here $\cos \theta = \frac{K(x, x^\prime)}{\sqrt{K(x,x)K(x^\prime, x^\prime)}}$ and intermediate kernels $K(x,x)$, $K(x,x^\prime)$ and $K(x^\prime, x^\prime)$ of the linear layer $W_0 \, x + b_0$  are defined in \eqref{innerkernel}. A further change of variables $ a=r\cos \left(\frac{\psi}{2} + \frac{\pi}{4} \right)$ and $b=r\sin \left(\frac{\psi}{2} + \frac{\pi}{4} \right)$, followed by integrating out the variable $r$, results in
 \begin{eqnarray} 
V_{\text{ReLU-net}}^{\text{William}}(x, x^\prime) = \frac{N\,(\sin \theta)^3}{2\pi} \sqrt{K(x,x)K(x^\prime, x^\prime)} \int^{\pi/2}_{0} \frac{d\psi\, \, \cos \psi}{( 1 - \cos \psi \cos \theta)^2} \,\,. \label{reluintermediate}
\end{eqnarray}
Taking the derivative of the following trigonometric identity with respect to $\cos \theta$
\begin{eqnarray} 
 \int^{\pi/2}_{0} \frac{d\psi\, }{ 1 - \cos \psi \cos \theta} = \frac{\pi - \theta}{\sin \theta} \,,
\end{eqnarray}
and substituting it in \eqref{reluintermediate}, we obtain 
\begin{eqnarray} 
\label{relufinitwidth}
V_{\text{ReLU-net}}^{\text{William}}(x, x^\prime) = \frac{N}{2\pi} \sqrt{K(x,x)K(x^\prime, x^\prime)} \left(\sin \theta + (\pi - \theta)\cos \theta \right)
\end{eqnarray}
The final ReLU-net kernel, \amnew{after setting $\sigma_{W_1} = \frac{\sigma_{W_0}}{\sqrt{N}}$,} is given by
\begin{eqnarray}  \label{deriverelukernel}
\mathbb{E}_{\text{ReLU-net}}[f(x)f(x^\prime)] &=& \sigma_{b_1}^2 + \sigma_{W_0}^2\, \frac{1}{2\pi} \sqrt{(\sigma_{b_0}^2 + \sigma_{W_0}^2\, x\cdot x)(\sigma_{b_0}^2 + \sigma_{W_0}^2\, x^\prime\cdot x^\prime)}(\sin \theta + (\pi - \theta)\cos\theta ) \nonumber \\
\theta &=& \arccos \Bigg[\frac{\sigma_{b_0}^2 + \sigma_{W_0}^2\, x\cdot x^\prime}{\sqrt{(\sigma_{b_0}^2 + \sigma_{W_0}^2\, x\cdot x)(\sigma_{b_0}^2 + \sigma_{W_0}^2\, x^\prime\cdot x^\prime)}} \Bigg] .
\end{eqnarray}

In the infinite width limit, the $2$-pt function of the hidden layer, given in terms of PDF of the output $f(x)$ following prescription in \cite{yang}, can be expressed in terms of Heaviside function as well:
\begin{eqnarray}
V^{\text{Yang}}_{\text{ReLU-net}} (x, x^\prime)= \frac{ \int\, df\, \Theta(f(x)) \Theta(f(x^\prime)) f(x) f(x^\prime)\, e^{-\frac{1}{2}f(x)\, K^{-1}(x,x^\prime)\, f(x^\prime) } }{\int\, df\,  e^{-\frac{1}{2}f(x)\, K^{-1}(x,x^\prime)\, f(x^\prime)}} .
\end{eqnarray}
After a similar change of variables and basis transformations, this results in the same expression as \eqref{relufinitwidth} at infinite width. This shows that \eqref{deriverelukernel} is the exact kernel for ReLU-net at any width.
             
\subsection*{\gnet}
We introduce a new activation function in order to obtain a translation invariant GP kernel. This architecture is obtained by having a normalization layer after an initial exponential activation, as the following
\begin{eqnarray}
x \rightarrow \exp(W_0 \, x + b_0 ) \rightarrow \frac{\exp(W_0 \, x + b_0 )}{\sqrt{K_{\text{exp}}(x,x)}} \nonumber \\
\implies f(x) = W_1 \Bigg(\frac{\exp(W_0 \, x + b_0 )}{\sqrt{K_{\text{exp}}(x,x)}} \Bigg) + b_1 \,,
\label{gnetactivationappendix}
\end{eqnarray}
where $K_{\text{exp}}(x,x) = V_{\text{exp}}(x,x) = \exp{ [ 2(\sigma_{b}^{2} + {\sigma_{W_0}^{2}} x^{2}) ] }$ is the $2$-pt function of the  intermediate exponential activation layer given by $\phi ^\prime(x) = \exp(W_0 \cdot x + b_0)$.

The resulting activation
is\footnote{We thank Greg Yang for discussions of activations
that yield translationally invariant kernels.}
\begin{eqnarray}
  \phi(x) = \frac{\exp{(W \, x + b)}}{\sqrt{K_{\text{exp}}(x,x)} }.
\end{eqnarray}
It is easy to check that the final kernel of \gnet~architecture is given by
\begin{eqnarray}
\mathbb{E}_{\text{\gnet}}[f(x)f(x^\prime)] = \sigma_{b_1}^2 + \sigma_{W_1}^2\, N \exp\left[- \frac{\sigma^2_{W_0}|x - x^\prime |^2}{2} \right].
\end{eqnarray}

At width $1$, the kernel of the exponential activation layer can be obtained by substituting $\phi^\prime (x) = \exp(W_0 \cdot x + b_0)$ in \eqref{william}, followed by computing the Gaussian integral, to obtain
 \begin{eqnarray}
 \label{expkernel}
V^{\text{William}}_{\text{exp}}(x, x^\prime) =\exp\left( \frac{1}{2}(K(x,x) + 2 K(x,x^\prime) + K(x^\prime, x^\prime)) \right) ,
\end{eqnarray}
where $K(x,x)$, $K(x,x^\prime)$ and $K(x^\prime, x^\prime)$ are the intermediate kernels for linear layer $W_0 \, x + b_0$, as defined in \eqref{innerkernel}. \amnew{Substituting \eqref{expkernel} in \eqref{william} for activation \eqref{gnetactivationappendix} results in the following.}
\begin{eqnarray}
V^{\text{William}}_{\text{\gnet}}(x, x^\prime) &=& \frac{N\, V^{\text{William}}_{\text{exp}}(x, x^\prime) }{\sqrt{V^{\text{William}}_{\text{exp}}(x, x) V^{\text{William}}_{\text{exp}}(x^\prime, x^\prime) } } \nonumber \\
&=& N \exp\left[- \frac{\sigma^2_{W_0}|x - x^\prime |^2}{2} \right].
\end{eqnarray}

In the infinite width limit, the kernel or $2$-pt function of the exponential activation layer is obtained using the PDF of output of this layer, following \cite{yang}, as
\begin{eqnarray}
V^{\text{Yang}}_{\text{exp}}(x, x^\prime) &=& \frac{ \int\,df\, e^{f(x)} \, e^{f(x^\prime)}\,e^{-\frac{1}{2} f(x)\, K^{-1}(x,x^\prime)\, f(x^\prime)  }  }{\int \,df\,e^{-\frac{1}{2} f(x)\, K^{-1}(x,x^\prime)\, f(x^\prime)} } \nonumber \\
&=& \frac{ \int\,df\, e^{-\frac{1}{2} f(x)\, K^{-1}(x,x^\prime)\, f(x^\prime) + \int\,dy\,J(y)f(y) }  }{\int \,df\,e^{-\frac{1}{2} f(x)\, K^{-1}(x,x^\prime)\, f(x^\prime)} }
\end{eqnarray}
with $J(y) = [\delta(y-x) + \delta(y - x^\prime)]$. Field theory methods, described in the previous Section, can be used to simplify this and obtain
\begin{eqnarray}
V^{\text{Yang}}_{\text{exp}}(x, x^\prime) &=& \exp\Bigg[\frac{1}{2}\left(\delta(w - x) + \delta(w-x^\prime) \right)K(w,z)\left(\delta(z-x) + \delta(z-x^\prime)\right) \Bigg] \nonumber \\
&=& \exp\left( \frac{1}{2}(K(x,x) + 2 K(x,x^\prime) + K(x^\prime, x^\prime)) \right) .
\end{eqnarray}
\amnew{Setting $\sigma_{W_1} = \frac{\sigma_{W_0}}{\sqrt{N}}$,} the final expression of Gauss-net kernel is given by
\begin{eqnarray} \label{Gaussnet}
\mathbb{E}_{\text{\gnet}}[f(x) f(x^\prime)] = \sigma_{b_1}^2 + \sigma_{W_0}^2\, \exp\left[- \frac{\sigma^2_{W_0}|x - x^\prime |^2}{2} \right],
\end{eqnarray}
which is translation invariant.

\bibliographystyle{unsrt}
\bibliography{refs}

\end{document}